\documentclass[journal=jacsat,manuscript=article]{achemso}

\usepackage{chemformula}
\usepackage[T1]{fontenc}
\usepackage{hyperref}
\usepackage{amsfonts}
\usepackage{amsmath}
\hypersetup{
    colorlinks=true,
    linkcolor=blue,
    urlcolor=blue,
    citecolor=blue
}

\usepackage{booktabs}
\usepackage{tabularx}
\usepackage{array}
\usepackage{ragged2e}
\usepackage{float}            
\usepackage[section]{placeins} 
\newcolumntype{Y}{>{\RaggedRight\arraybackslash}X} 
\newcolumntype{L}{>{\RaggedRight\arraybackslash}p{0.30\textwidth}}
\newcolumntype{M}{>{\RaggedRight\arraybackslash}p{0.28\textwidth}}
\author{Alexander Aghili}
\email{awaghili@ucsc.edu}
\author{Andy Bruce}
\email{acbruce@ucsc.edu}
\author{Daniel Sabo}
\author{Sanya Murdeshwar}
\email{smurdesh@ucsc.edu}
\author{Kevin Bachelor}
\author{Ionut Mistreanu}
\affiliation[University of California, Santa Cruz]
{Baskin Engineering, University of California - Santa Cruz, Santa Cruz, CA}
\author{Ashwin Lokapally}
\author{Razvan Marinescu}
\affiliation[GiwoTech Inc]
{GiwoTech Inc, Cambridge, MA}
\alsoaffiliation[University of California, Santa Cruz]
{Baskin Engineering, University of California - Santa Cruz, Santa Cruz, CA}

\title[MD Benchmark]
  {A Standardized Benchmark for Machine-Learned Molecular Dynamics using Weighted Ensemble Sampling}

\abbreviations{IR,NMR,UV}

\setkeys{acs}{etalmode=truncate, maxauthors = 10}

\begin{document}
\begin{abstract}
The rapid evolution of molecular dynamics (MD) methods, including machine-learned dynamics, has outpaced the development of standardized tools for method validation. Objective comparison between simulation approaches is often hindered by inconsistent evaluation metrics, insufficient sampling of rare conformational states, and the absence of reproducible benchmarks. To address these challenges, we introduce a modular benchmarking framework that systematically evaluates protein MD methods using enhanced sampling analysis. Our approach uses weighted ensemble (WE) sampling via \emph{The Weighted Ensemble Simulation Toolkit with Parallelization and Analysis} (WESTPA), based on progress coordinates derived from \emph{Time-lagged Independent Component Analysis} (TICA), enabling fast and efficient exploration of protein conformational space. The framework includes a flexible, lightweight propagator interface that supports arbitrary simulation engines, allowing both classical force fields and machine learning-based models. Additionally, the framework offers a comprehensive evaluation suite capable of computing more than 19 different metrics and visualizations across a variety of domains. 

We further contribute a dataset of nine diverse proteins, ranging from 10 to 224 residues, that span a variety of folding complexities and topologies. Each protein has been extensively simulated at 300K for one million MD steps per starting point (4 ns). 

To demonstrate the utility of our framework, we perform validation tests using classic MD simulations with implicit solvent and compare protein conformational sampling using a fully trained versus under-trained CGSchNet model. By standardizing evaluation protocols and enabling direct, reproducible comparisons across MD approaches, our open-source platform lays the groundwork for consistent, rigorous benchmarking across the molecular simulation community.

\end{abstract}

\section{Introduction}
Molecular dynamics (MD) is a computational technique used to simulate the behavior of atoms and molecules over time \cite{Karplus2002}. By applying the principles of classical physics, MD tracks how a molecular system evolves, capturing the dynamic motions and interactions between its components. Unlike static representations such as crystal structures or those generated by recent AI models (e.g. AlphaFold\cite{Jumper2021}), MD provides a dynamic view of how biological macromolecules, such as proteins, DNA, and ligand bindings, behave in more realistic, flexible environments \cite{Hollingsworth2018}. This ability to model atomic-level dynamics makes MD a powerful tool in drug discovery \cite{pr9010071}, where it enhances understanding of how drug candidates interact with target proteins. By going beyond static models, MD improves predictions of binding modes, stability, and affinity, and can reveal hidden or transient binding sites that may serve as novel targets for therapeutic intervention \cite{VAJDA20181}.

A central challenge in MD is the limited timescale accessible to atomistic simulations, which, even on powerful hardware, typically takes hours to generate nanoseconds of data \cite{Dror2012}. This duration is far shorter than many biologically relevant processes, which span microseconds to seconds. Enhanced sampling strategies such as metadynamics \cite{doi:10.1073/pnas.202427399,PhysRevLett.100.020603} and simulated tempering \cite{lenner2016continuous} improve sampling efficiency by biasing the system's energy landscape or temperature distribution. Among these approaches, weighted ensemble (WE) simulations \cite{annurev:/content/journals/10.1146/annurev-biophys-070816-033834} \cite{huber1996weighted} further enhance sampling through parallelization and trajectory resampling. WEs run multiple replicas of a system and periodically resample them based on user-defined metrics of conformational space coverage called progress coordinates. This adaptive allocation of computational resources increases the likelihood of observing rare events within tractable timeframes, making WE a powerful tool for capturing critical transitions in complex systems. We utilized WESTPA 2.0 \cite{doi:10.1021/ct5010615} \cite{Russo2022}, an open-source weighted ensemble implementation.

Another key strategy for extending accessible timescales is coarse-graining (CG), which reduces molecular complexity by grouping atoms into coarse-grained beads \cite{Stevens2023}. This simplification lowers computational cost and enables simulations of larger systems over longer durations. Methods such as carbon alpha (\(C\alpha\)) models, which retain only backbone \(C\alpha\) atoms, and the Martini force field \cite{doi:10.1021/jp071097f}, which maps multiple atoms into interaction sites, are commonly used CG approaches. Although CG sacrifices atomic detail, fine-grained interactions, and physical accuracy, it remains a valuable technique when the goal is to explore long-term or large-scale molecular dynamics.

Finally, machine learning (ML) models have recently emerged as compelling alternatives to traditional force fields for both atomistic and coarse-grained MD. In particular, graph neural networks (GNNs) such as SchNet \cite{10.5555/3294771.3294866}, DimeNet \cite{NEURIPS2021_82489c97}, and PhysNet \cite{Unke_2019} leverage molecular graph representations to learn energy landscapes, forces, and chemical properties directly from data. Variants like CGSchNet \cite{Husic2020} \cite{Pelaez2024TorchMDNet} extend these models to CG simulations. While promising for extending timescales and capturing complex interactions, ML-based models still face challenges in ensuring physical consistency with atomistic data, generalization to unseen systems, and maintaining computational efficiency during long-term simulations.

Although machine learning models are rapidly advancing in their capabilities, a standardized framework for evaluating their physical accuracy and computational efficiency has yet to be established. The three methods discussed collectively enhance the ability to approximate atomic dynamics efficiently, thereby offering a foundation for developing a systematic and rigorous evaluation framework. An ideal evaluation would establish 1) a common set of observables and metrics that will be used as yardsticks to measure model progress, 2) a common ground-truth dataset to be used universally across models and 3) a common enhanced sampling methodology that will be used across all potential energy functions, either classical or machine-learned. Such an evaluation would significantly aid progress in classical and machine-learned MD by highlighting the areas for improvement and by providing detailed qualitative and quantitative information.

In this work we present a modular and reproducible benchmarking framework that addresses the aforementioned criteria. Towards 1), we evaluate structural fidelity, slow-mode accuracy, and statistical consistency through a suite of over \textbf{19} metrics and visualizations, which include both global and local metrics on model performance. Towards 2), we release a ground truth dataset of MD trajectories from nine diverse proteins, that exhaustively cover their conformational space. Towards 3), our benchmark leverages weighted ensemble sampling, using WESTPA, to ensure broad conformational coverage. Our flexible benchmark can rigorously evaluate both classical and machine-learned molecular dynamics methods. We demonstrate its utility by comparing fully trained and under-trained CGSchNet models, as well as implicit solvent MD, showing the benchmark can meaningfully differentiate between methods. By standardizing evaluation across a diverse protein dataset, our platform addresses a critical gap in the MD field, enabling fair, reproducible comparisons and accelerating progress in simulation methodology development.

\section{Methods}

\subsection{Overview of Benchmark}

\begin{figure}
  \includegraphics[width=\textwidth]{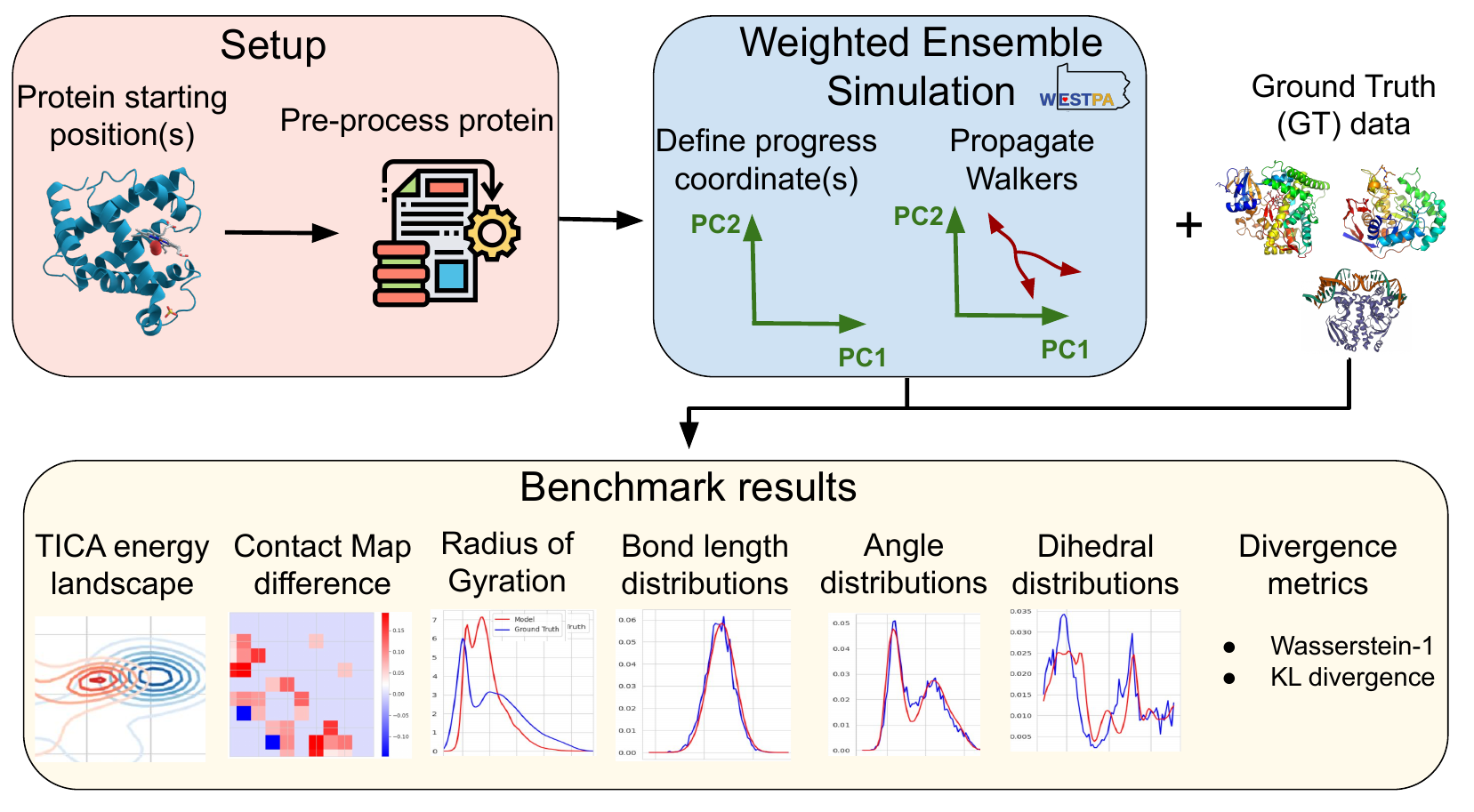}
      \caption{Architecture of our weighted ensemble (WE) benchmarking framework. }
  \label{fgr:system_arch}
\end{figure}

An overview of the benchmark suite is given in Fig. \ref{fgr:system_arch}. We first pre-process protein conformations, and then run a WESTPA-based weighted ensemble simulation, which requires defining the progress coordinate and propagating walkers. We then compare the WE with ground truth data along several dimensions covering both global properties, such as Time-lagged Independent Component Analysis (TICA) \cite{Scherer2015} energy landscapes, contact map differences and distributions for the radius of gyration (RoG), bond lengths, angles and dihedrals. Finally, we also compute quantitative divergence metrics, mainly Wasserstein-1 and Kullback-Leibler divergences across all the relevant analyses.

\newpage

\subsection{Ground Truth Data}
\label{sec_gtdata}

As shown in Table \ref{tab:proteins}, we selected nine proteins spanning diverse folds and sizes from Lindorff-Larsen et al. \cite{LindorffLarsen2011}.
\begin{table}[!ht]
\centering
\resizebox{\textwidth}{!}{%
\begin{tabular}{|lllll|}
\hline
\textbf{Protein} & \textbf{Residues} & \textbf{Fold} & \textbf{Description} & \textbf{PDB} \\
\hline
Chignolin & 10 & \(\beta\)-hairpin & Tests basic secondary structure formation & \href{https://www.rcsb.org/structure/1UAO}{1UAO} \\
Trp-cage & 20 & \(\alpha\)-helix & Fast-folding with hydrophobic collapse & \href{https://www.rcsb.org/structure/1L2Y}{1L2Y} \\
BBA & 28 & \(\beta\beta\alpha\) & Mixed secondary structure competition & \href{https://www.rcsb.org/structure/1FME}{1FME} \\
a3D & 73 & 3-helix & Synthetic helical bundle, tests designability & \href{https://www.rcsb.org/structure/2A3D}{2A3D} \\
Protein B & 53 & 3-helix & Helix packing dynamics & \href{https://www.rcsb.org/structure/1PRB}{1PRB} \\
Protein G & 56 & \(\alpha/\beta\) & Complex topology formation & \href{https://www.rcsb.org/structure/1PGA}{1PGA} \\
\(\lambda\)-repressor & 224 & 5-helix & Tests scalability & \href{https://www.rcsb.org/structure/1LMB}{1LMB} \\
Homeodomain & 54 & 3-helix bundle & DNA-binding domain with stable fold & \href{https://www.rcsb.org/structure/1ENH}{1ENH} \\
WW domain & 37 & \(\beta\)-sheet & Challenging \(\beta\)-sheet topology & \href{https://www.rcsb.org/structure/1E0L}{1E0L} \\
\hline
\end{tabular}
}
\caption{Protein benchmarks with representative structures.}
\label{tab:proteins}
\end{table}

\begin{figure}
  \includegraphics[width=0.65\textwidth]{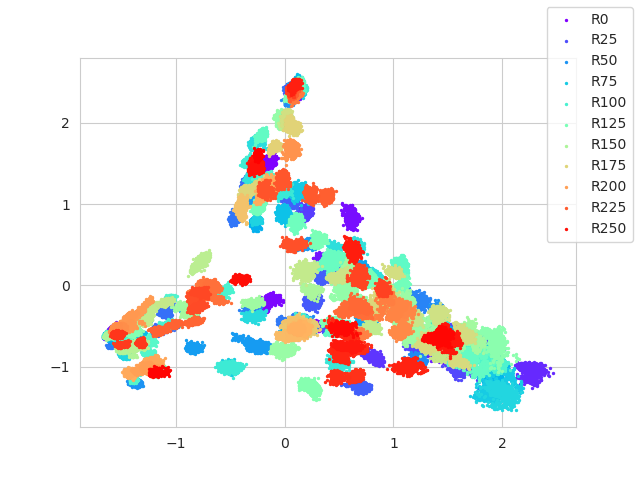}
      \caption{Ground truth trajectories of Protein G in TICA 0/1 space with a 10\% sampling of starting points, colored by starting point number. The legend provides the starting point number using R\# where \# is the starting point associated with the color (divided by 10). This is not representative of the explored space, but shows how little the dynamics will deviate from the original point.}
  \label{fgr:gt_proteing_full}
\end{figure}

To generate reference data against which models will be compared, we ran MD simulations from a total of 9919 different starting points provided by Majewski et al. \cite{Majewski2023} (Prepared at 350K). From each starting point, we ran 1,000,000 steps at a 4 femtosecond (fs) timestep resulting in a total of 4 nanosecond (ns) per starting point at 300K. The number of starting points ran per protein differed ranging from a minimum of 372 with Chignolin to a maximum of 2560 for Protein G. Since the starting points provide a detailed coverage of the conformation space, the simulation length of 1,000,000 steps per starting point was selected to ensure adequate sampling of dynamics across the conformational space for these well-characterized proteins. Each starting point did not explore much beyond its immediate conformations as shown in Fig. \ref{fgr:gt_proteing_full}, but the conformations overlap between all the starting points in the final configuration to construct the full possible conformation transitions as seen in Fig. \ref{fgr:gt_proteing_full}. 

\begin{figure}
  \includegraphics[trim={9.42in 9.35in 2.5cm 2.5cm},clip,scale=0.85]{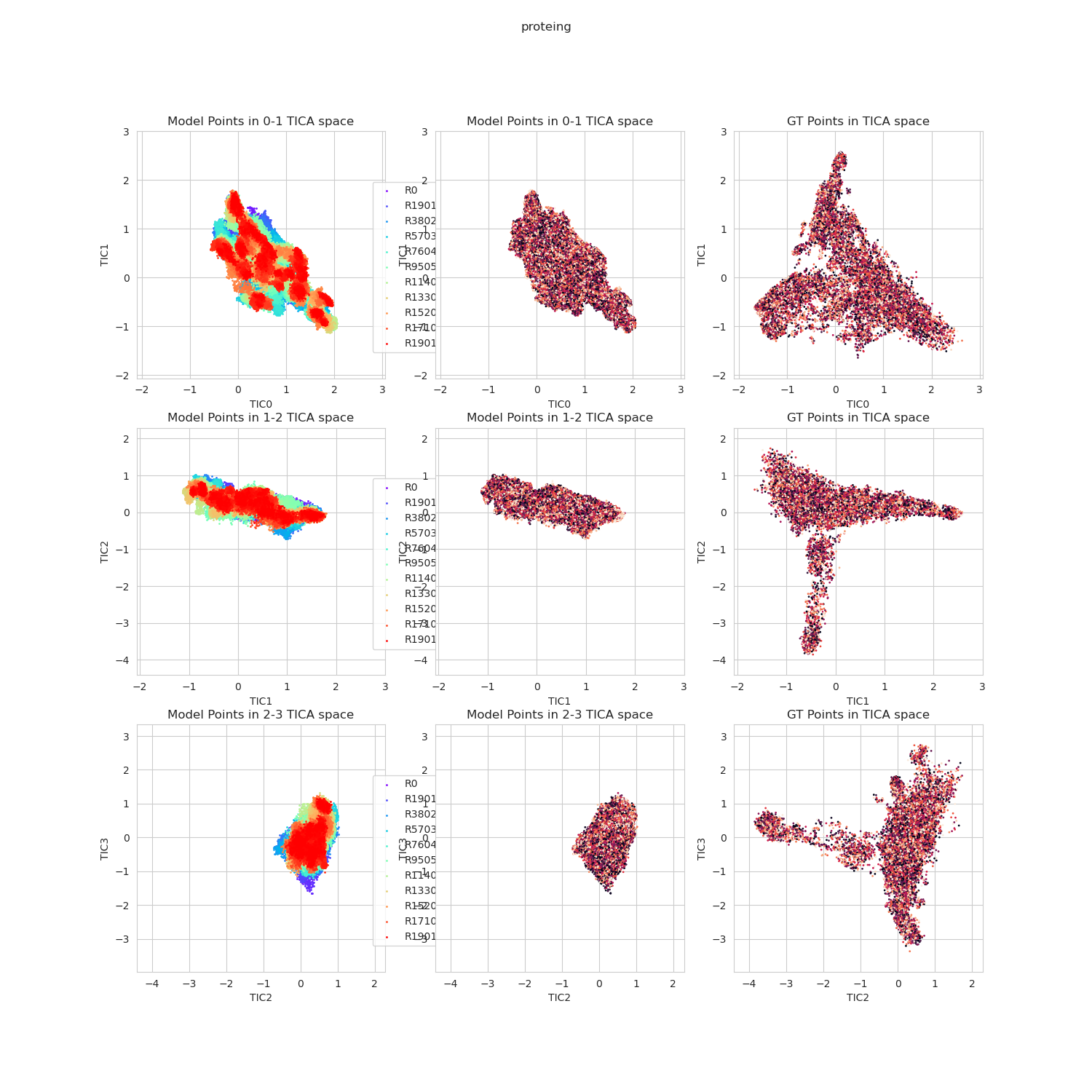}
      \caption{Complete set of ground truth points for Protein G in TICA 0/1 space with a stride of 100. The color of each point is the value of the 3rd TICA component.}
  \label{fgr:gt_wwdomain_full}
\end{figure}
\newpage

All MD simulations were performed using OpenMM \cite{Eastman2023OpenMM8} 8.2.0  with explicit solvent models. Initial protein structures were obtained from the Protein Data Bank (PDB) \cite{Burley2025_RCSBUpdatedResources} and processed using pdbfixer to repair missing residues, atoms, and termini, as well as to assign standard protonation states at pH 7.0. The systems were modeled using the AMBER14 all-atom force field (amber14-all.xml) in combination with the TIP3P-FB water model (amber14/tip3pfb.xml). Solvation was applied with 1.0 nm padding and 0.15 M NaCl ionic strength. Electrostatics were modeled with Particle Mesh Ewald (PME). All bonds involving hydrogen were constrained (HBonds). A Monte Carlo barostat maintained pressure at 1 atm with coupling every 25 steps. Temperature was held at 300K using Langevin Middle integration with a friction coefficient of 1 ps\(^{-1}\). Additional simulation parameters are listed in Table \ref{tab:gt_sim_params}.

\begin{table}[H]
\centering
\begin{tabular}{|lcc|}
\hline
\textbf{Parameter} & \textbf{Value} & \textbf{Unit} \\
\hline
Time step (\(\Delta t\)) & 4.0 & fs \\
Temperature (\(T\)) & 300 & K \\
Pressure (\(P\)) & 1.0 & atm \\
Friction coefficient (\(\gamma\)) & 1.0 & ps\(^{-1}\) \\
Nonbonded cutoff & 1.0 & nm \\
PME error tolerance & \(5 \times 10^{-4}\) & - \\
Hydrogen mass & 1.5 & amu \\
Constraint tolerance & \(10^{-6}\) & - \\
Barostat interval & 25 & steps \\
Equilibration time & 10 & ps \\
Solvent model & TIP3P-FB & - \\
\hline
\end{tabular}
\caption{Simulation parameters used for all explicit solvent OpenMM runs}
\label{tab:gt_sim_params}
\end{table}

Simulations were executed on the Delta supercomputer at NCSA. Each individual job utilized 1 compute node with 4 GPUs each. GPU bindings were configured to ensure efficient hardware allocation, and multithreading was enabled via OpenMP with 32 threads per task. Multiple simulation processes were run concurrently on each node to maximize throughput. To generate the trajectories, we used 25 nodes per protein (batched across starting points). Table \ref{tbl:gt_times} shows the total node hours consumed for each protein. In total, the simulations consumed approximately 2693.6 node-hours.

\begin{table}[h!]
\centering
\begin{tabular}{|c|c|}
\hline
\textbf{System} & \textbf{Compute Time \textsuperscript{\emph{1}}} \\
\hline
a3D & 15-15:03:09.568 \\
BBA & 08-19:09:16.537 \\
Chignolin & 02-19:01:34.386 \\
Homeodomain & 16-05:33:05.456 \\
$\lambda$-repressor & 11-20:19:47.807 \\
Protein B & 03-23:52:17.511 \\
Protein G & 25-22:22:37.020 \\
Trp-cage & 03-12:55:02.634 \\
WWdomain & 08-12:53:27.203 \\
\hline
\end{tabular}
\caption{Total node simulation time for ground truth proteins.}
\vspace{3ex}
\textsuperscript{\emph{1}} Compute time is in the form (days-hh:mm:ss) and represents total node time (not wall-clock time). A single node had 4 GPUs, each running in parallel 4 different MD simulations.
\label{tbl:gt_times}

\end{table}

\subsection{Benchmark Suite}
\label{sec:benchmark_suite}

To evaluate the accuracy of molecular simulation models, we provide a benchmark suite comprising 19 plots across five major classes of metrics. These metrics were carefully selected to capture a range of molecular behaviors, including long-timescale conformational dynamics, global structure, and local chemical geometry. Specifically, the benchmark suite includes \((1)\) TICA PDFs and 2D contour maps to examine slow collective motions, \((2)\) TICA point plots comparing model TICA points via WESTPA iterations, model TICA points via a macrostate model, and ground truth TICA points via a macrostate model, \((3)\) a Macrostate Model that determines the physical meaning of regions in TICA space,  \((4)\) Length, Angle, Dihedral, and Gyration-based metrics which assess local structural consistency and geometry, and \((5)\) a Contact Map illustrating residue-residue distances. These 5 classes of metrics comprise 19 visualizations in the unified benchmark suite, allowing users to thoroughly evaluate a model's temporal, structural, and geometric accuracy. Metrics were generated with the assistance of the deeptime library \cite{hoffmann2021deeptime}. By comparing the generated trajectories to ground truth simulations, the benchmark provides both visual and quantitative tools for identifying where a model succeeds or fails to reproduce key physical and structural features.

An important technical consideration throughout this benchmark is the use of WESTPA weights. WESTPA accelerates rare event sampling by resampling trajectory branches in a way that does not follow the natural Boltzmann statistics. To correct for this, each frame is assigned a statistical weight that reflects its contribution to equilibrium properties. We incorporate these weights when computing kernel density estimates (KDEs), histograms, averages in the TICA space, contact maps, radius of gyration, and bond, angle, and dihedral distributions. Without applying these weights, the resulting distributions would be heavily skewed toward over-sampled regions of phase space, particularly in transition pathways. The ground truth data, generated without WESTPA, does not use weights but instead relies on raw density and values for benchmarking.

\subsubsection{Time-lagged Independent Component Analysis}
A major component of the benchmark involves using TICA to extract and visualize the slowest collective motions in the system. This method is a dimensionality reduction technique that transforms the physical dimensions that make up a conformation, into an ordered orthogonal basis. Each basis vector is some combination of the original dimensions such that after the transform, such that after the resulting basis represent the modes of motion in the order of slowest to fastest. 

First, we calculate a TICA model on the ground truth reference data for each protein. We apply it as a measurement to both the reference, as  well as the model generated data.  Mathematically this can be defined as follows:

\[
z^\top(t) = r^\top(t)U = r^\top(t)W\Sigma^{-1}V
\]

with \(r(t)\) being the conformation pre-transformation, \(U = W\Sigma^{-1}V\) being the learned transform, and \(z(t)\) being the transformed data. Once we have the learned projection \(U\), we then project each trajectory's frames into the top four TICA components (0, 1, 2 and 3). From these, we generate three sets of 2D scatter plots: TICA components 0 vs 1, 1 vs 2, and 2 vs 3. These projections help reveal how well the model captures slow conformational changes. 

\begin{figure}
  \includegraphics[width=\textwidth]{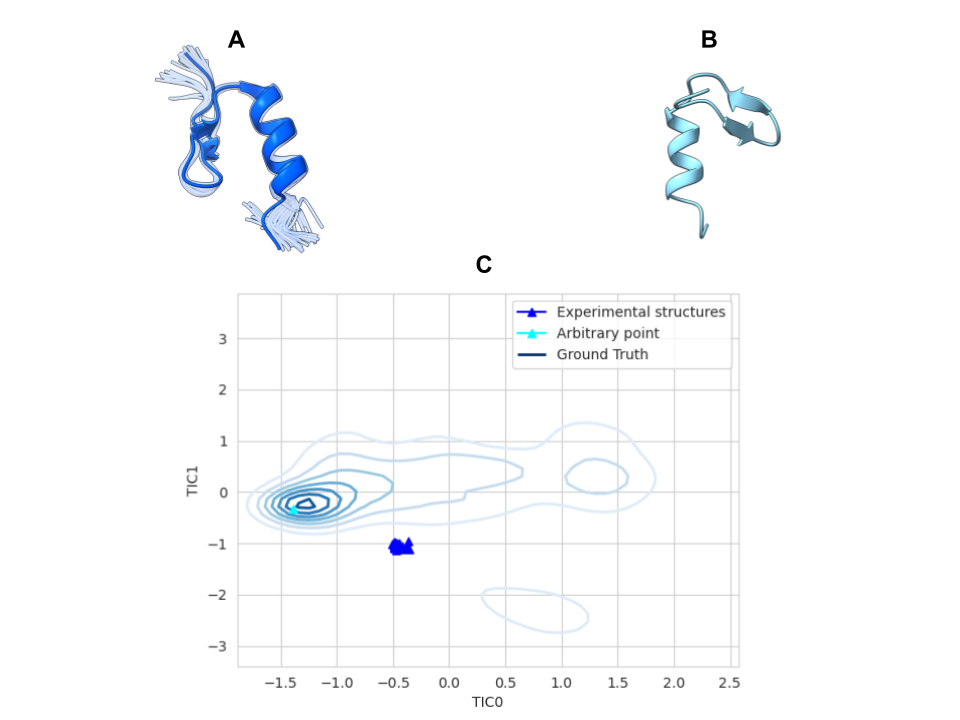}
      \caption{Conformations randomly chosen based on TICA proximity, visualized along side their location in TICA space. A: Many overlayed visualizations of close conformations with close TICA proximity. B: Visualization of randomly chosen conformation located far away in TICA space. C: Projection of all conformations visualized into TIC 0, and TIC 1 space.}
  \label{fgr:tica_distance}
\end{figure}

These slow moving modes of motion also serve as a good proxy for the probability density of the conformation space. Fig. \ref{fgr:tica_distance} demonstrates some randomly picked samples with high proximity in TICA space, and one sample farther away. The similarity in conformation aligning with the TICA distance gives a good example of TICA working as a measurable representation of conformation space.

To go beyond point-based comparisons, we also compute a weighted KDE of the distribution in the TICA 0/1 subspace for both ground truth and model trajectories. The KDE is calculated using the WESTPA-assigned weights \(w_i\) for each frame, following the formulation:

\[
p(x) = \sum_i w_i K_h(x - x_i)
\]

where \(K_h\) is a smoothing kernel and \(x_i\) is the truncated (up to 2 or 4) vector of the TICA principal components for frame \(i\). Additionally, we include histograms of the marginal distributions of TICA components 0 through 3. 

\subsubsection{Kullback-Leibler divergence and Wasserstein distance}
For all TICA-based metrics, we compute two statistical measures of distributional divergence: the Kullback-Leibler (KL) divergence and the Wasserstein-L1 (W1) distance. The KL divergence quantifies the relative entropy between two distributions \(p(x)\) and \(q(x)\):

\[
D_{\text{KL}}(P \parallel Q) = \int p(x) \log\left( \frac{p(x)}{q(x)} \right) dx
\]

The W1 distance measures the minimum ``transport cost" of transforming one distribution into another:

\[
W(P, Q) = \inf_{\gamma \in \Gamma(P, Q)} \int |x - y| \, d\gamma(x, y)
\]

where \(\gamma\) defines transport plan and \(\Gamma\) is the space of all transport plans. Together, these visual and statistical comparisons highlight whether a model samples the correct regions of conformational space.

\subsubsection{Contact map difference}
Contact maps serve as another important global structural metric in our benchmark. Rather than a thresholded binary contact, we compute the difference in mean distances (in \AA) between residue pairs using C$_\alpha$-C$_\alpha$ distances. For each residue pair $(i,j)$,
\[
C_{ij} = \big\langle d_{ij} \big\rangle_M - \big\langle d_{ij} \big\rangle_{\mathrm{GT}}
\]
where $d_{ij}$ is the C$_\alpha$ distance and the model average $\langle \cdot \rangle_{M}$ is computed with WESTPA weights. Positive values indicate the model samples larger average separations than ground truth; negative values indicate separations are closer than the ground truth. Deviations between model and GT contact maps can reveal structural misfolding, loss of compactness, or overrepresentation of non-native interactions.

\subsubsection{Radius of gyration}
To assess the global shape and compaction of the molecule, we compute the radius of gyration \(R_g\), defined as:

\[
R_g = \sqrt{\frac{1}{N} \sum_{i=1}^N \|\mathbf{r}_i - \mathbf{r}_{\text{com}}\|^2}
\]

where \(\mathbf{r}_{\text{com}}\) is the center of mass of the structure. A histogram of \(R_g\) values across the trajectory shows whether the model appropriately samples between extended and compact states. Like the TICA and contact map metrics, the \(R_g\) distributions are weighted using WESTPA-derived weights to ensure accurate representation of equilibrium statistics.

\subsubsection{Bond lengths, angles and dihedrals}
Local structural fidelity is evaluated through the distributions of bond lengths, bond angles, and dihedral angles, (BAD) computed for all relevant residual triplets and quadruplets throughout the trajectories. The BAD coordinates of a protein represent the full structure up to rotations. Bond lengths are directly measured as:

\[
d_{ij} = \|\mathbf{r}_i - \mathbf{r}_j\|
\]

while bond angles and dihedrals are calculated using cosine and cross-product relationships among bonded atoms. Histograms of these quantities are constructed for both ground truth and model trajectories, using WESTPA weights to account for sampling bias. These histograms are particularly diagnostic: for example, if the bond length histogram has significant probability density below 3.5 Å or above 4.5 Å, the model likely violates basic chemical constraints. 

\subsubsection{Computing steady-state dynamics through Markov State Models}
In order to compute an unbiased estimate of the dynamics from the biased, weighted data generated with WESTPA, we run a Markov State Model \cite{husic2018markov} (MSM) constructed by representing the system's dynamics as a Markov chain over a discretized state space. The TICA space is partitioned into \(n\) discrete states \(S = \{s_1, s_2, \dots, s_n\}\) using an \(n\times n\) (We selected \(n=80\) arbitrarily) rectilinear binning scheme. Transitions between these states are counted at a fixed lag time \(\tau\) to build a transition count matrix \(C \in \mathbb{R}^{n \times n}\), where \(C_{ij}\) is the number of observed transitions from state \(i\) to state \(j\). This matrix is row-normalized to yield the transition probability matrix \(T\), with \(T_{ij} = \frac{C_{ij}}{\sum_k C_{ik}}\) and \(\sum_j T_{ij} = 1\), which approximates the system's propagator over lag time \(\tau\) under the Markovian assumption that future states depend only on the current state. From \(T\), the stationary distribution (if unique) \(\pi \in \mathbb{R}^n\) is obtained as the normalized left eigenvector corresponding to eigenvalue 1, satisfying \(\pi^T T = \pi^T\) and \(\sum_i \pi_i = 1\). These MSM-based histograms serve as an optional, debiased reference and provide an alternative to weighted estimates. Users may choose between using WESTPA weights, MSM corrections, or raw unweighted counts.

\begin{figure}
    \centering
    \includegraphics[width=\linewidth]{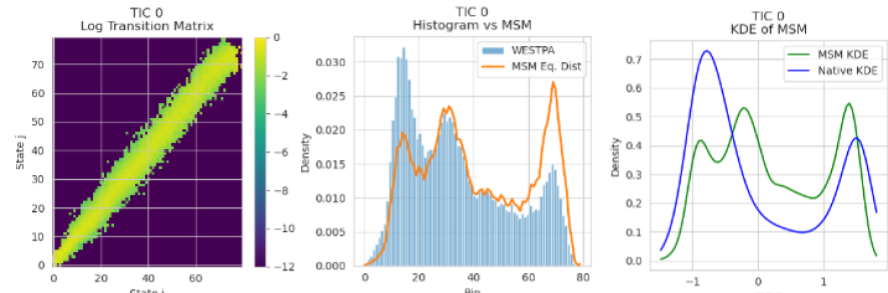}
    \caption{Construction of the MSM starts with (left) the Log probability of transition matrix between the discrete bins, then (middle) the equilibrium distribution using the MSM vs the raw quantities from the WESTPA run, and finally (right) constructing the two KDEs (Green is the SSMSM, Blue is the Raw Data). }
    \label{fgr:ssmsm}
\end{figure}
\clearpage
\subsubsection{Macrostates}
We utilize TICA space to construct the macrostates of a protein, which are the states with significant alterations of the conformation of the protein. These macrostates represent different stable states of the protein, such as being unfolded vs folded. The macrostates are calculated by first running K-means with 100 clusters using the Euclidean distance on the first ten TICA components. Then a Markov State Model is fitted to the transition probabilities between clusters. We then assign clusters to five macrostates with PCCA+ \cite{Rblitz2013}, though the number of clusters can be edited as desired. We utilize the graphic of the 2D TICA point graphs and color each point to identify which macrostate they are apart of. 

Taken together, these metrics and graph form a comprehensive suite for benchmarking MD and AI-based simulations. By capturing a spectrum of global, local and geometric features, the framework enables robust and reproducible comparisons across traditional and machine-learned models. 

\subsection{Propagator, Progress Coordinates, and Modular Design}
\label{sec:propagator}

\subsubsection{WESTPA-based Weighted Ensemble (WE)}
Efficiently sampling proteins' conformational spaces with classical all-atom MD or even with coarse-grained models is computationally prohibitive. In order to accelerate sampling of rare events such as transitions between metastable states, we have leveraged WESTPA-2.0, the weighted ensemble enhanced sampling framework, which we integrated into our benchmark. We have developed a propagator module that enables coupling with a variety of propagators for molecular simulation. We implemented a custom propagator for CGSchNet, supporting coarse-grained, machine learning-driven molecular dynamics, as well as for implicit and explicit OpenMM simulations. 

\subsubsection{Propagators}
Each propagator is designed to be flexible in how simulation data is saved. Users can specify whether to store trajectory data in DCD format (widely supported by visualization tools), compressed NumPy .npz format (ideal for lightweight post-processing and machine learning workflows), or both, depending on downstream analysis needs. The modular architecture also facilitates rapid development of new propagators tailored to other simulation engines or specialized workflows. Developers can implement custom propagators by extending a straightforward interface that cleanly separates the concerns of trajectory propagation, data collection, and integration with WESTPA's adaptive sampling loop. 

Each propagator is responsible for executing the MD trajectory segments and managing all associated data collection. Following each simulation segment, WESTPA's adaptive resampling algorithm is automatically triggered based on a user-defined progress coordinate. 

\subsubsection{TICA-based progress coordinates}
To support enhanced exploration of conformational landscapes, we developed a custom progress coordinate using TICA principal components. Unlike traditional progress coordinates such as root-mean-square deviation (RMSD), TICA-based coordinates are particularly well-suited for capturing the slow collective motions intrinsic to protein folding, binding, and conformational transitions. These slow modes are often the key drivers of rare-event dynamics. Our implementation allows for an arbitrary number of TICA components (TICs) to be used in defining the progress coordinate; however, in practice, we typically only utilize the first two TICs. These often account for the majority of the slow motion variance in the system, while also keeping the dimensionality of the binning scheme computationally tractable. Since the cost of binning increases exponentially with increased dimensionality, this trade-off ensures a balance between resolution and efficiency.

For binning, we employ the Minimal Adaptive Binning (MAB) scheme \cite{torrillo2021}, a dynamic binning strategy designed to facilitate efficient sampling over high free-energy barriers. MAB adaptively adjusts bin placement based on observed sampling density, obviating the need for apriori specification of bin boundaries.

\subsection{Experiments Compute and Setup}
\subsubsection{Implicit Solvent Experiments}
To validate our benchmarking methodology, we performed implicit solvent all-atom MD simulations. Starting from a single conformation, we demonstrate comprehensive exploration of conformational space, confirming that the TICA space is fully connected and traversable. This indicates that regions of the space are not isolated partitions; rather, any two points can be connected given sufficient time, making it reasonable to expect a protein to explore the entire space from a single starting structure. Utilizing implicit all-atom MD as the propagator also provides a substantially higher degree of physical fidelity compared to current machine learning based models. Consequently, the resulting benchmark is expected to more accurately capture the underlying physical behavior of the system, thereby offering a more reliable standard of comparison.

Implicit solvent simulations with WESTPA were executed on the NCSA supercomputer (Delta) and the NERSC supercomputer (Perlmutter). Each individual job utilized 8-16 compute nodes with 4 GPUs per node. GPU bindings were configured to ensure efficient hardware allocation, and multithreading was enabled via OpenMP with 16 threads per task. Each node ran 4 tasks concurrently, each bound to a single GPU. Wall times ranged from 8 to 12 hours, depending on the specific configuration. In total, these simulations consumed approximately 8,348 node hours.

For the WESTPA configuration, we used of a two-dimensional progress coordinate (TIC 0 and TIC 1), recorded at 11 time points per segment. MAB binning was applied using 7 bins in each TICA dimension with each bin targeted to maintain 3 walkers.

These MD simulations were carried out using OpenMM 8.2.0 with implicit solvent enabled via the GBN2 model (implicit/gbn2.xml), paired with the AMBER14 all-atom force field (amber14-all.xml). Initial structures were derived from preprocessed PDB files with appropriate topological repairs. The simulations were performed using mixed GPU precision. A 4 fs time step was used with Langevin Middle integration at 300 K and a friction coefficient of 1 ps\(^{-1}\). Each iteration is run for 1000 steps with data saved every 100 steps. 

\subsubsection{Model Experiments}
CGSchNet simulations with WESTPA were executed on the NERSC supercomputer (Perlmutter). Each individual job utilized 2 compute nodes with 4 GPUs per node. GPU bindings were configured to ensure efficient hardware allocation, and multithreading was enabled via OpenMP with 64 threads per task. Each node ran 1 task, bound to four GPUs. Wall times were typically 2 hours per job.

Similarly to the implicit solvent data, we used TIC 0 and TIC 1 as a 2D progress coordinate. MAB binning was applied using 7 bins in each TICA dimension and 3 walkers per bin. Simulation runs were constrained to a maximum of 200 total WESTPA iterations.

To demonstrate the effectiveness of our benchmark, we trained two models based on the CGSchNet architecture from Husic et al. The first model was intentionally under-trained using only 10\% of the frames and data from the entire protein dataset, designed to perform poorly. The second model was also trained on the full dataset, incorporating all available frames and all proteins. Full details of shared parameters and differences can be found in Table \ref{tab:majewski-shared} and Table \ref{tab:majewski-diffs}, respectively.
 
\begin{table}[H]
\centering
\small
\caption{Shared parameters and training setup between fully trained and under-trained CGSchNet models.}
\begin{tabularx}{\textwidth}{|L M Y|}
\hline
\textbf{Hyperparameter} & \textbf{Name / Flag} & \textbf{Value} \\
\hline
Model type & \texttt{model} & graph-network \\
Output head & \texttt{output\_model} & Scalar \\
Precision & -- & FP32 (\texttt{precision: 32}) \\
Embedding dimension & \texttt{embedding\_dimension} & \textbf{128} \\
Number of interaction layers & \texttt{num\_layers} & \textbf{4} \\
Number of RBF & \texttt{num\_rbf} & \textbf{20} \\
RBF type & \texttt{rbf\_type} & \textbf{expnorm} \\
Trainable RBF & \texttt{trainable\_rbf} & \textbf{true} \\
Activation (message/attn) & \texttt{activation}, \texttt{attn\_activation} & \textbf{silu} \\
Equivariance group & \texttt{equivariance\_invariance} & \textbf{O(3)} \\
Attention heads & \texttt{num\_heads} & \textbf{8} \\
Aggregation & \texttt{aggr} & add \\
Neighbor embedding & \texttt{neighbor\_embedding} & false \\
Distance influence & \texttt{distance\_influence} & both \\
Max neighbors & \texttt{max\_num\_neighbors} & 32 \\
Cutoff lower / upper (\AA) & \texttt{cutoff\_lower}, \texttt{cutoff\_upper} & \textbf{2.0} / \textbf{12.0} \\
Derivatives (forces) & \texttt{derivative} & true \\
Charge / Spin & \texttt{charge}, \texttt{spin} & false / false \\
Standardize & \texttt{standardize} & false \\
Initial learning rate & \texttt{--lr} & 1e-4 \\
Weight decay & \texttt{--wd} & 0.0 \\
Early stopping patience & \texttt{--early-stopping} & 1 \\
Train/Val split & \texttt{--val-ratio} & 0.10 (90\% / 10\%) \\
Proteins used & \texttt{--proteins} & All Benchmark Proteins \\
Micro-batching effect & Split batch into APC sub-batches & Split batch into APC sub-batches \\
\hline
\end{tabularx}
\label{tab:majewski-shared}
\end{table}

\begin{table}[H] %
\centering
\small
\caption{Key differences between fully trained and under-trained CGSchNet models.}
\begin{tabularx}{\textwidth}{|L M Y|}
\hline 
\textbf{Aspect} & \textbf{Fully-trained model} & \textbf{Under-trained model} \\
\hline 
LR schedule & --plateau-lr & Constant \\

Frame selection & \textbf{All frames} per protein & \textbf{10\% of frames} per protein (random) \\
APC (\textit{atoms-per-call}) & \textbf{20000} (\texttt{--apc 20000}) & --apc 60000 \\
\hline
\end{tabularx}
\label{tab:majewski-diffs}
\end{table}

\section{Results}

\subsection{Benchmark Results}

\begin{table}[ht]
\centering
\begin{tabular}{|lccc|}
\hline
\textbf{Protein} & \textbf{\% TICA Explored \textsuperscript{\emph{1}}} & \textbf{Compute Time \textsuperscript{\emph{2}}} & \textbf{Max Timescale (ns) \textsuperscript{\emph{3}}}\\
\hline
a3D            & 51.24\% & 49-13:28:50 & 51.644 \\
BBA            & 90.89\% & 28-10:12:53 & 29.300 \\
Chignolin      & 93.05\% & 09-03:45:33 & 21.928\\
Homeodomain    & 82.72\% & 83-11:43:32 & 69.604\\
\(\lambda\)-repressor  & 68.15\% & 74-22:28:24 & 74.184 \\
Protein B      & 75.12\% & 78-03:29:37 & 46.508\\
Protein G      & 73.47\% & 65-22:24:44 & 71.696\\
Trp-cage       & 96.44\% & 13-16:51:52 & 23.924\\
WW Domain      & 83.26\% & 42-11:47:23 & 32.808\\
\hline
\end{tabular}
\caption{Protein Conformational Space Exploration using Weighted Ensembles}
\label{tab:protein_metrics_transposed}
\vspace{3ex}
\textsuperscript{\emph{1}} Metrics are from implicit solvent all-atom MD simulations. Coverage (\% Explored) is computed by discretizing the 2D TICA-based conformational space into a 100\(\times\)100 grid; a grid cell is counted as explored if it contains any ground truth point and at least one model point.

\textsuperscript{\emph{2}} Compute time is in the form (days-hh:mm:ss) and represents total node time (not wall-clock time). A single node had 4 GPUs, each running in parallel 4 different MD simulations.

\textsuperscript{\emph{3}} Maximum timescale is a measure of the how much time has passed from the starting point at the final iteration. At 4 fs per step and 1,000 steps per iteration, each iteration corresponds to 4 picoseconds (0.004 ns). The maximum timescale is found by taking the total iterations for a protein and multiplying by 4 ps. Note that this is not a metric of compute required, since there maybe hundreds of segments per iteration that are parallel trajectories in time. 
\end{table}

Table \ref{tab:protein_metrics_transposed} shows the percentage of explored ground-truth 2D TICA space using all-atom, implicit solvent MD simulations propagated using WESTPA. The reference data against which this was compared is the ground truth data described in the methods. Some proteins such as Chignolin and Trp-cage explored more than 93\% of the conformational space, while taking the equivalent of 9 to 13 days on a single node (note that we ran these in parallel on multiple nodes so the actual time was significantly reduced). On the other side, a protein like a3D only explored 51\% after the equivalent of 49 days of single-node hours. 

For the stopping criteria, we attempted to stop each run after reaching over 75\% exploration, a threshold met by most systems. Although certain proteins, such as a3D, exhibited lower effective sampling relative to others of comparable size, their maximum timescales were similar or greater, indicating an equivalent or higher number of iterations. Notably, qualitative analysis revealed that larger proteins, including $\lambda$-repressor, Protein G, and particularly a3D, tended to plateau in their exploration. This behavior is likely attributable to the presence of complex conformational landscapes characterized by high energy barriers separating distinct metastable states.

It is important to emphasize that the computational times observed for the simulations are greater than those associated with the ground truth calculations. However, a direct comparison of compute time between the ground truth generation and the implicit all-atom weighted ensemble approach is not meaningful due to fundamental methodological differences. The ground truth is established through multiple independent simulations initiated from distinct starting configurations that collectively span the conformational landscape. Each trajectory is propagated for a fixed duration to ensure comprehensive sampling of the system’s conformational dynamics.

In contrast, the weighted ensemble methodology initiates from a single starting configuration, from which all subsequent conformational exploration emerges through a resampling strategy. Attempting to reproduce such global exploration from a single initial condition using conventional, unbiased OpenMM explicit all-atom molecular dynamics simulations, analogous to the ground truth protocol, would require orders of magnitude more computational effort than either the ensemble of parallel ground truth trajectories or the WE simulations.

The primary motivation for employing the weighted ensemble strategy is to prevent the model from becoming confined to local regions of configurational space and to facilitate efficient, systematic exploration of the entire conformational landscape from a single initial condition. While it is expected that, given sufficient time, the ground truth simulations would eventually sample the entire space, such guarantees cannot be extended to machine learning driven models. Consequently, using weighted ensembles serves as a necessary mechanism to verify model correctness and ensure adequate sampling diversity, rather than relying on parallelized trajectories.

Although comparisons of computational cost between the weighted ensemble method and the ground truth are not directly meaningful, it remains valid to assess the computational efficiency of different models who are using the weighted ensemble methodology.

\subsection{Full Benchmark on Chignolin}

\begin{figure}
    \centering
    \includegraphics[width=\linewidth]{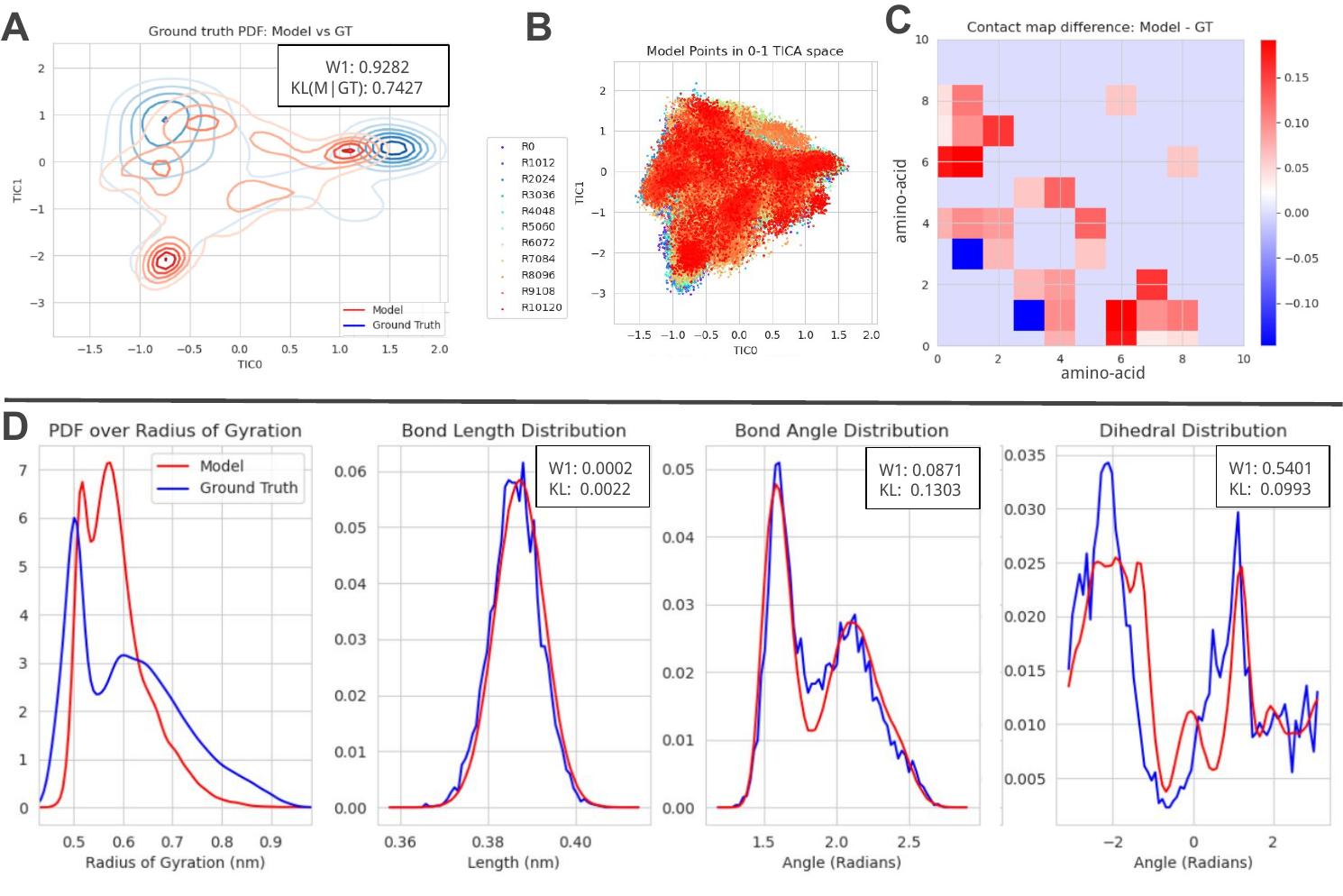}
    \caption{Main results of our benchmark evaluated on Chignolin. The ground truth (GT) data (in blue) is described in the methods section, the model (in red) is an implicit-solvent all-atom MD that was enhanced with a WESTPA-based weighted ensemble. (A) TICA contour plot between model vs ground-truth (B) model points in TICA space sampled from WESTPA, colored by WESTPA iteration. (C) Difference in contact maps between model and ground-truth (see \nameref{sec:benchmark_suite}). (D) Distributions for radius of gyration, bond lengths, 3-atom angles and 4-atom dihedral.  Note the W1 and KL values are added on the graphic for clarity, actual benchmark graphics do not include W1 and KL values which are instead found in a table.}
    \label{fgr:main_benchmark}
\end{figure}

In Fig. \ref{fgr:main_benchmark} we highlight the main capabilities of our benchmark on Chignolin, by comparing the explicit-solvent MD ground truth data from the methods section vs an all-atom implicit-solvent MD simulation enhanced with a WESTPA-based weighted ensemble (i.e. ``the model``). The 2D TICA-based density plots (Fig. \ref{fgr:main_benchmark}A) show significant overlap between the model and the GT data, albeit there are slight shifts in the peak locations and amplitudes. Fig. \ref{fgr:main_benchmark}B shows a scatter plot of the model points which have been generated from multiple WESTPA-based segments (597,210 in total across 5482 iterations). In Fig. \ref{fgr:main_benchmark}C we show the contact map difference, showing which amino-acid pairs in the protein are further away from one another in the model simulations as opposed to ground truth data (red squares) and which ones are closer together (blue squares). Results show that the model generally places some pairs of amino-acids closer together as opposed to ground truth data (light blue), with only pair (3,1) being placed consistently closer together than in the ground-truth data (dark blue). The four plots in Fig. \ref{fgr:main_benchmark}D (bottom row) show the distributions of four key metrics: the radius of gyration, bond lengths, 3-atom angles and 4-atom dihedrals. While the distributions for the radius of gyration and dihedrals show some difference between the all-atom implicit solvent model vs ground truth, the bond length and bond angle distributions match the ground truth very well. All other complete protein benchmarks can be found in the Supplementary Information. 
 
\subsection{Comparison Between Under-Trained and Fully-Trained ML Models}
\begin{figure}
    \includegraphics[trim={0 0 0 1cm},clip,scale=0.62]{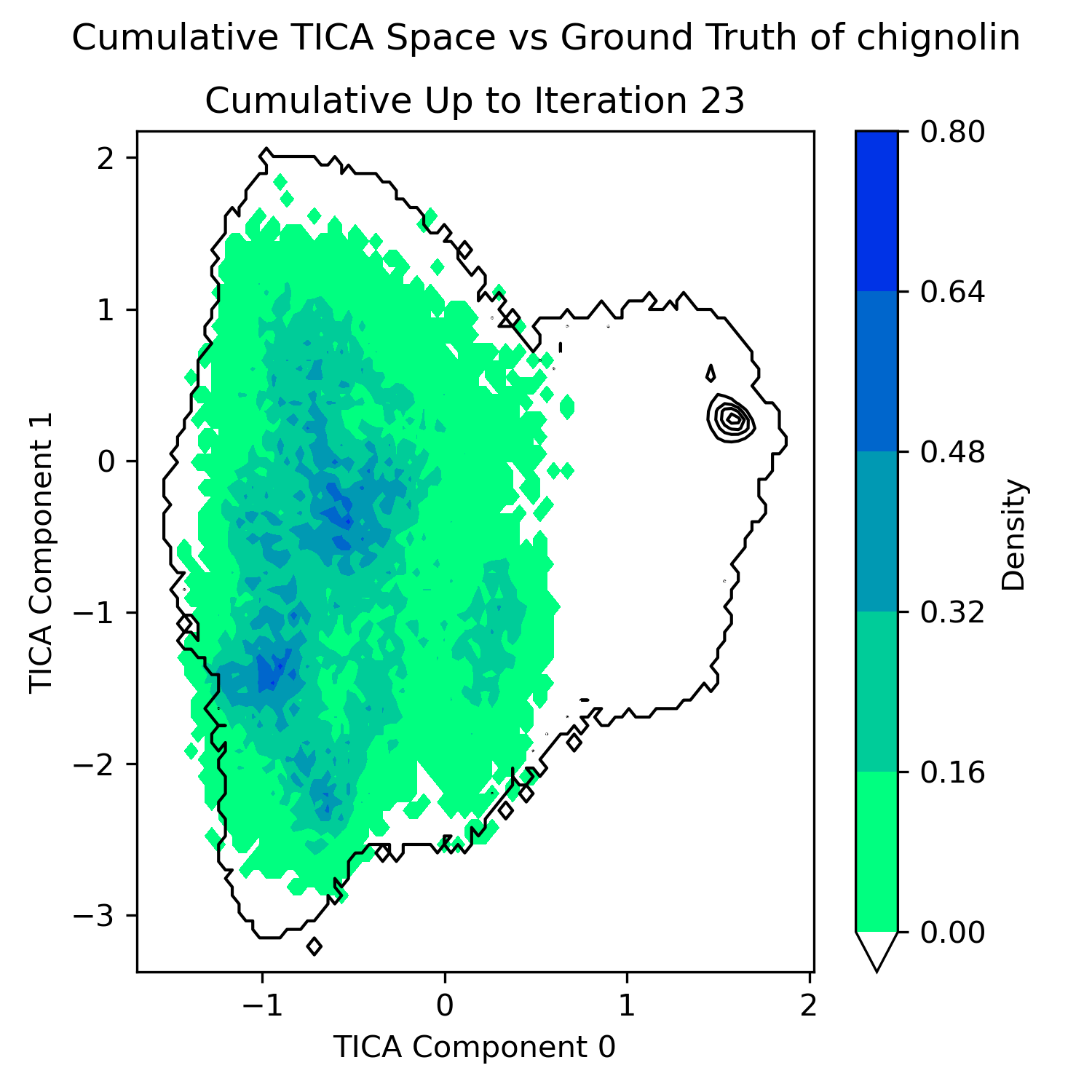}
    \includegraphics[trim={0 0 0 2cm},clip,scale=0.2]{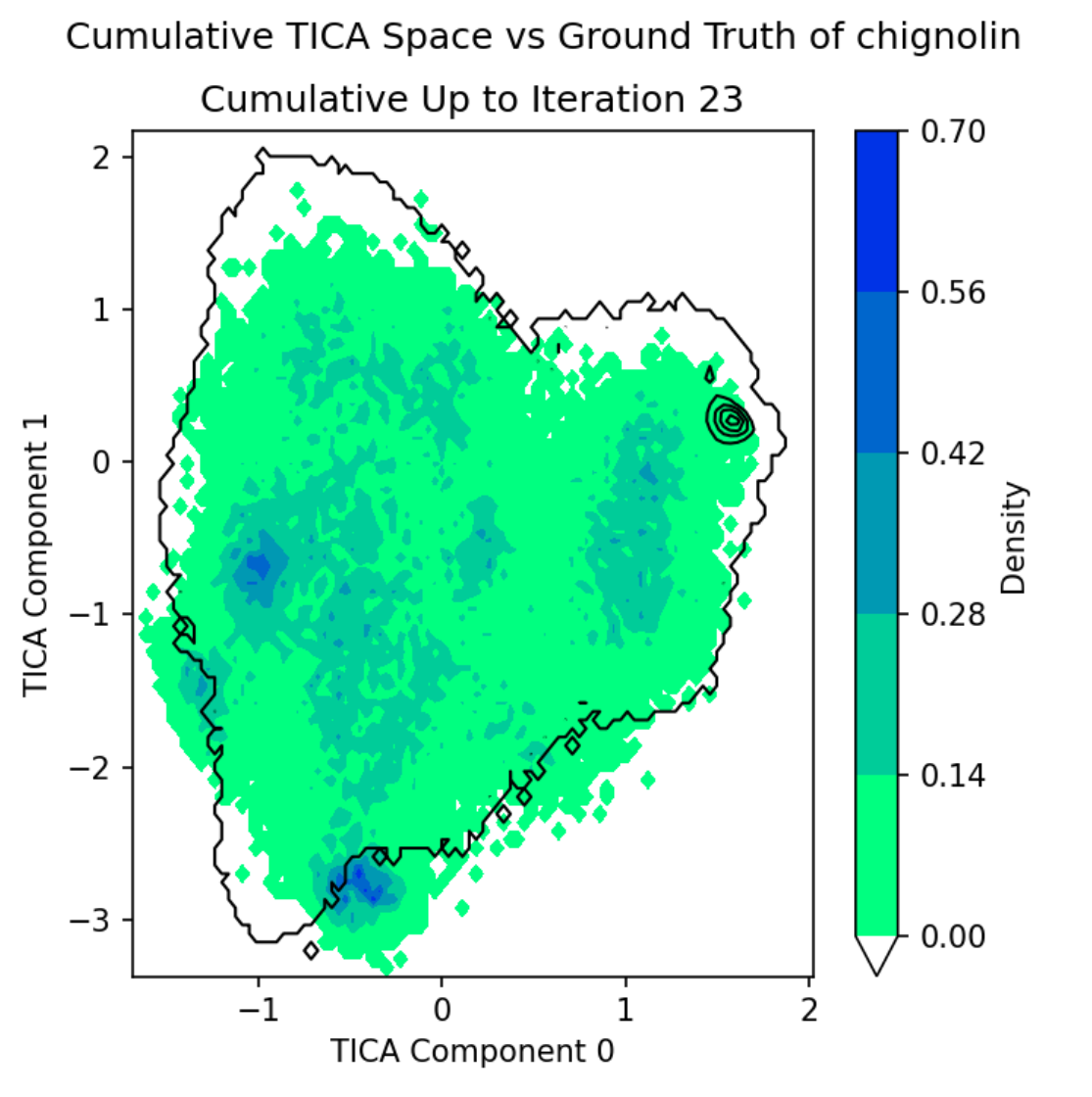}
      \caption{Sampling of the Chignolin conformational space with (left) an under-trained CGSchNet model vs (right) a fully-trained model. The black contour is the ground truth region, and the color represents the raw density of conformational states.}
  \label{fgr:good_model_explored}
\end{figure}

In Fig. \ref{fgr:good_model_explored} we show the conformational space explored by a CGSchNet model trained with 10\% of data (under-trained) versus a model trained on the entire dataset (fully-trained). Both models ran for 23 WESTPA iterations on 120 segments with 1,000 CG steps per segment at each iteration. The conformational space explored by the trained CGSchNet model is broader and more physically meaningful compared to that of the under-trained model. This indicates that a fully trained model can accurately capture the dynamics of the protein, whereas an under-trained model fails to do so.

\begin{figure}
    \includegraphics[clip,scale=0.115]{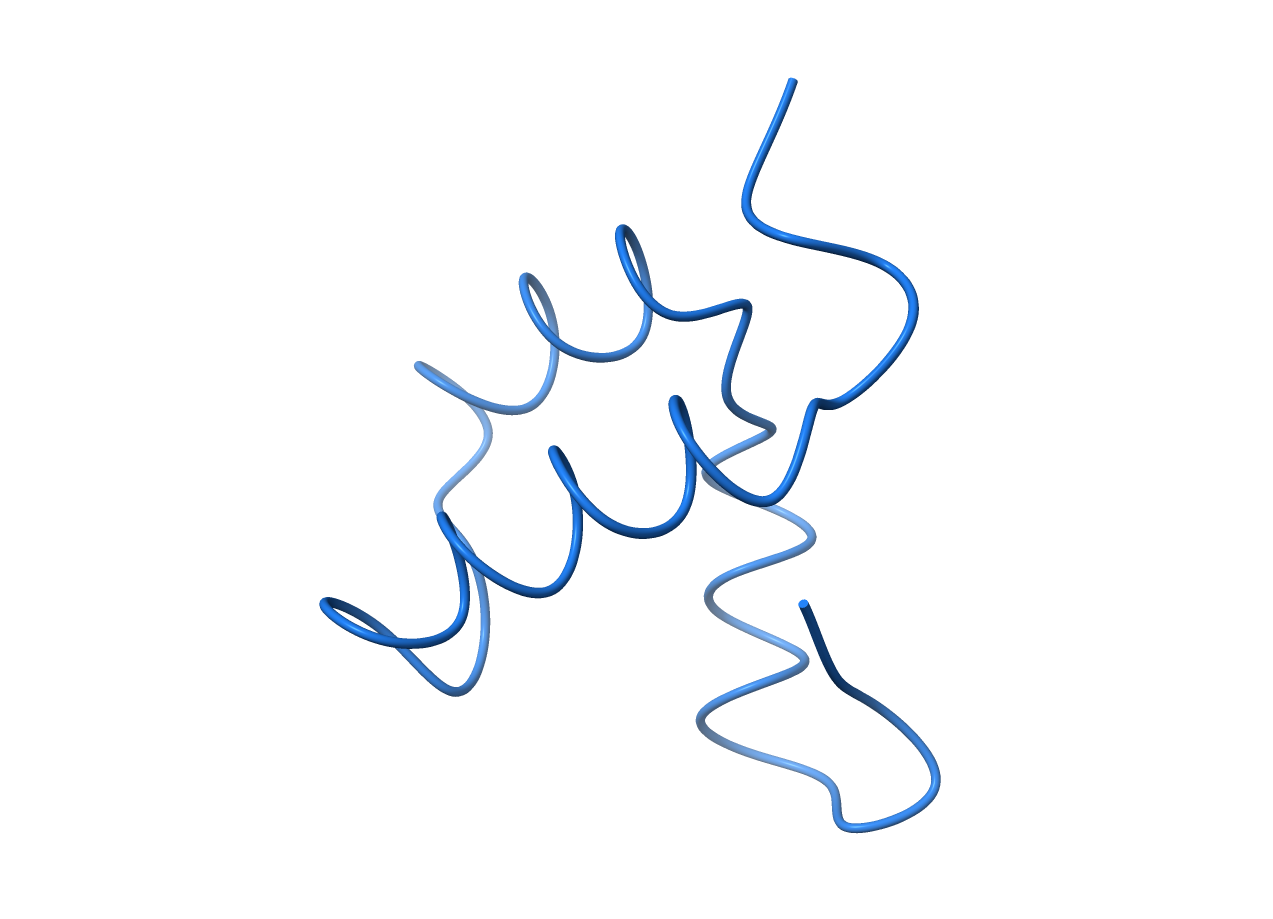}
    \includegraphics[clip,scale=0.115]{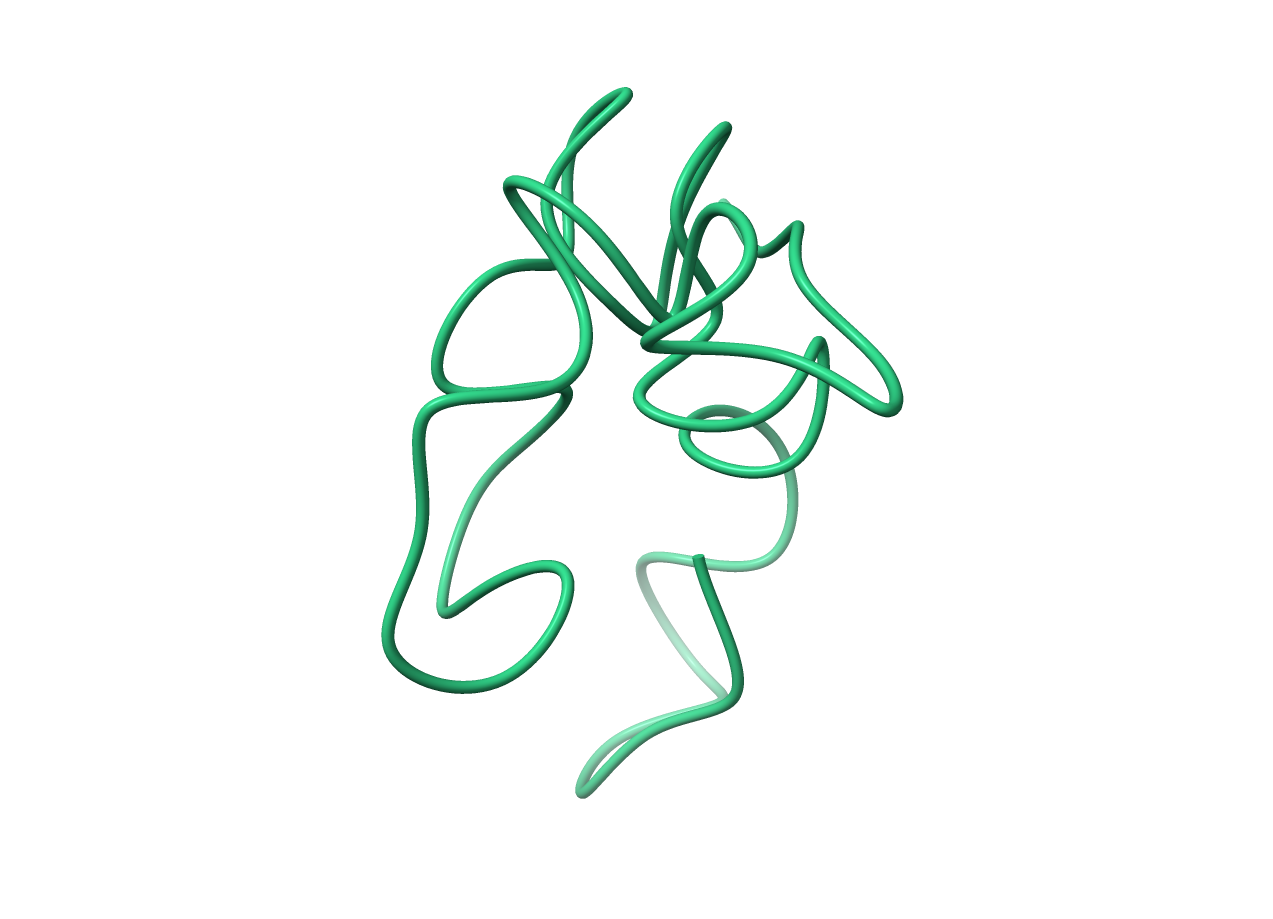}
    \includegraphics[clip,scale=0.115]{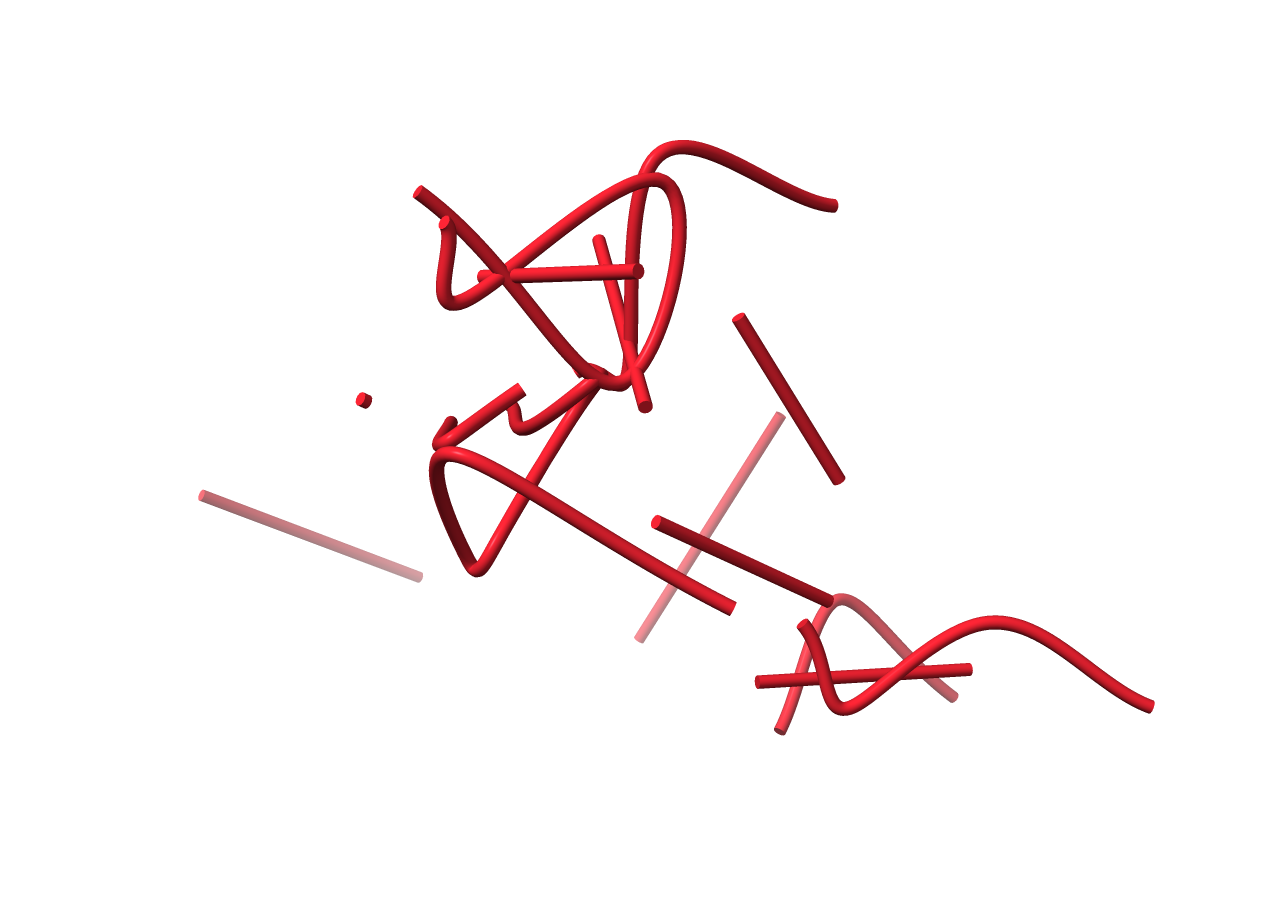}
    \caption{Model-induced instabilities lead to broken proteins. A stable structure of Homeodomain (Left, Blue) is the starting position of the protein. A bad model may implode (Middle, Green) or explode (Right, Red) from the starting position due to being poorly trained. }
  \label{fgr:homedomain_stability}
\end{figure}

\begin{figure}
    \includegraphics[trim={5cm 19in 5cm 2cm},clip,scale=0.3]{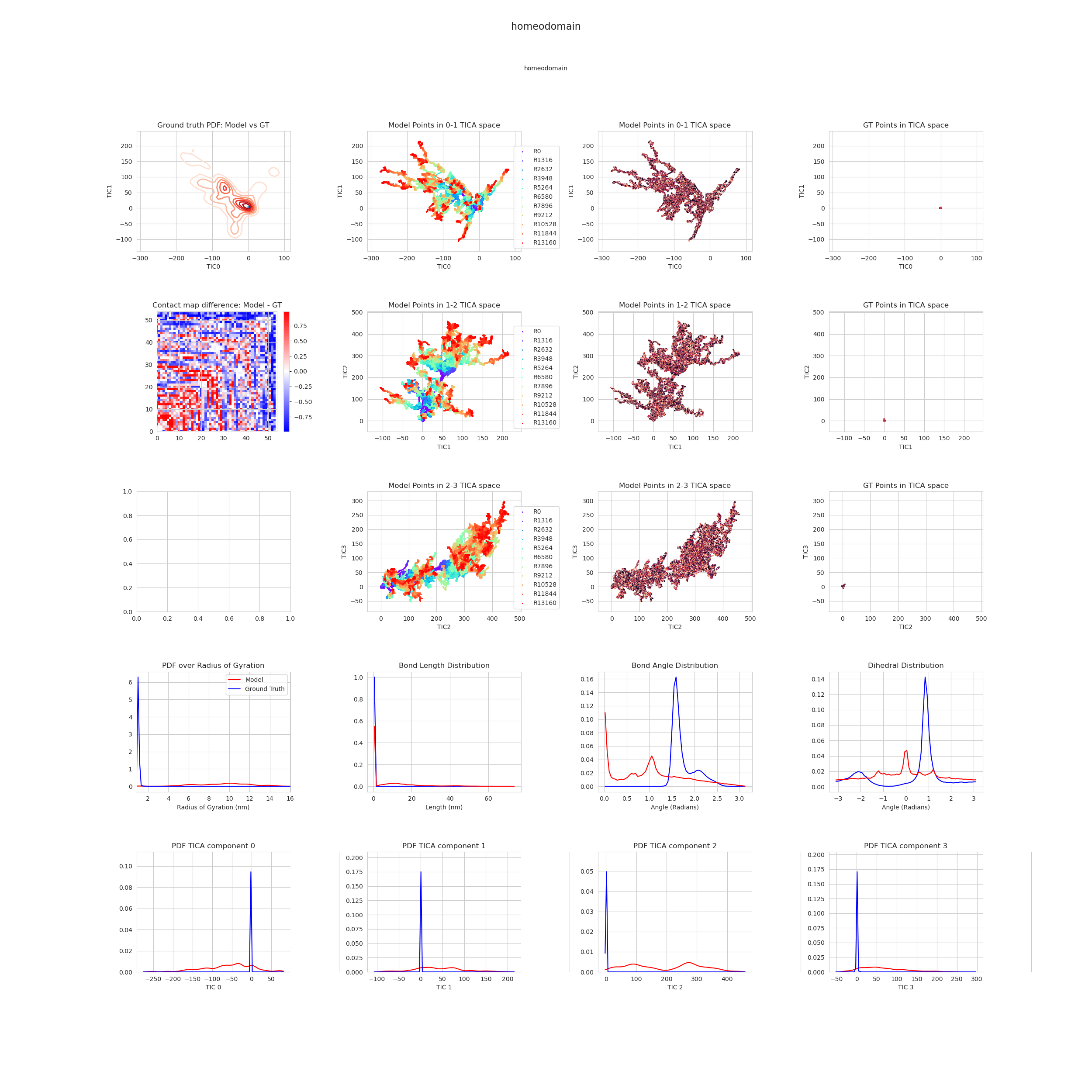}
    \includegraphics[trim={5cm 6in 5cm 15in},clip,scale=0.3]{images/bad_model_benchmark/bad_homeodomain_benchmark.png}
    \caption{Partial benchmark graphics for an exploded Homeodomain. The graphics clearly show poor performance of the model with nonphysical values compared to the ground truth.}
  \label{fgr:homedomain_exploded_benchmark}
\end{figure}

In many cases, under-trained models produce unstable protein trajectories, with some proteins exploding or imploding due to model-induced instabilities as shown in Fig. \ref{fgr:homedomain_stability}. If the bond lengths or angles reach physically implausible values, then the model is often unable to continue running as the GNN may create bonds that exceed a maximum threshold (used to determine when it implodes). In such cases the benchmark (Fig. \ref{fgr:homedomain_exploded_benchmark}) may appear extreme suggesting such an explosion having occurred. \\

\begin{figure}
    \includegraphics[trim={2.5cm 0 2.5cm 5.13in},clip,scale=0.45]{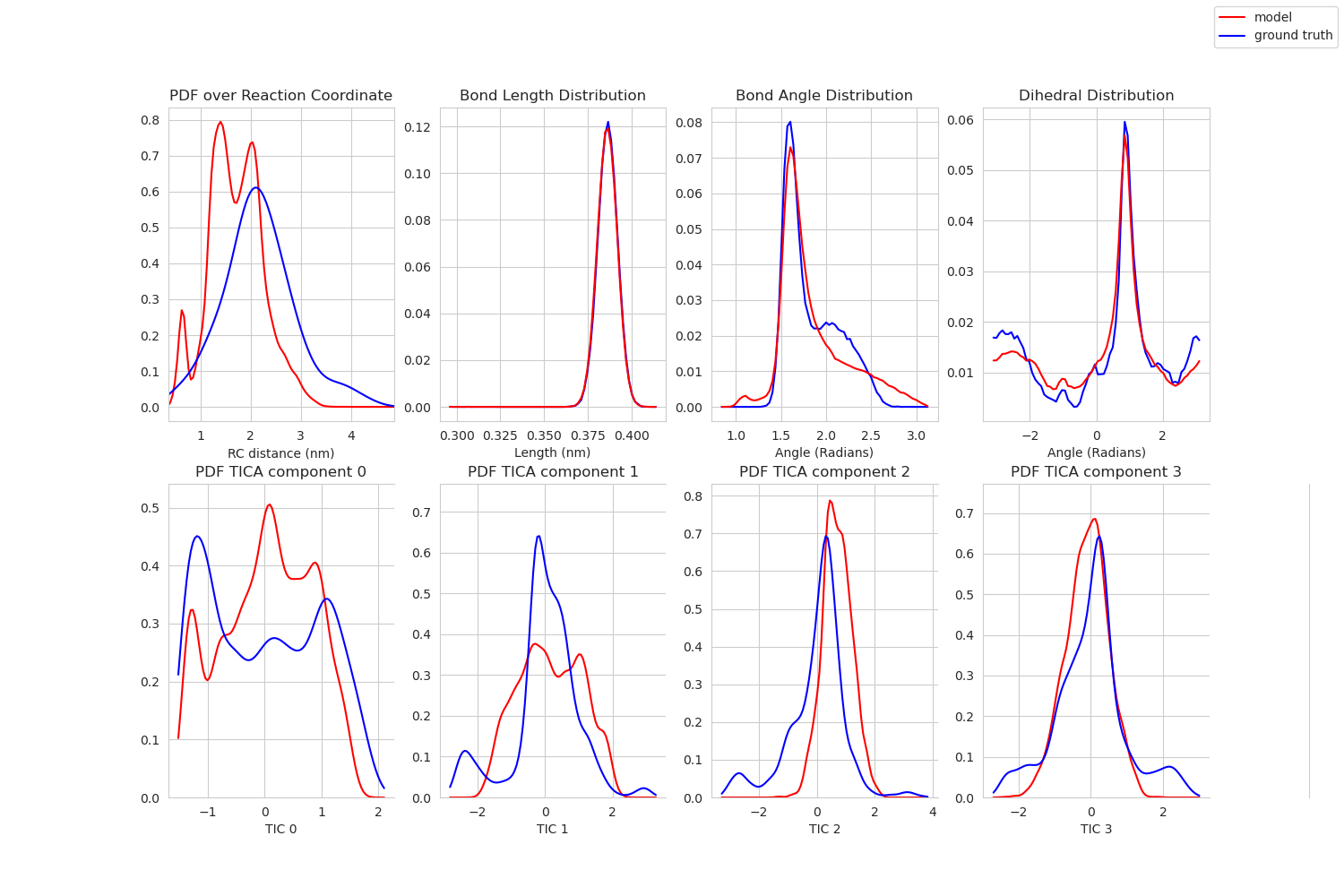}
    \includegraphics[trim={2.5cm 0 2.5cm 5.13in},clip,scale=0.45]{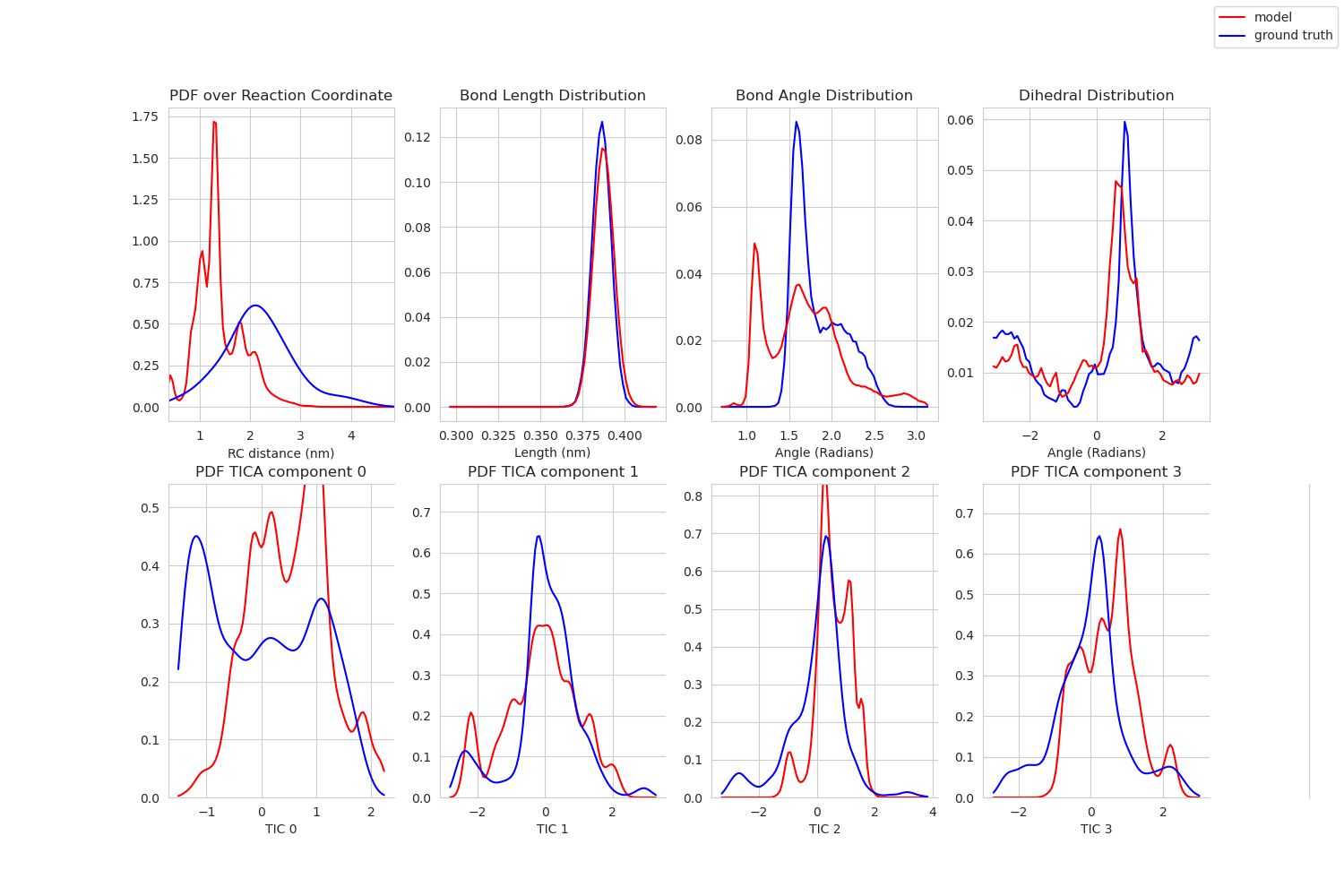}
    \caption{Under Trained (Bottom) vs Fully Trained (Top) CGSchNet Model BBA Final Benchmark 1D TICA Plots}
  \label{fgr:model_difference}
\end{figure}

\clearpage
In the TICA benchmark plots shown in Fig. \ref{fgr:model_difference}, the fully trained model (top) can better reproduce the key conformational features when compared to the under-trained model (bottom). This demonstrates the importance of model quality for accurately recovering low-dimensional projections of the energy landscape.

Table \ref{tab:w1dist} presents Wasserstein-1 distances between the model-predicted and ground-truth probability density functions across multiple metrics. The fully trained models generally achieve lower W\(_1\) distances, indicating a closer match to the true protein behavior compared to the under-trained counterparts. In Table \ref{tab:kldiv} we report the KL-divergences for the same distributions. 

\begin{table}[!ht]
\centering
\resizebox{\textwidth}{!}{%
\begin{tabular}{|l|cc|cc|cc|cc|cc|cc|}
\hline
\textbf{Protein} & \multicolumn{2}{c|}{\textbf{TIC 0}} & \multicolumn{2}{c|}{\textbf{TIC 1}} & \multicolumn{2}{c|}{\textbf{Bonds}} & \multicolumn{2}{c|}{\textbf{Angles}} & \multicolumn{2}{c|}{\textbf{Dihedrals}} \\
 & UT & FT & UT & FT & UT & FT & UT & FT & UT & FT  \\
\hline
a3D         & 0.6139 & 0.7346 & 0.3653 & 0.4382 & 0.0031 & 0.0004 & 0.0885 & 0.0529 & 0.4242 & 0.1107  \\
BBA             &  0.4902  & 0.2109 & 0.2095  & 0.2781 &0.0007 & 0.0006  & 0.0605  & 0.0441 & 0.4272 & 0.3215  \\
Chignolin       & 0.3609 & 0.1904 & 0.8948 & 0.8934 &  0.0023 & 0.0003 & 0.1803 & 0.0871 & 0.6544 &  0.5291 \\
Homeodomain     &  0.4917 & 0.4284 &  0.7547 & 0.4722 & 0.0026 & 0.0002 & 0.1845 & 0.0158 & 0.3238 & 0.2223 \\
\(\lambda\)-repressor & 0.7695 & 0.4016 & 0.6949 & 0.4392 & 0.0002 & 0.0003 & 0.0641 & 0.0569 & 0.2406 & 0.1836\\
Protein B       & 1.1227 & 0.5474 &  1.7827 & 0.5817 &  0.0014 & 0.0000 & 0.1276 & 0.2287 & 0.4887 & 0.1878 \\
Protein G      & 0.7457      & 0.5210 & 0.2944      & 0.4907 & 0.0019 & 0.0007 & 0.0471 & 0.0951 & 0.2949 & 0.4118  \\
Trp-cage        &  0.4597 & 0.4348 &  0.7833 & 0.3665 &  0.0003 & 0.0004 & 0.0565 &0.0460  & 0.1069 & 0.1854  \\
WW domain       &  0.4825  & 0.3321 & 0.3110 & 0.2071 & 0.0016 & 0.0005 & 0.0659 & 0.0356 & 0.5642 &  0.4661 \\
\hline
\end{tabular}
}
\caption{Wasserstein-1 distance between the under-trained (UT) and the fully-trained (FT) CGSchNet model PDFs across TIC 0, TIC 1, and structural features: bond lengths, angles and dihedrals for all proteins in our benchmark.}
\label{tab:w1dist}
\end{table}

\begin{table}[!ht]
\centering
\resizebox{\textwidth}{!}{%
\begin{tabular}{|l|cc|cc|cc|cc|cc|cc|}
\hline
\textbf{Protein} & \multicolumn{2}{c|}{\textbf{TIC 0}} & \multicolumn{2}{c|}{\textbf{TIC 1}} & \multicolumn{2}{c|}{\textbf{Bonds}} & \multicolumn{2}{c|}{\textbf{Angles}} & \multicolumn{2}{c|}{\textbf{Dihedrals}} \\
 & UT & FT & UT & FT & UT & FT & UT & FT & UT & FT\\
\hline
a3D     & 0.7080 & 4.8922 & 0.8750 & 3.8715 & 0.1545 & 0.0048 & 0.2076 & 0.0851 & 0.2134 & 0.1601 \\
BBA             & 0.6981 & 0.1614 &  0.2568 & 0.8866 & 0.0154  & 0.0086 & 0.4750 & 0.1100 & 0.2025 & 0.0925 \\
Chignolin       & 2.8054 & 0.2368  & 0.5400 & 0.4806 & 0.0842 & 0.0033 & 0.5551 & 0.1331 & 0.1776 & 0.0996 \\
Homeodomain     & 0.9845 & 3.7665 & 2.1244 & 2.1008 & 0.1561 & 0.0015 & 0.6603 & 0.0390 & 0.3144 & 0.0552  \\
\(\lambda\)-repressor & 3.1606 & 2.1323 & 3.7179 & 2.5487 & 0.0025 & 0.0019 & 0.1091 & 0.0748 & 0.0660 & 0.0361 \\
Protein B       & 5.6274 & 0.4249 & 0.9901 & 0.7843 & 0.0570 & 0.0008 & 0.5957 & 0.4226 & 0.2948 & 0.0644 \\
Protein G       &  3.4520 & 3.1381 & 1.0588 & 1.3059 & 0.0708 & 0.0076 & 0.2731 & 0.1556 & 0.1056 & 0.0848 \\
Trp-cage        & 2.9204 & 1.6817 & 1.0706 & 0.8146 & 0.0024 & 0.0038 & 0.1683 & 0.1234 & 0.0785 & 0.0554 \\
WW domain       & 5.0102 & 0.9144 & 1.3795 & 0.6805 &  0.0465 & 0.0059 & 0.3646 & 0.2254 & 0.4473 & 0.0841 \\
\hline
\end{tabular}
}
\caption{KL-Divergence -- \(D_\text{KL}(\text{GT} \parallel \text{Model})\) -- for the under-trained (UT) and the fully-trained (FT) CGSchNet model PDFs across TIC 0, TIC 1, and structural features: bond lengths, angles and dihedrals for all proteins in our benchmark.}
\label{tab:kldiv}
\end{table}

\subsubsection{Computation Times}

In Table \ref{tab:computationtimes}, we show the GPU time it takes for the CGSchNet model as well as implicit-solvent all-atom MD to reach coverages of 20\%, 50\% and 80\% of Chignolin's energy landscape. The computation times need to significantly increase (15x) in order to move from 50\% coverage up to 80\% for the all-atom when compared to the ML model. We also notice that the ML model offers a speedup of approximately 10-25x compared with all-atom implicit solvent MD simulations.

\begin{center}
\begin{table}[!ht]
\resizebox{\textwidth}{!}{ 
\begin{tabular}{|c|c c c|c c c|c|}
\hline
\textbf{Coverage} & \multicolumn{3}{c|}{\textbf{CG ML}} & \multicolumn{3}{c|}{\textbf{Implicit Solvent All Atom MD}} & \textbf{Speedup} \\
\cline{2-7}
 & Segments & GPU Time (s) & GPU Time (min) & Segments & GPU Time (s) & GPU Time (min) & \\
\hline
20\% & 102 & 74.29 & 1.24 & 906 & 841.67 & 14.03 & \textbf{11.33x} \\
50\% & 534 & 388.91 & 6.48 & 4,554 & 4230.67 & 70.51 & \textbf{10.88x} \\
80\% & 1782 & 1302.82 & 21.71 & 35,745 & 33207.11 & 553.45 & \textbf{25.49x} \\
\hline
\end{tabular}
}
\caption{Comparison of performance between CG ML (CGSchNet) fully trained model and implicit-solvent all-atom MD at different coverage levels, for Chignolin. Calculated by taking the average GPU time per segment and multiplying by the number of segments as direct GPU time was not accessible.}
\label{tab:computationtimes}

\end{table}
\end{center}

\section{Discussion}

Our modular benchmark offers an extendable, unbiased, and computationally efficient methodology for evaluating MD simulations across conformational exploration and accuracy, and through the weighted ensembles, the acceleration required to capture slow kinetic processes. Its modular design enables the addition of new simulation methods, data, and visualizations, fostering continual expansion and collaboration within the community. 

When doing analysis on components of our benchmark, we noticed the MSM and the weighted data methodologies used to generate the TICA PDFs are closely aligned, as both are derived from the estimating equilibrium dynamics. Thus, it is expected that the two will generally be consistent despite minor discrepancies that occasionally occur. It is worth noting that the weighting system requires the use of WESTPA, whereas a MSM can be derived independently of WESTPA.

The majority of the benchmark results can be found in the supporting information, where there are some interesting trends. As an example, while the under trained CGSchNet model mostly performed poorly in comparison to the fully trained model, a3D did perform better in key benchmark metrics, indeed exploring beyond the fully trained model and even the implicit data. While a surprising result, our benchmark has identified this anomaly, helping us ask questions about the model and the reasoning for its performance.

Using the weighted ensemble enhanced sampling framework, we enable MD and ML models to dynamically transition between conformational states. Additionally, we validate that it exhibits physically realistic behavior within each state using the local structural properties. This provides a strong basis for evaluating the fidelity of a protein model. Such an approach ensures that the conformational transitions observed are the result of intrinsic thermodynamic and structural properties of the system, rather than artifacts of initialization. 

When simulations capture accurate intramolecular behavior (correct bond lengths, bond angles, dihedral angle distributions) and transition dynamics, it serves as compelling evidence that the model accurately reflects the underlying energy landscape and dynamic profile of the protein.



\section{Conclusion}

We have introduced a modular, extensible benchmarking framework for machine-learned molecular dynamics, grounded in enhanced sampling via the weighted ensemble method and rigorous slow motion analysis using tools such as TICA and Markov State Models. By comparing conformational sampling and structural fidelity against high-quality ground truth data, the framework enables robust and reproducible evaluations across diverse simulation methodologies. Our integration of custom propagators, adaptive binning, and multidimensional progress coordinates makes the platform versatile. The modularity allows for compatibility with both classical force fields and machine learning-based models. Through extensive testing on a curated set of proteins, we demonstrated the benchmark's ability to distinguish between under-trained and fully trained models (Fig. \ref{fgr:good_model_explored}), as well as its capacity to sample the space effectively from only a single starting conformation.

While we believe the benchmark provides significant value to the community, several important limitations remain. Most notably, current simulation strategies primarily focus on driving the protein as far as possible across the conformational landscape. However, this approach overlooks other crucial aspects of protein dynamics. For instance, demonstrating the ability to return to the initial state would validate that the observed transitions are reversible and not merely exploratory. Additionally, keeping track of transitions between distinct macrostates would offer deeper insights into the protein's functional behavior and the completeness of the sampled conformational space. 

Looking forward, this benchmark sets the stage for community-wide standardization in MD validation, offering a consistent methodology to assess performance, accuracy, and efficiency. Its open-source nature encourages collaborative development, with opportunities to expand into larger proteins or alternative sampling techniques. Future iterations may incorporate metrics for reversibility, transition pathway analysis, and state recrossing behavior, which are essential for validating long-timescale protein kinetics. The benchmark strategy of utilizing weighted ensembles and TICA may be used for more complex processes such as protein interactions or self-assembly processes. By providing a transparent and physically grounded benchmark, we hope to accelerate the development of reliable, scalable, and interpretable, and most importantly consistently measurable simulation tools in computational biophysics.

\section{Data Availability}
We have provided the ground truth dataset, benchmark constructions files, and results which are accessible via the following link: \href{http://raz.ucsc.edu/benchmark\_data\_300K/}{http://raz.ucsc.edu/benchmark\_data\_300K/}.\\
The code for our benchmark can be found here: \\\href{ https://github.com/BioMedAI-UCSC/CGML\_Driver.git}{ https://github.com/BioMedAI-UCSC/CGML\_Driver.git}\\

\section{Supporting Information Description}
\begin{itemize}
    \item Additional ground truth data, figures, and details. Also, a complete benchmark set of all proteins evaluated for the Implicit All-Atom, CGSchNet Undertrained Model, and the CGSchNet Fully Trained model. A complete benchmark includes the contact map diagram, PDFs over structure and TICA components, 2D TICA contours, Macrostates, comparison points in TICA space, and KL/W1 tables as described in the manuscript. 
\end{itemize}

\section{Acknowledgments}
This research used resources of the National Energy Research Scientific Computing Center (NERSC), a Department of Energy User Facility using NERSC award m4229-ERCAP 0033479.

This research also used the Delta advanced computing and data resource which is supported by the National Science Foundation (award OAC 2005572) and the State of Illinois. Delta is a joint effort of the University of Illinois Urbana-Champaign and its National Center for Supercomputing Applications.

This work was supported by the U.S. National Science Foundation under Grant No. 2451680 from the Directorate for Technology, Innovation and Partnerships.

\bibliography{cites}

\clearpage 

\begin{tocentry}
\centering
\includegraphics[width=3.25in]{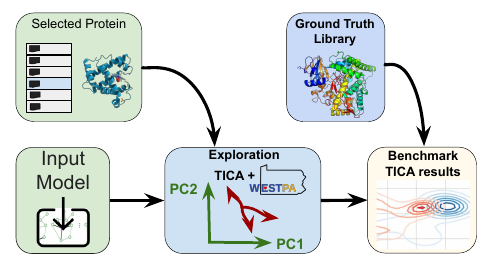}
\label{TOC Graphic}
\end{tocentry}

\end{document}


\vspace{1em}

\newpage
\setcounter{page}{1}
\renewcommand{\thepage}{S\arabic{page}}

\subsection{Ground Truth Data}
\begin{table}[h!]
\centering
\begin{tabular}{|c|c|}
\hline
\textbf{Protein} & \textbf{Number of Starting Points} \\
\hline
a3D & 1580 \\
BBA & 858 \\
Chignolin & 372 \\
Homeodomain & 1668 \\
$\lambda$-repressor & 1199 \\
Protein B & 401 \\
Protein G & 2560 \\
Trp-cage & 456 \\
WWdomain & 900 \\
\hline
\end{tabular}
\caption{Total number of starting points for each benchmark protein for ground truth data construction.}
\label{tab:starting_points}
\end{table}

\subsubsection{Ground Truth Exploration by Starting Point}
Note: For the follow images, each starting point is sampled at approximately 1 in 10 and therefore this is just a subset. This shows how each individual starting point doesn't explore a significant amount of the conformational space, but combined they create a detailed view of possible conformations and their transitions. The legend on the side (R\#) where \# is the starting point number (divided by 10). 

\begin{figure}[htbp]
    \centering
    \includegraphics[trim={0 0 0 0},clip,width=\linewidth]{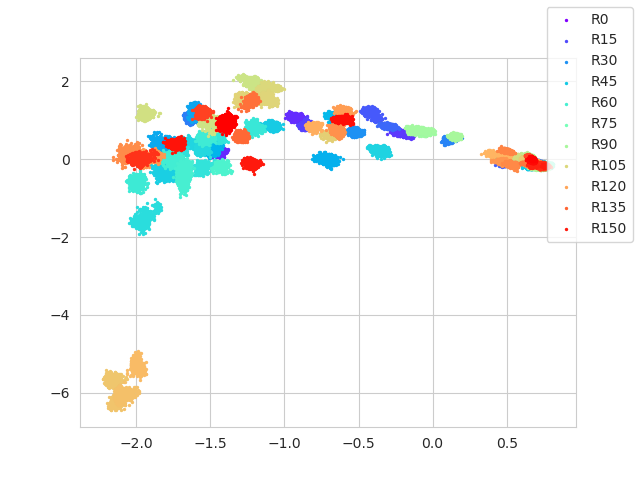}
    \caption{a3D Ground Truth Exploration Colored by Starting Point}
\end{figure}
\begin{figure}[htbp]
    \centering
    \includegraphics[trim={0 0 0 0},clip,width=\linewidth]{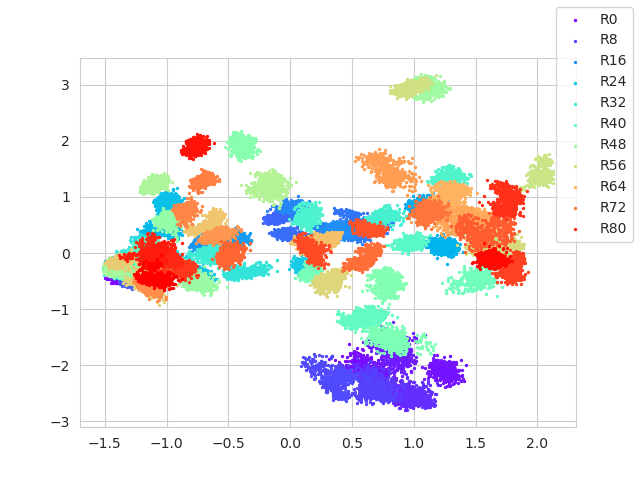}
    \caption{BBA Ground Truth Exploration Colored by Starting Point}
\end{figure}
\begin{figure}[htbp]
    \centering
    \includegraphics[trim={0 0 0 0},clip,width=\linewidth]{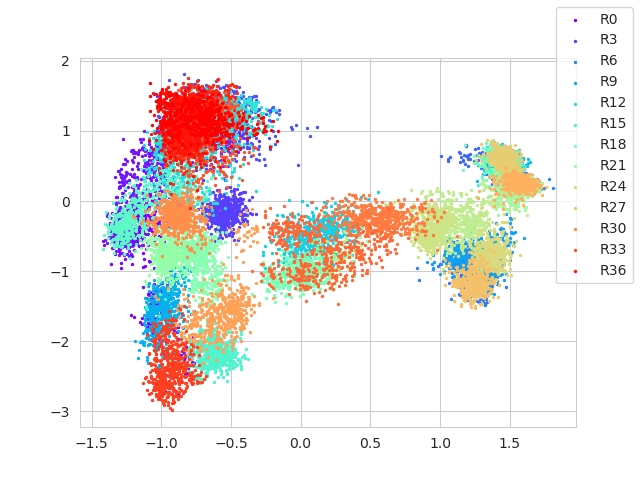}
    \caption{Chignolin Ground Truth Exploration Colored by Starting Point}
\end{figure}
\begin{figure}[htbp]
    \centering
    \includegraphics[trim={0 0 0 0},clip,width=\linewidth]{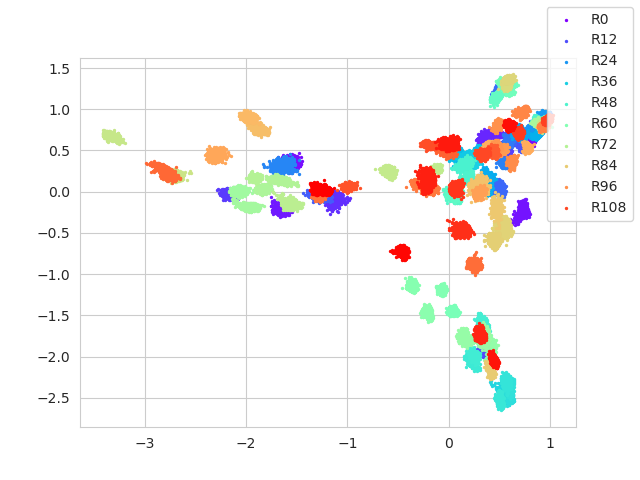}
    \caption{$\lambda$-repressor Ground Truth Exploration Colored by Starting Point}
\end{figure}
\begin{figure}[htbp]
    \centering
    \includegraphics[trim={0 0 0 0},clip,width=\linewidth]{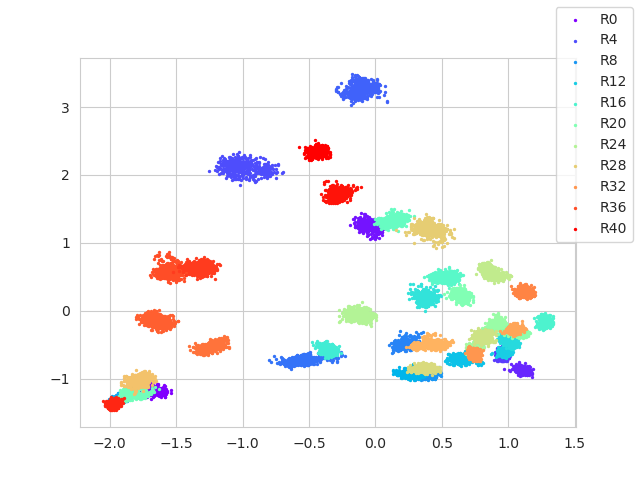}
    \caption{Protein B Ground Truth Exploration Colored by Starting Point}
\end{figure}
\begin{figure}[htbp]
    \centering
    \includegraphics[trim={0 0 0 0},clip,width=\linewidth]{images/ground_truth/gt_plot_proteing.png}
    \caption{Protein G Ground Truth Exploration Colored by Starting Point}
\end{figure}
\begin{figure}[htbp]
    \centering
    \includegraphics[trim={0 0 0 0},clip,width=\linewidth]{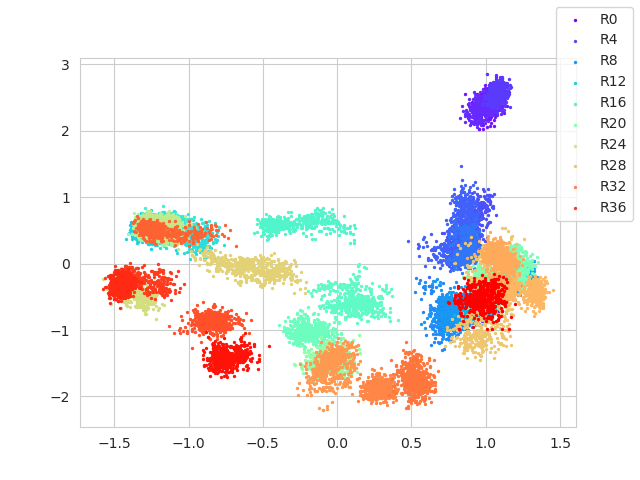}
    \caption{Trp-cage Ground Truth Exploration Colored by Starting Point}
\end{figure}
\begin{figure}[htbp]
    \centering
    \includegraphics[trim={0 0 0 0},clip,width=\linewidth]{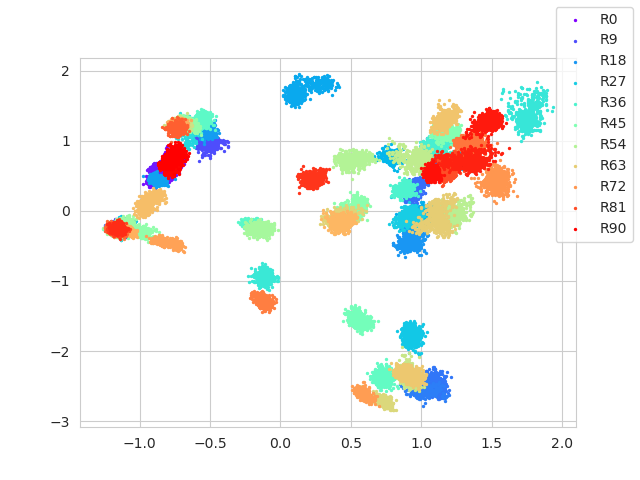}
    \caption{WWdomain Ground Truth Exploration Colored by Starting Point}
\end{figure}
\clearpage

\subsection{Full Benchmark Diagrams - Implicit Solvent All Atom MD}

\subsubsection{a3D}

\begin{figure}[htbp]
    \centering
    \includegraphics[trim={0 0 0 0.75cm},clip,width=0.5\linewidth]{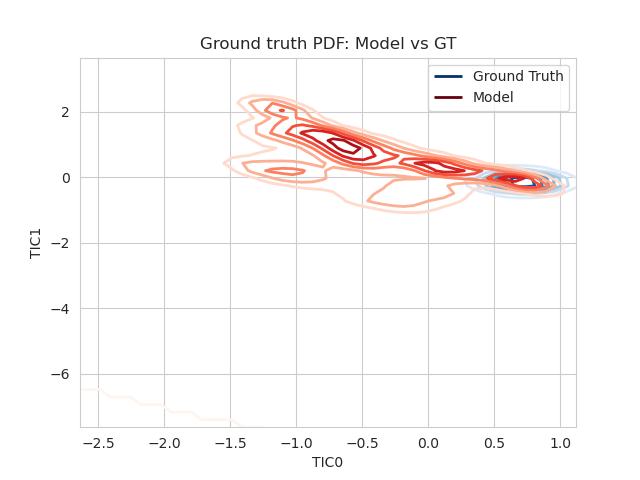}
    \caption{TICA contours for a3D - Implicit Solvent All Atom MD}
    \label{fgr:tica_contours_bba}
\end{figure}

\begin{figure}[htbp]
    \centering
    \includegraphics[trim={0 0 0 0.75cm},clip,width=0.5\linewidth]{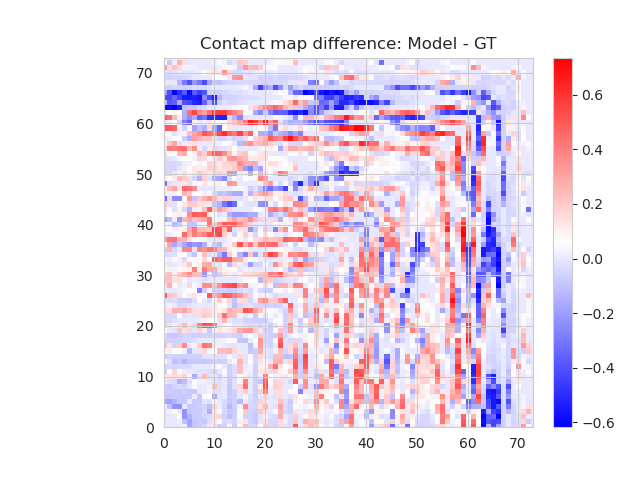}
    \caption{Contact map for a3D - Implicit Solvent All Atom MD}
\end{figure}

\begin{figure}[htbp]
    \centering
    \includegraphics[trim={0 0 0 0cm},clip,width=\linewidth]{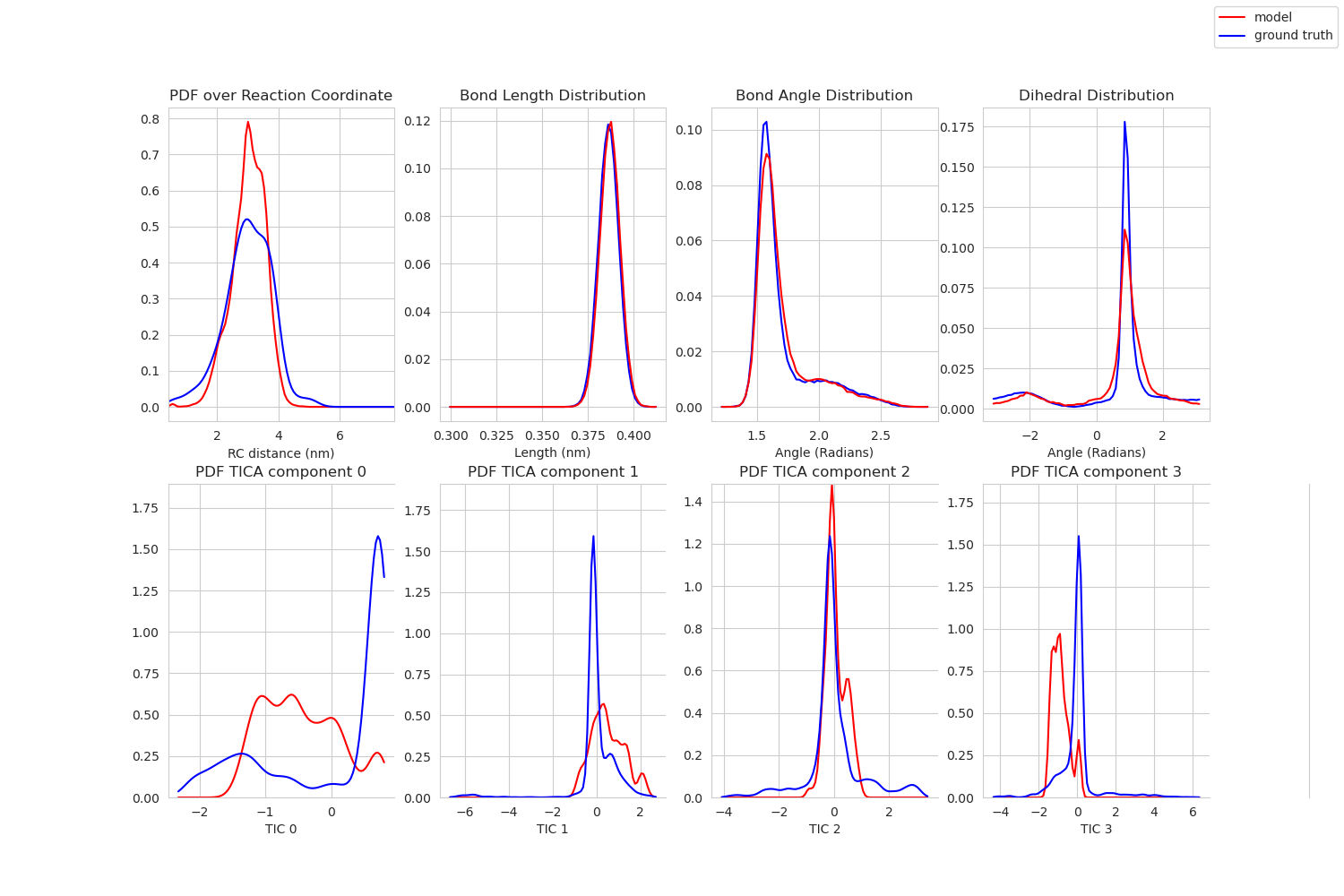}
    \caption{PDFs for a3D - Implicit Solvent All Atom MD}
    \label{fgr:pdfs_bba}
\end{figure}

\begin{figure}[htbp]
    \centering
    \includegraphics[trim={0 0 0 0},clip,width=\linewidth]{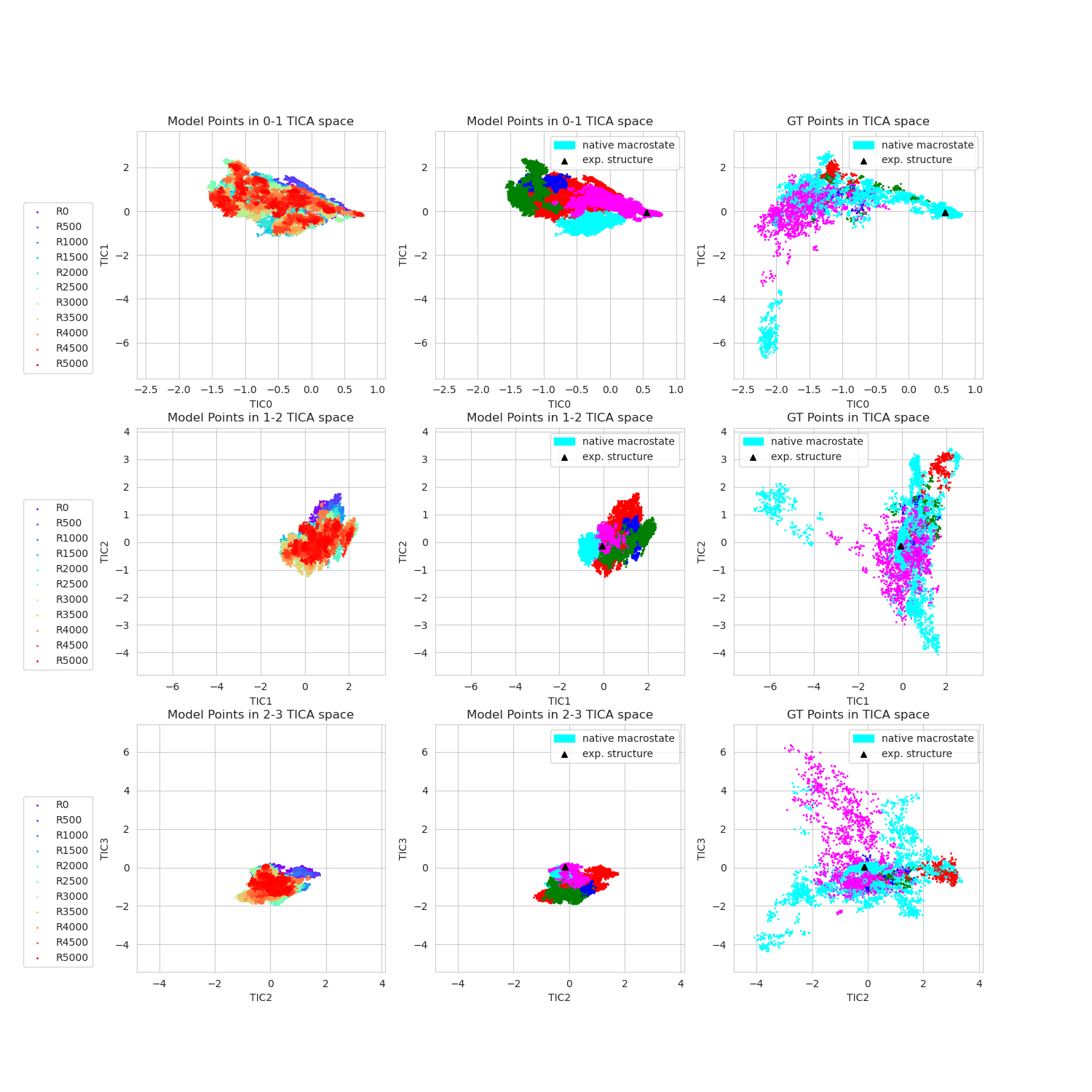}
    \caption{TICA space projections for a3D - Implicit Solvent All Atom MD}
    \label{fgr:tica_spaces_bba}
\end{figure}
\clearpage
\begin{table}[h!]
\centering
\resizebox{\textwidth}{!}{%
\begin{tabular}{|l|c|c|c|c|c|c|c|c|}
\hline
\textbf{Metric} & \textbf{TIC 0} & \textbf{TIC 1} & \textbf{TIC 2} & \textbf{TIC 3} & \textbf{Bonds} & \textbf{Angles} & \textbf{Dihedrals} & \textbf{Gyration} \\
\hline
KL & 1.6437 & 0.7745 & 1.6595 & 1.8519 & 0.0075 & 0.0187 & 0.2106 & 1.6698 \\
W1 & 0.5673 & 0.5367 & 0.3729 & 0.9209 & 0.0006 & 0.0144 & 0.3540 & 0.1492 \\
\hline
\end{tabular}
}
\caption{KL and W1 metrics for a3D with Implicit All Atom MD}
\end{table}

\clearpage

\subsubsection{BBA}

\begin{figure}[htbp]
    \centering
    \includegraphics[trim={0 0 0 0.75cm},clip,width=0.5\linewidth]{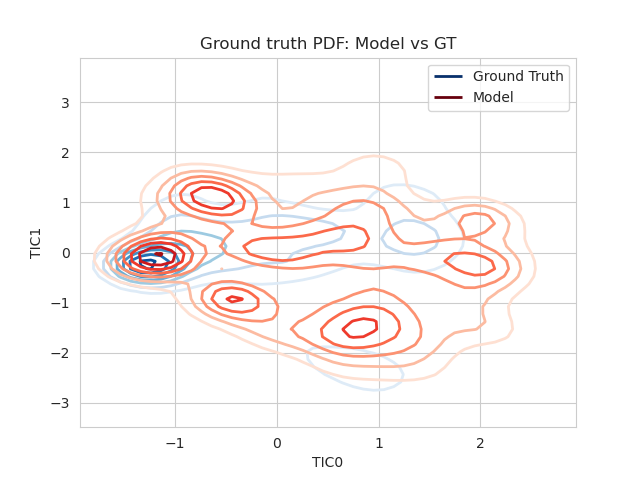}
    \caption{TICA contours for BBA - Implicit Solvent All Atom MD}
    \label{fgr:tica_contours_bba}
\end{figure}

\begin{figure}[htbp]
    \centering
    \includegraphics[trim={0 0 0 0.75cm},clip,width=0.5\linewidth]{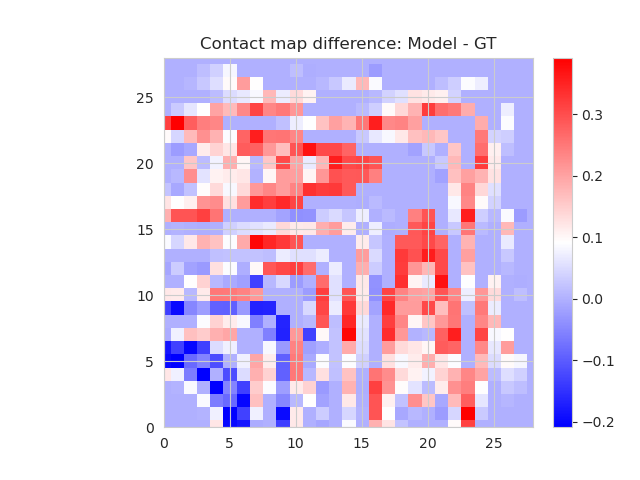}
    \caption{Contact map for BBA - Implicit Solvent All Atom MD}
\end{figure}

\begin{figure}[htbp]
    \centering
    \includegraphics[trim={0 0 0 0cm},clip,width=\linewidth]{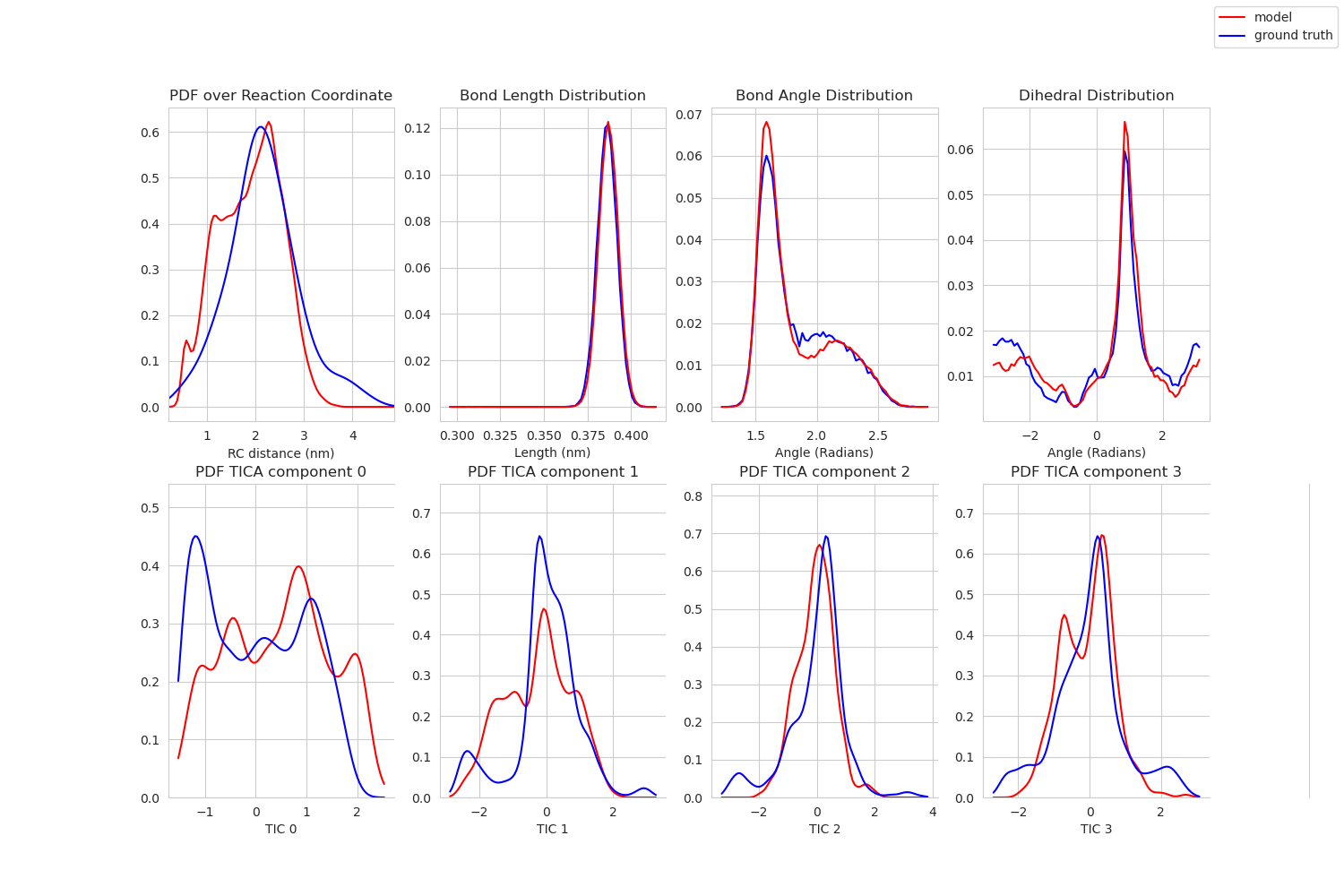}
    \caption{PDFs for BBA - Implicit Solvent All Atom MD}
    \label{fgr:pdfs_bba}
\end{figure}

\begin{figure}[htbp]
    \centering
    \includegraphics[trim={0 0 0 0},clip,width=\linewidth]{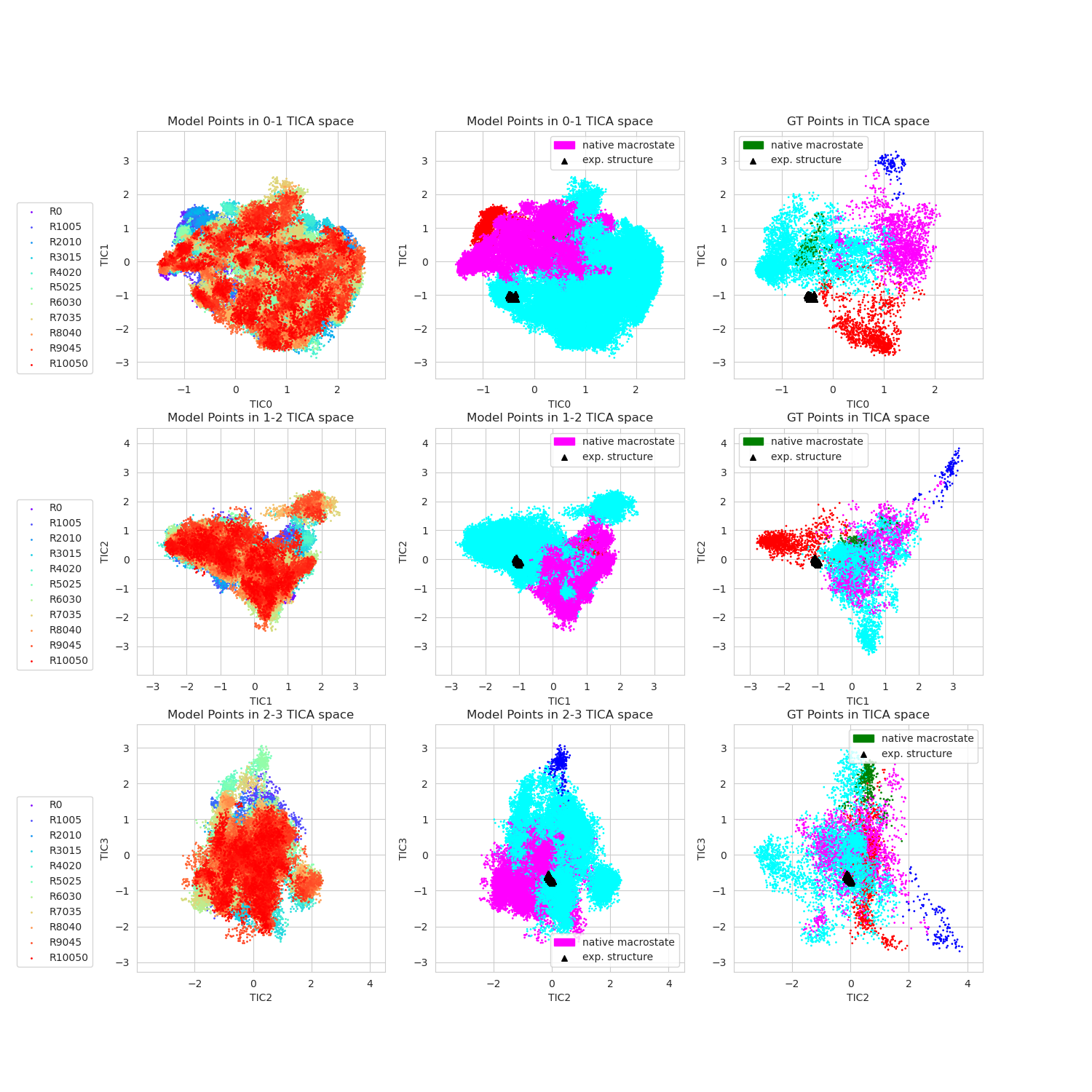}
    \caption{TICA space projections for BBA - Implicit Solvent All Atom MD}
    \label{fgr:tica_spaces_bba}
\end{figure}
\clearpage
\begin{table}[h!]
\centering
\resizebox{\textwidth}{!}{%
\begin{tabular}{|l|c|c|c|c|c|c|c|c|}
\hline
\textbf{Metric} & \textbf{TIC 0} & \textbf{TIC 1} & \textbf{TIC 2} & \textbf{TIC 3} & \textbf{Bonds} & \textbf{Angles} & \textbf{Dihedrals} & \textbf{Gyration} \\
\hline
KL & 0.1826 & 0.3306 & 0.6010 & 0.2696 & 0.0066 & 0.0073 & 0.0697 & 0.2199 \\
W1 & 0.4290 & 0.2915 & 0.2544 & 0.1743 & 0.0005 & 0.0273 & 0.4423 & 0.0843 \\
\hline
\end{tabular}
}
\caption{KL and W1 metrics for BBA with Implicit All Atom MD}
\end{table}
\clearpage

\subsubsection{Chignolin}

\begin{figure}[htbp]
    \centering
    \includegraphics[trim={0 0 0 0.75cm},clip,width=0.5\linewidth]{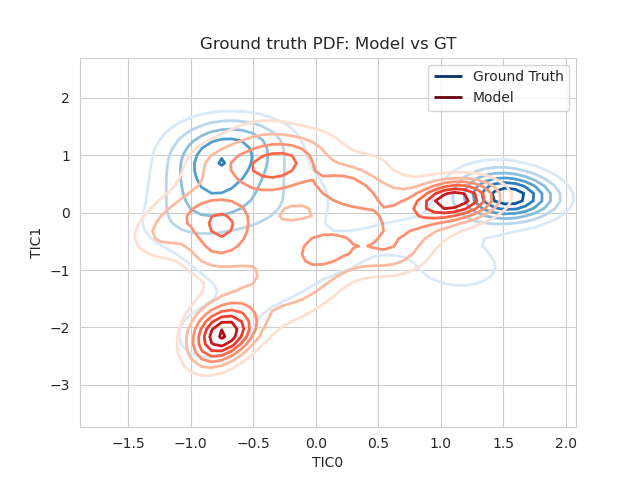}
    \caption{TICA contours for Chignolin - Implicit Solvent All Atom MD}
    \label{fgr:tica_contours_chignolin}
\end{figure}

\begin{figure}[htbp]
    \centering
    \includegraphics[trim={0 0 0 0.75cm},clip,width=0.5\linewidth]{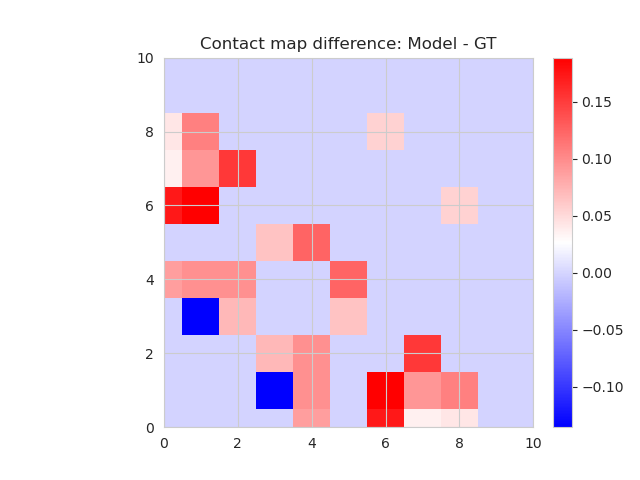}
    \caption{Contact map for Chignolin - Implicit Solvent All Atom MD}
\end{figure}

\begin{figure}[htbp]
    \centering
    \includegraphics[trim={0 0 0 0cm},clip,width=\linewidth]{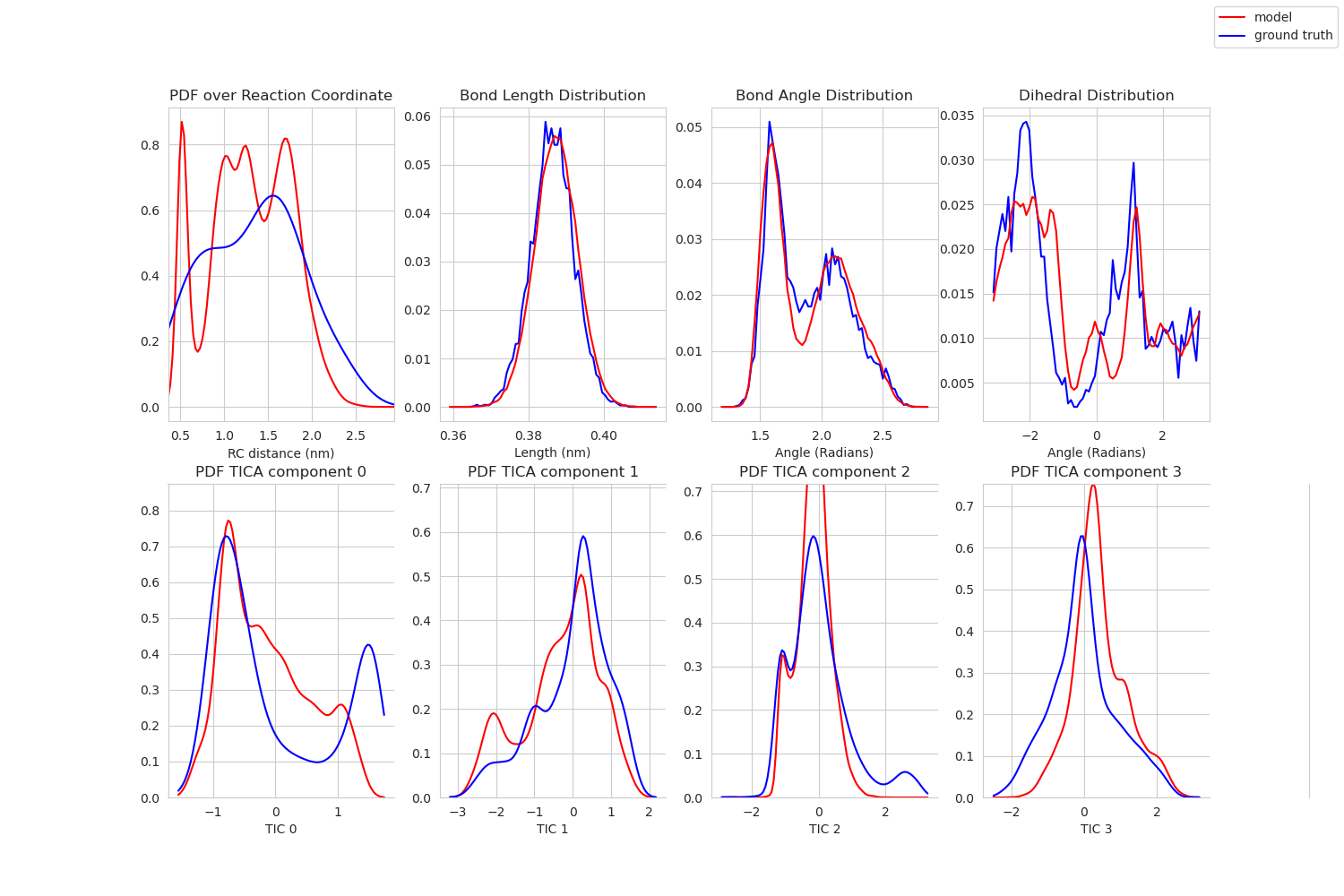}
    \caption{PDFs for Chignolin - Implicit Solvent All Atom MD}
    \label{fgr:pdfs_chignolin}
\end{figure}

\begin{figure}[htbp]
    \centering
    \includegraphics[trim={0 0 0 0},clip,width=\linewidth]{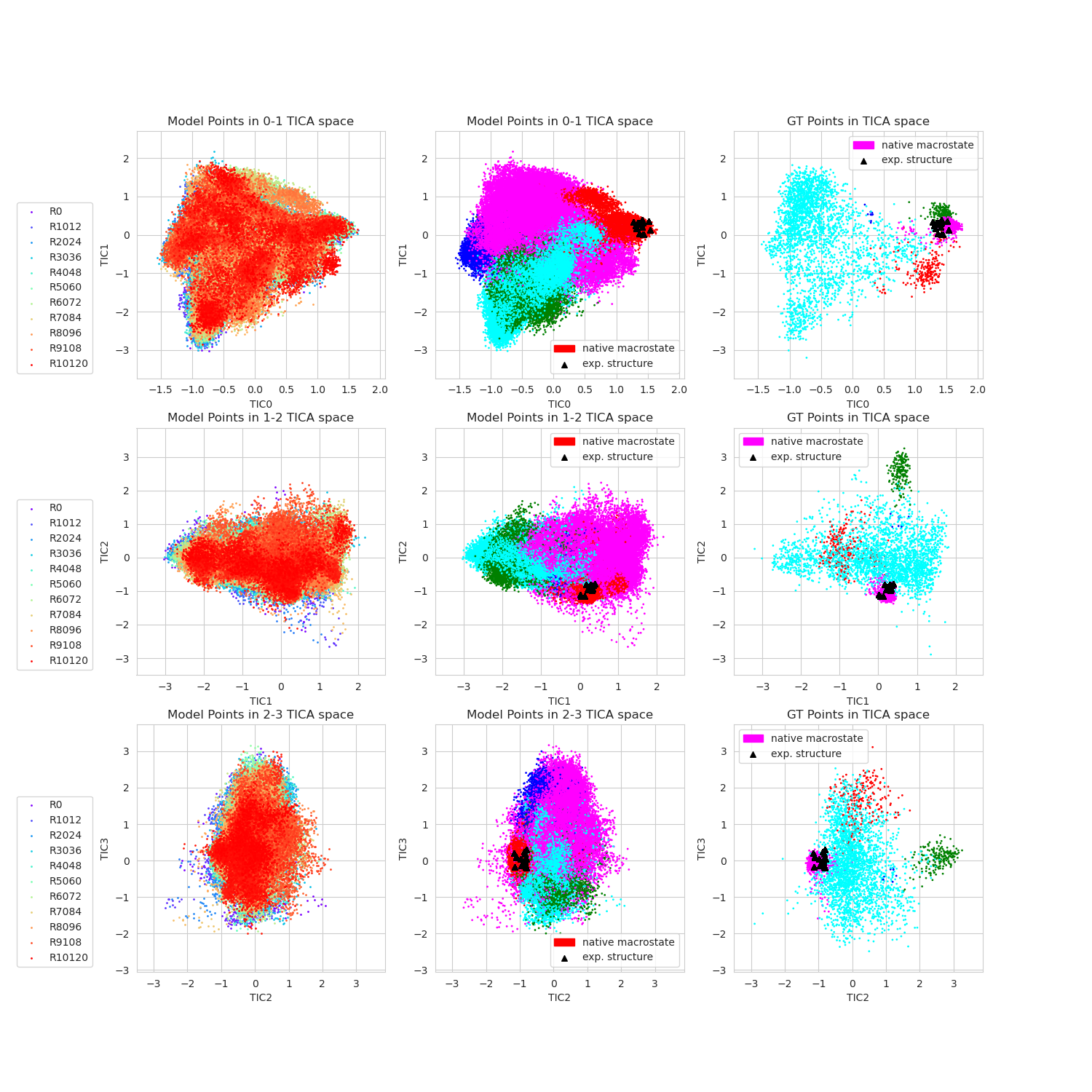}
    \caption{TICA space projections for Chignolin - Implicit Solvent All Atom MD}
    \label{fgr:tica_spaces_chignolin}
\end{figure}
\clearpage
\begin{table}[h!]
\centering
\resizebox{\textwidth}{!}{%
\begin{tabular}{|l|c|c|c|c|c|c|c|c|}
\hline
\textbf{Metric} & \textbf{TIC 0} & \textbf{TIC 1} & \textbf{TIC 2} & \textbf{TIC 3} & \textbf{Bonds} & \textbf{Angles} & \textbf{Dihedrals} & \textbf{Gyration} \\
\hline
KL & 0.2368 & 0.4806 & 0.1442 & 0.1353 & 0.0033 & 0.1331 & 0.0996 & 0.9233 \\
W1 & 0.1904 & 0.8934 & 0.2074 & 0.2368 & 0.0003 & 0.0871 & 0.5291 & 0.0947 \\
\hline
\end{tabular}
}
\caption{KL and W1 metrics for Chignolin with Implicit All Atom MD}
\end{table}
\clearpage

\subsubsection{Homeodomain}

\begin{figure}[htbp]
    \centering
    \includegraphics[trim={0 0 0 0.75cm},clip,width=0.5\linewidth]{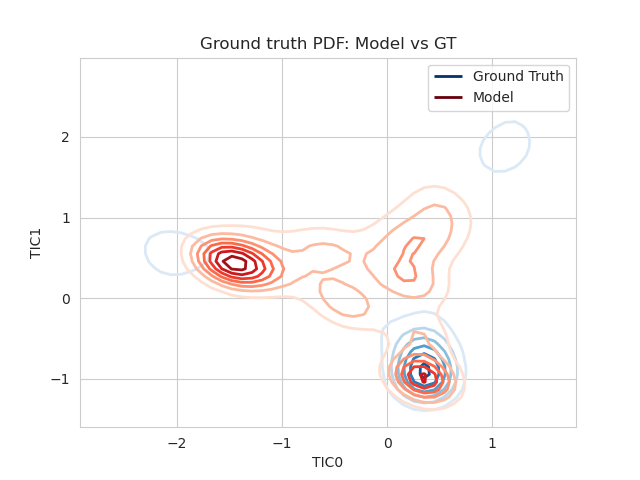}
    \caption{TICA contours for Homeodomain - Implicit Solvent All Atom MD}
    \label{fgr:tica_contours_homeodomain}
\end{figure}

\begin{figure}[htbp]
    \centering
    \includegraphics[trim={0 0 0 0.75cm},clip,width=0.5\linewidth]{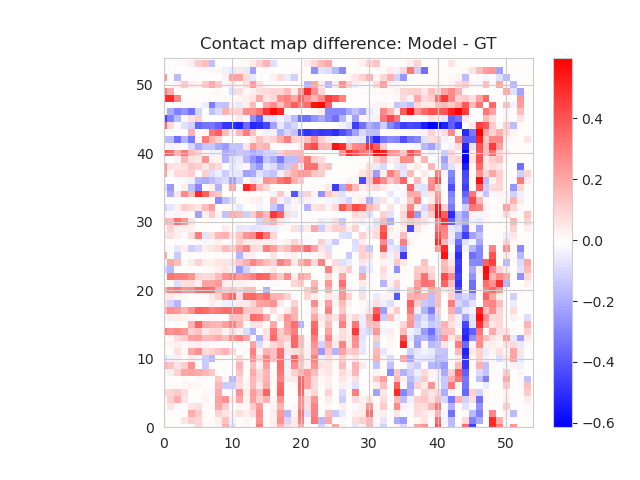}
    \caption{Contact map for Homeodomain - Implicit Solvent All Atom MD}
\end{figure}

\begin{figure}[htbp]
    \centering
    \includegraphics[trim={0 0 0 0cm},clip,width=\linewidth]{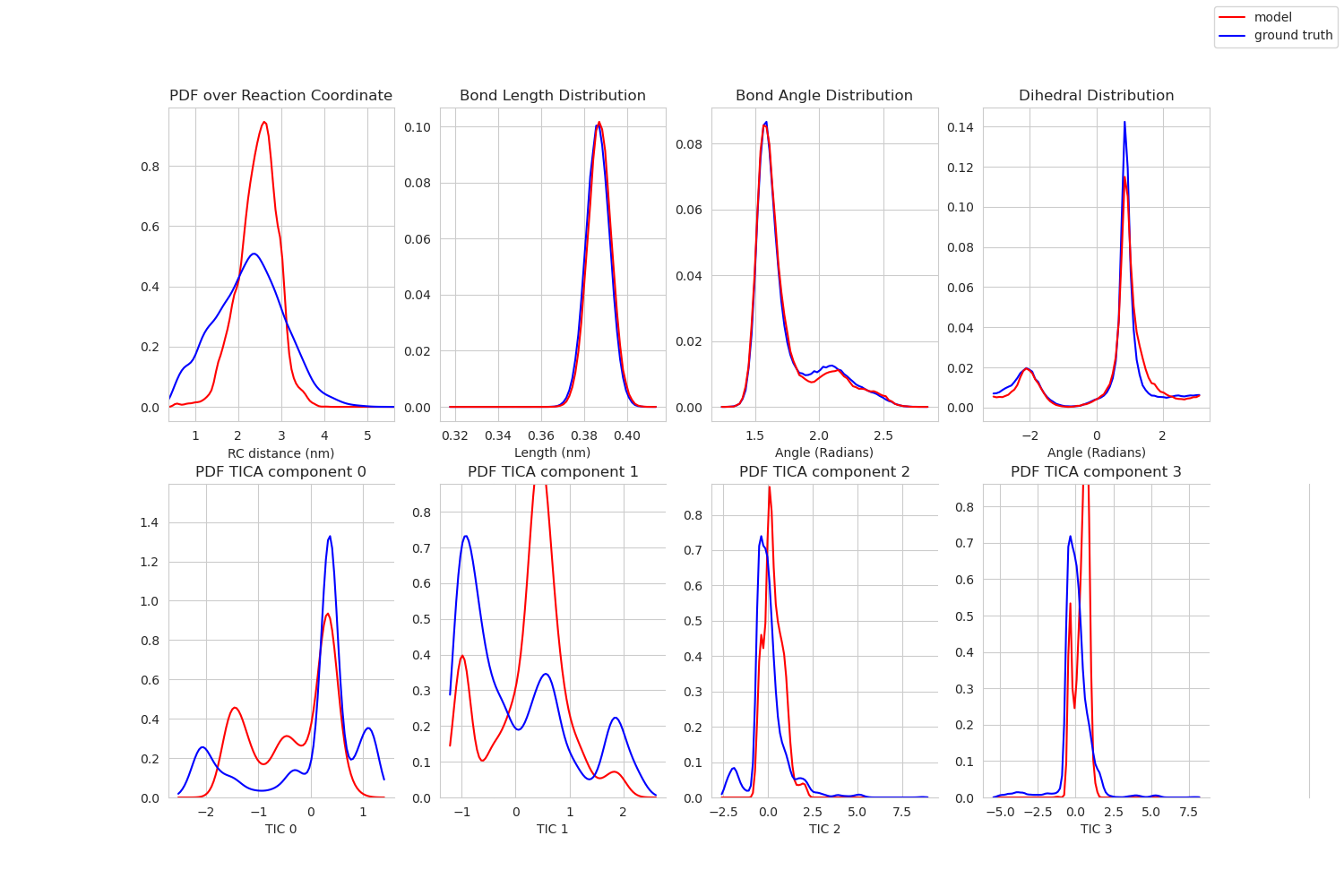}
    \caption{PDFs for Homeodomain - Implicit Solvent All Atom MD}
    \label{fgr:pdfs_homeodomain}
\end{figure}

\begin{figure}[htbp]
    \centering
    \includegraphics[trim={0 0 0 0},clip,width=\linewidth]{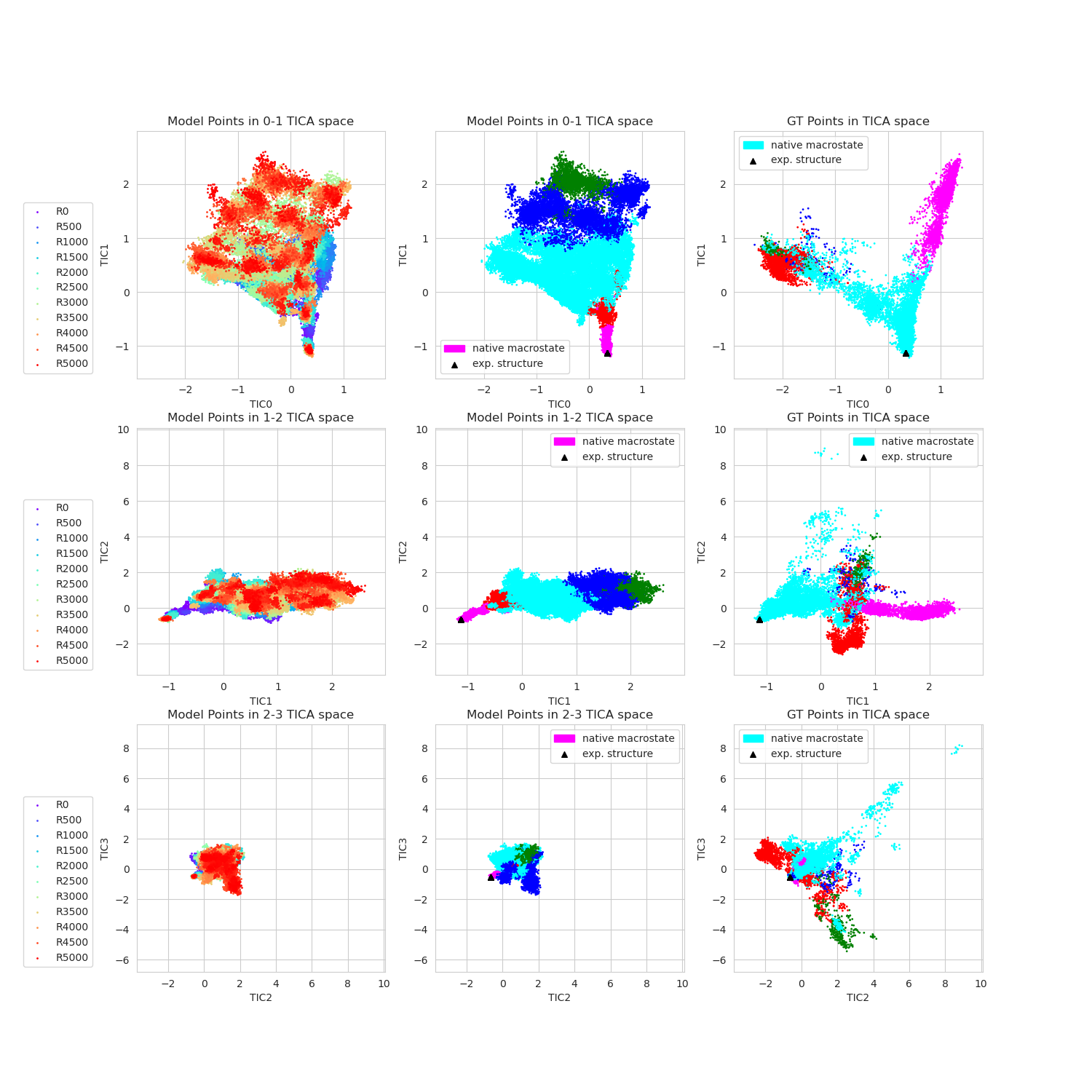}
    \caption{TICA space projections for Homeodomain - Implicit Solvent All Atom MD}
    \label{fgr:tica_spaces_homeodomain}
\end{figure}
\clearpage

\begin{table}[h!]
\centering
\resizebox{\textwidth}{!}{%
\begin{tabular}{|l|c|c|c|c|c|c|c|c|}
\hline
\textbf{Metric} & \textbf{TIC 0} & \textbf{TIC 1} & \textbf{TIC 2} & \textbf{TIC 3} & \textbf{Bonds} & \textbf{Angles} & \textbf{Dihedrals} & \textbf{Gyration} \\
\hline
KL & 1.1512 & 0.4092 & 1.0635 & 0.8944 & 0.0105 & 0.0164 & 0.0546 & 0.9459 \\
W1 & 0.4714 & 0.4423 & 0.4483 & 0.5334 & 0.0008 & 0.0309 & 0.1103 & 0.0463 \\
\hline
\end{tabular}
}
\caption{KL and W1 metrics for Homeodomain with Implicit All Atom MD}
\end{table}
\clearpage

\clearpage

\subsubsection{$\lambda$-repressor}

\begin{figure}[htbp]
    \centering
    \includegraphics[trim={0 0 0 0.75cm},clip,width=0.5\linewidth]{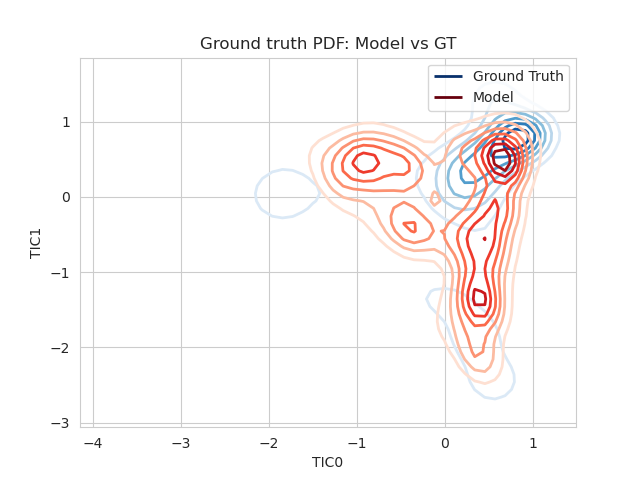}
    \caption{TICA contours for $\lambda$-repressor - Implicit Solvent All Atom MD}
    \label{fgr:tica_contours_homeodomain}
\end{figure}

\begin{figure}[htbp]
    \centering
    \includegraphics[trim={0 0 0 0.75cm},clip,width=0.5\linewidth]{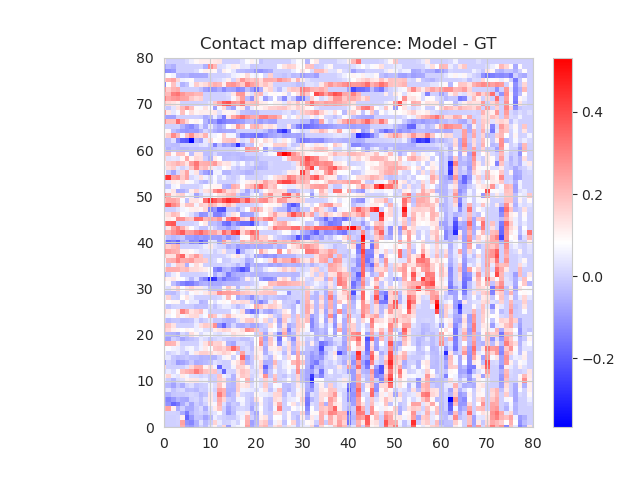}
    \caption{Contact map for $\lambda$-repressor - Implicit Solvent All Atom MD}
\end{figure}

\begin{figure}[htbp]
    \centering
    \includegraphics[trim={0 0 0 0cm},clip,width=\linewidth]{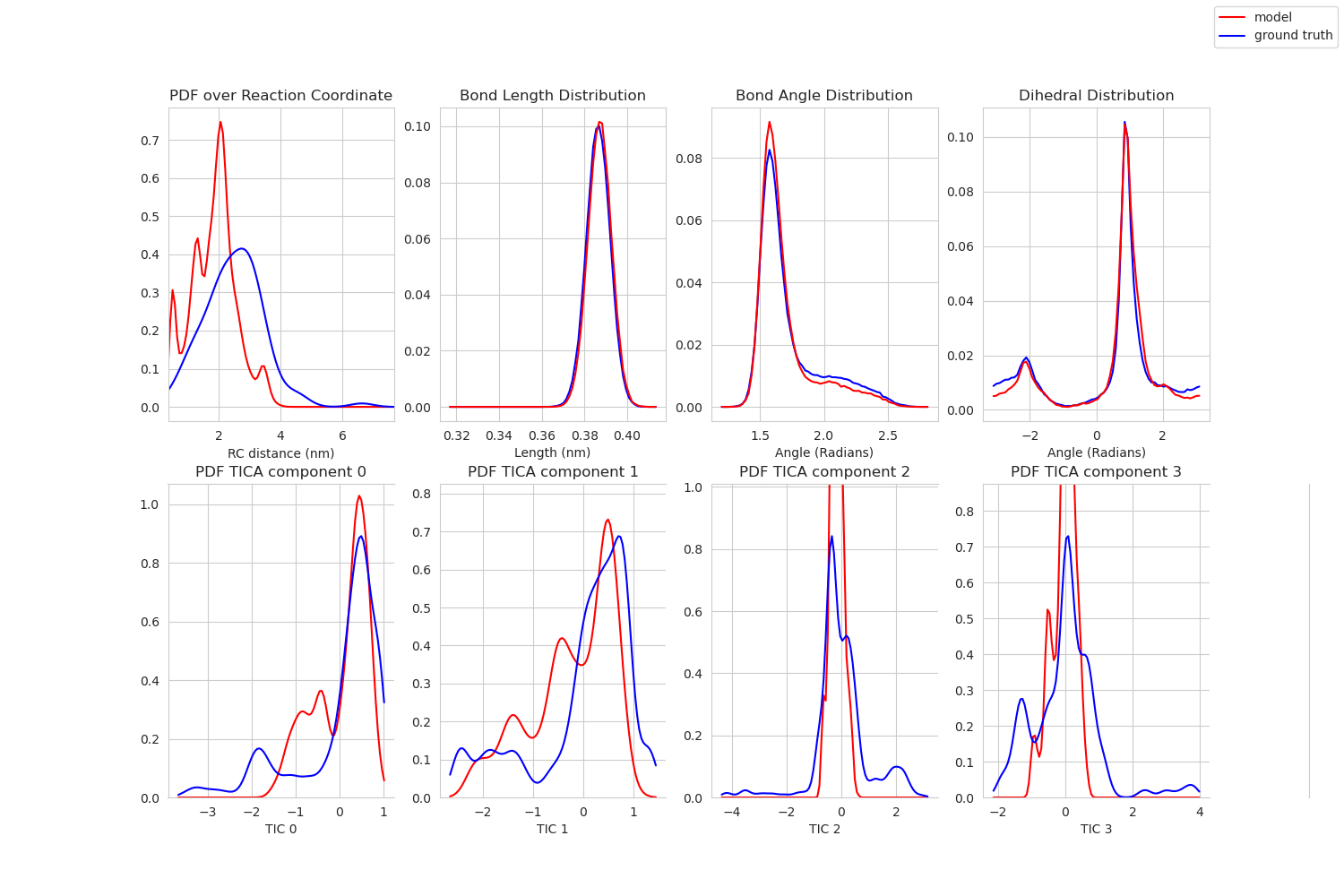}
    \caption{PDFs for $\lambda$-repressor - Implicit Solvent All Atom MD}
    \label{fgr:pdfs_homeodomain}
\end{figure}

\begin{figure}[htbp]
    \centering
    \includegraphics[trim={0 0 0 0},clip,width=\linewidth]{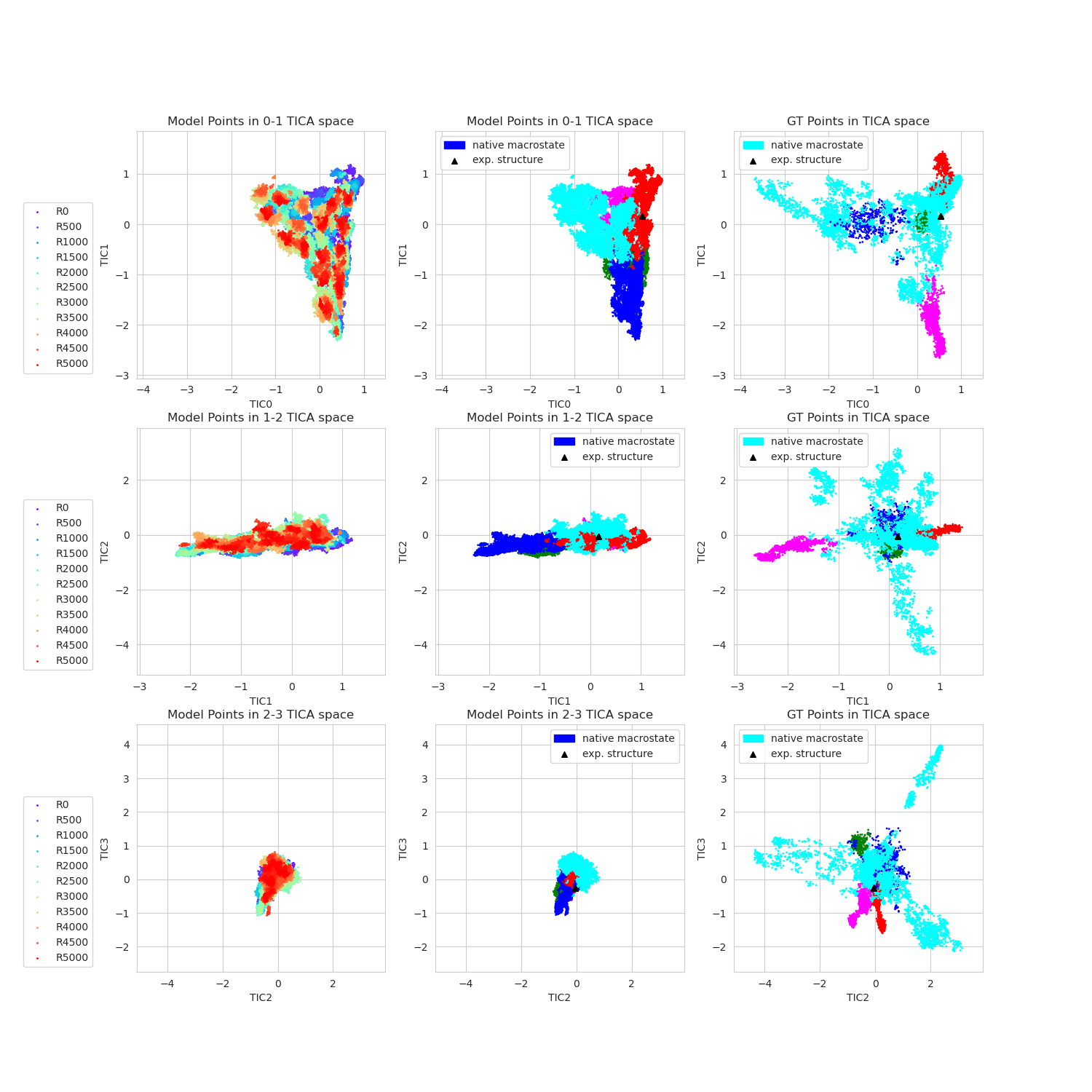}
    \caption{TICA space projections for $\lambda$-repressor - Implicit Solvent All Atom MD}
    \label{fgr:tica_spaces_homeodomain}
\end{figure}
\clearpage

\subsubsection{Protein B}

\begin{figure}[htbp]
    \centering
    \includegraphics[trim={0 0 0 0.75cm},clip,width=0.5\linewidth]{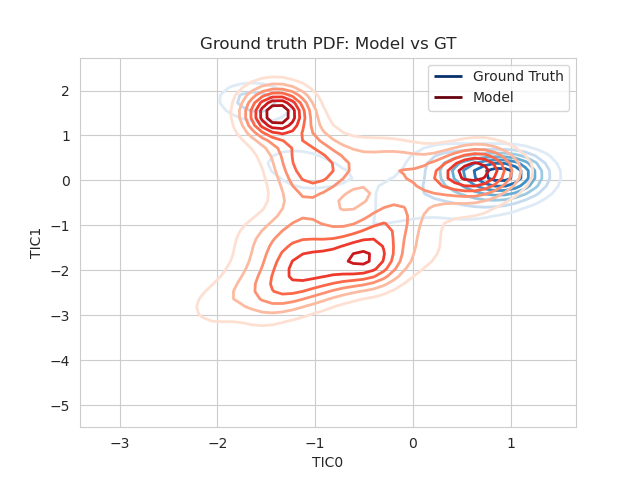}
    \caption{TICA contours for Protein B - Implicit Solvent All Atom MD}
    \label{fgr:tica_contours_proteinb}
\end{figure}

\begin{figure}[htbp]
    \centering
    \includegraphics[trim={0 0 0 0.75cm},clip,width=0.5\linewidth]{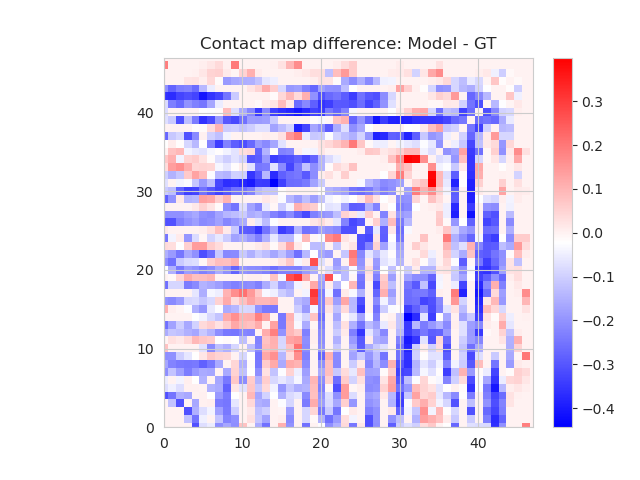}
    \caption{Contact map for Protein B - Implicit Solvent All Atom MD}
\end{figure}

\begin{figure}[htbp]
    \centering
    \includegraphics[trim={0 0 0 0cm},clip,width=\linewidth]{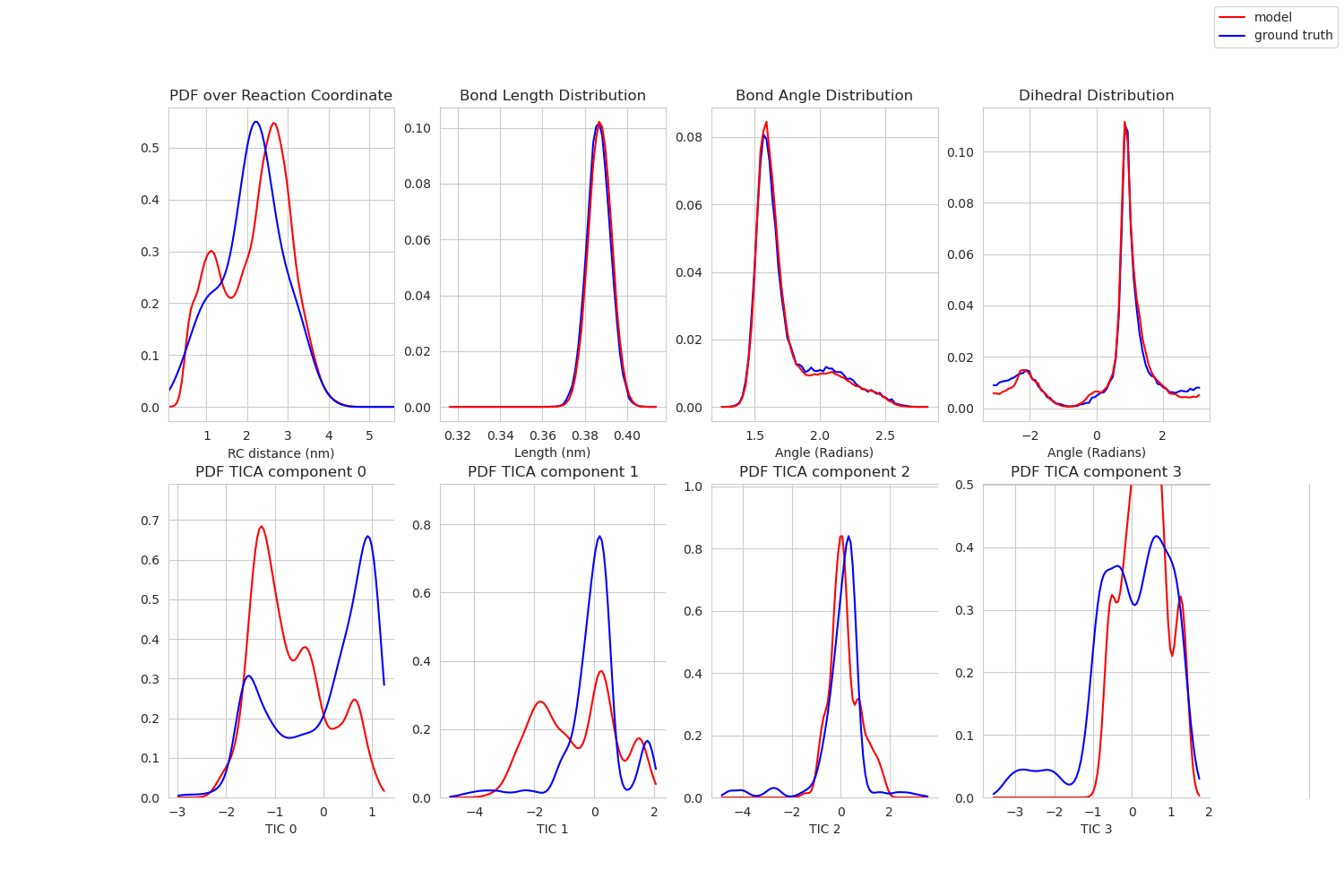}
    \caption{PDFs for Protein B - Implicit Solvent All Atom MD}
    \label{fgr:pdfs_proteinb}
\end{figure}

\begin{figure}[htbp]
    \centering
    \includegraphics[trim={0 0 0 0},clip,width=\linewidth]{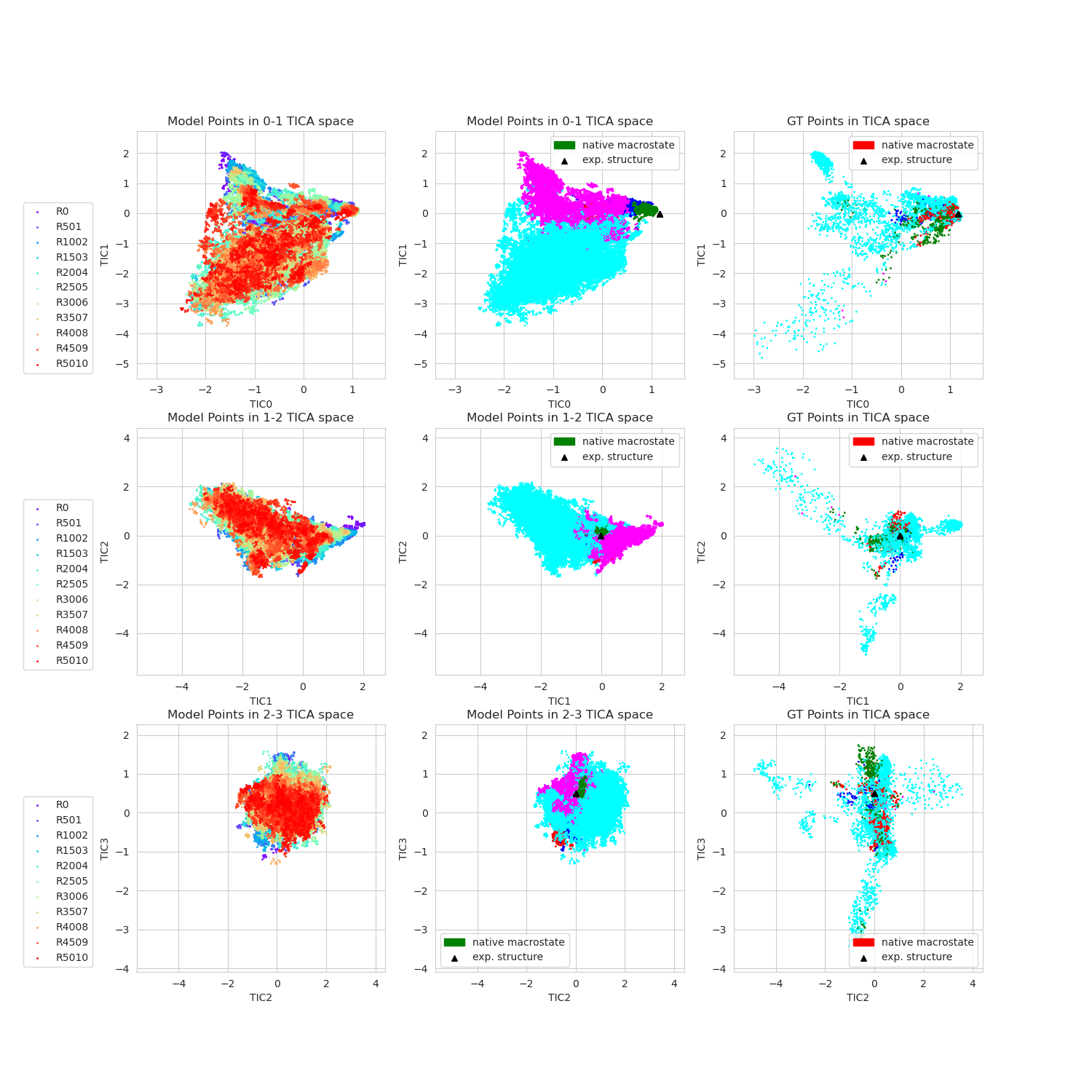}
    \caption{TICA space projections for Protein B - Implicit Solvent All Atom MD}
    \label{fgr:tica_spaces_proteinb}
\end{figure}
\clearpage
\begin{table}[h!]
\centering
\resizebox{\textwidth}{!}{%
\begin{tabular}{|l|c|c|c|c|c|c|c|c|}
\hline
\textbf{Metric} & \textbf{TIC 0} & \textbf{TIC 1} & \textbf{TIC 2} & \textbf{TIC 3} & \textbf{Bonds} & \textbf{Angles} & \textbf{Dihedrals} & \textbf{Gyration} \\
\hline
KL & 0.5772 & 0.5385 & 0.7651 & 1.0011 & 0.0111 & 0.1242 & 0.0112 & 0.7586 \\
W1 & 0.6635 & 0.6575 & 0.2576 & 0.3851 & 0.0008 & 0.1308 & 0.2180 & 0.4018 \\
\hline
\end{tabular}
}
\caption{KL and W1 metrics for Protein B with Implicit All Atom MD}
\end{table}

\clearpage

\subsubsection{Protein G}

\begin{figure}[htbp]
    \centering
    \includegraphics[trim={0 0 0 0.75cm},clip,width=0.5\linewidth]{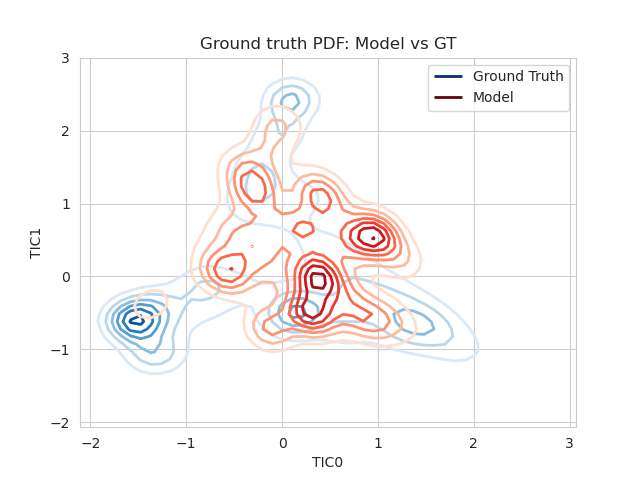}
    \caption{TICA contours for Protein G - Implicit Solvent All Atom MD}
    \label{fgr:tica_contours_proteing}
\end{figure}

\begin{figure}[htbp]
    \centering
    \includegraphics[trim={0 0 0 0.75cm},clip,width=0.5\linewidth]{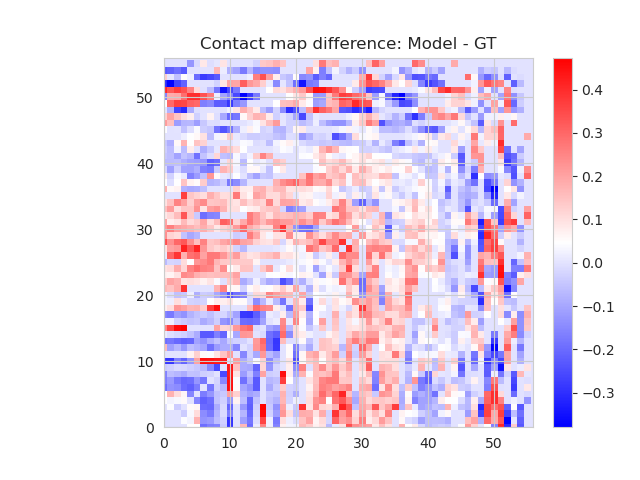}
    \caption{Contact map for Protein G - Implicit Solvent All Atom MD}
\end{figure}

\begin{figure}[htbp]
    \centering
    \includegraphics[trim={0 0 0 0cm},clip,width=\linewidth]{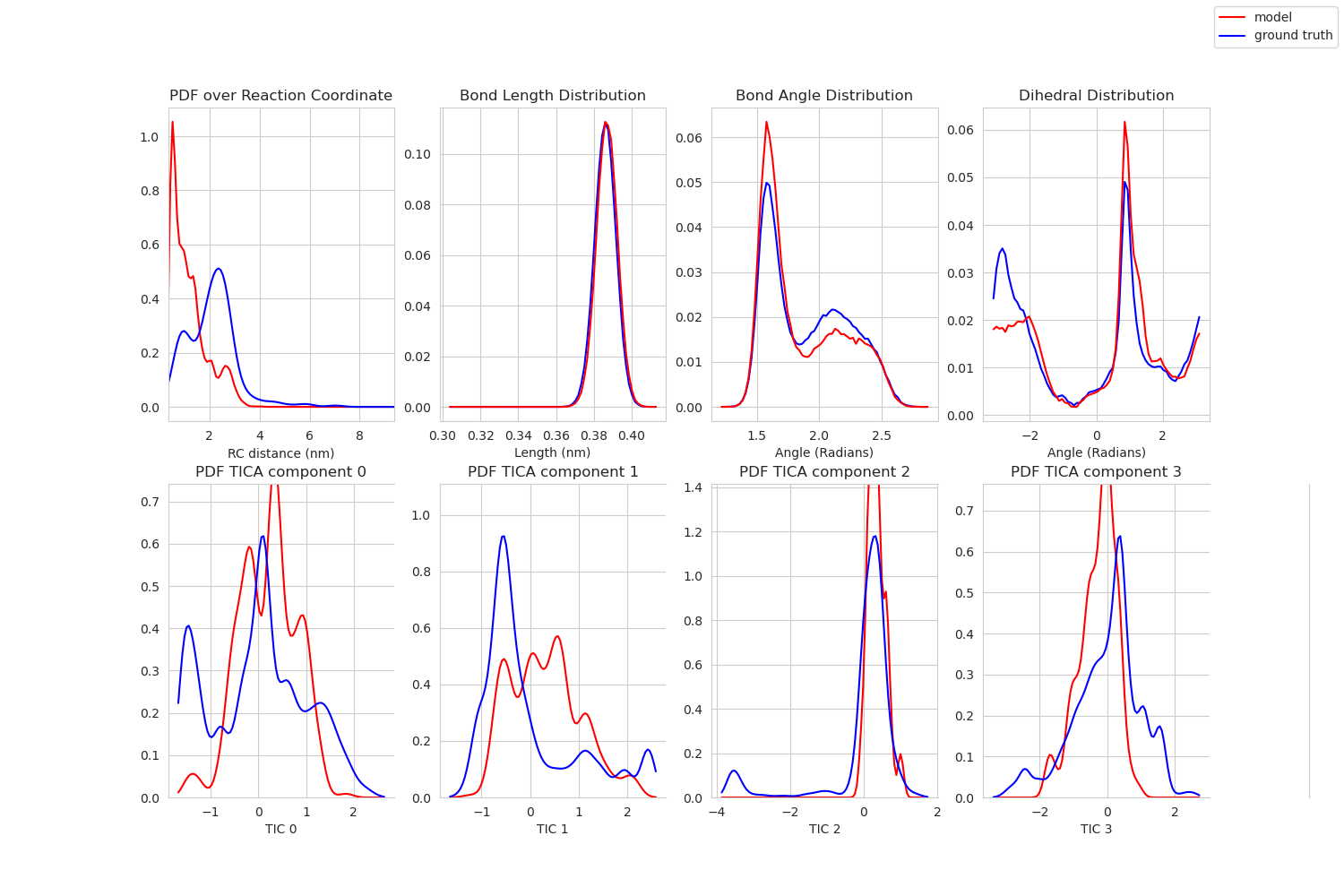}
    \caption{PDFs for Protein G - Implicit Solvent All Atom MD}
    \label{fgr:pdfs_proteing}
\end{figure}

\begin{figure}[htbp]
    \centering
    \includegraphics[trim={0 0 0 0},clip,width=\linewidth]{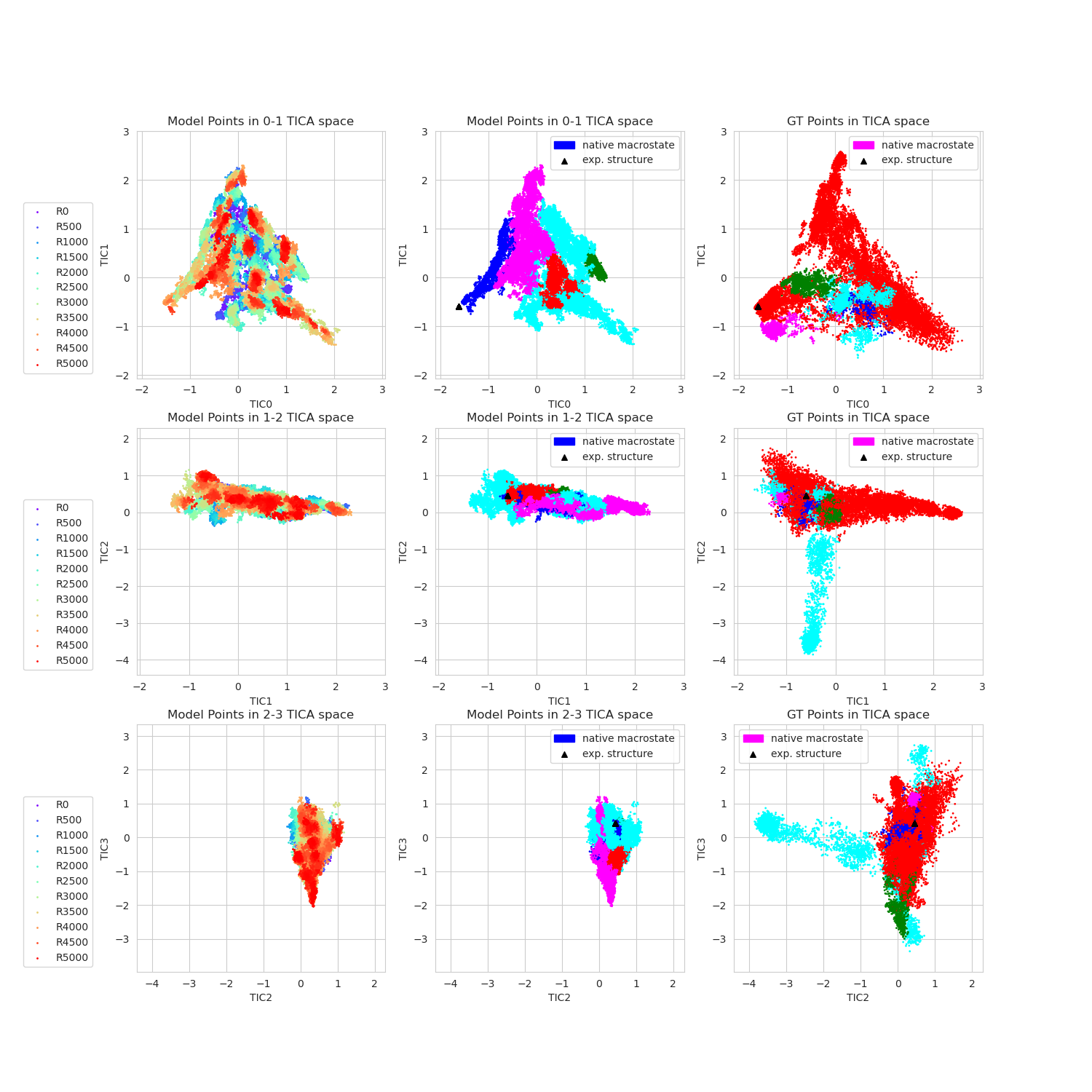}
    \caption{TICA space projections for Protein G - Implicit Solvent All Atom MD}
    \label{fgr:tica_spaces_proteing}
\end{figure}
\clearpage
\begin{table}[h!]
\centering
\resizebox{\textwidth}{!}{%
\begin{tabular}{|l|c|c|c|c|c|c|c|c|}
\hline
\textbf{Metric} & \textbf{TIC 0} & \textbf{TIC 1} & \textbf{TIC 2} & \textbf{TIC 3} & \textbf{Bonds} & \textbf{Angles} & \textbf{Dihedrals} & \textbf{Gyration} \\
\hline
KL & 0.6085 & 0.5601 & 1.1166 & 1.5702 & 0.0087 & 0.0350 & 0.0447 & 0.8245 \\
W1 & 0.3543 & 0.4543 & 0.3587 & 0.4238 & 0.0007 & 0.0519 & 0.2997 & 0.0860 \\
\hline
\end{tabular}
}
\caption{KL and W1 metrics for Protein G with Implicit All Atom MD}
\end{table}

\clearpage

\subsubsection{Trp-cage}

\begin{figure}[htbp]
    \centering
    \includegraphics[trim={0 0 0 0.75cm},clip,width=0.5\linewidth]{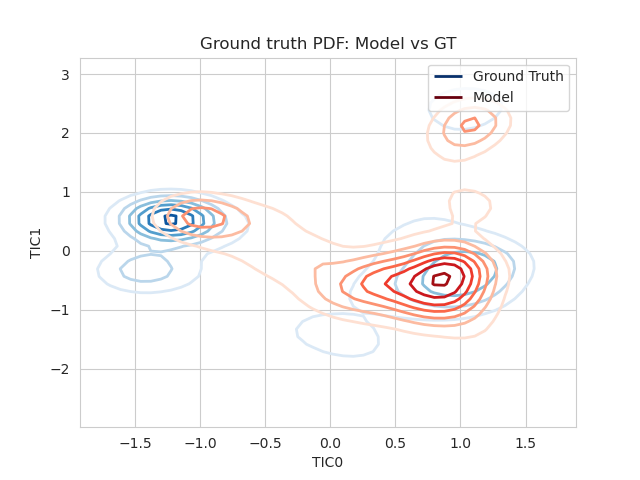}
    \caption{TICA contours for Trp-cage - Implicit Solvent All Atom MD}
    \label{fgr:tica_contours_trpcage}
\end{figure}

\begin{figure}[htbp]
    \centering
    \includegraphics[trim={0 0 0 0.75cm},clip,width=0.5\linewidth]{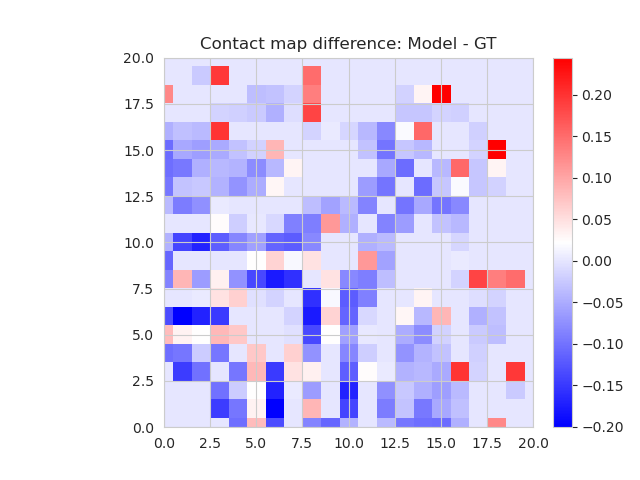}
    \caption{Contact map for Trp-cage - Implicit Solvent All Atom MD}
\end{figure}

\begin{figure}[htbp]
    \centering
    \includegraphics[trim={0 0 0 0cm},clip,width=\linewidth]{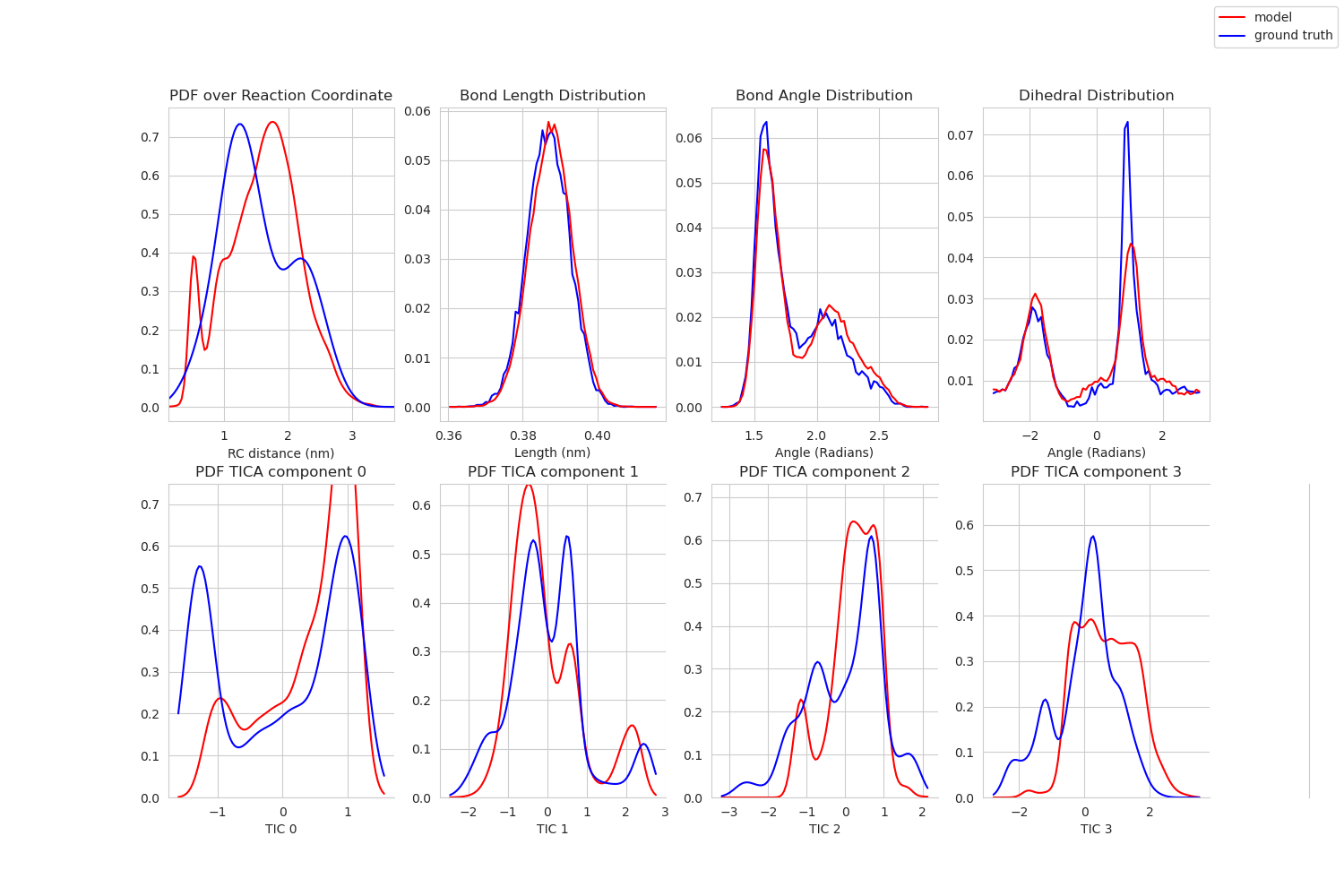}
    \caption{PDFs for Trp-cage - Implicit Solvent All Atom MD}
    \label{fgr:pdfs_trpcage}
\end{figure}

\begin{figure}[htbp]
    \centering
    \includegraphics[trim={0 0 0 0},clip,width=\linewidth]{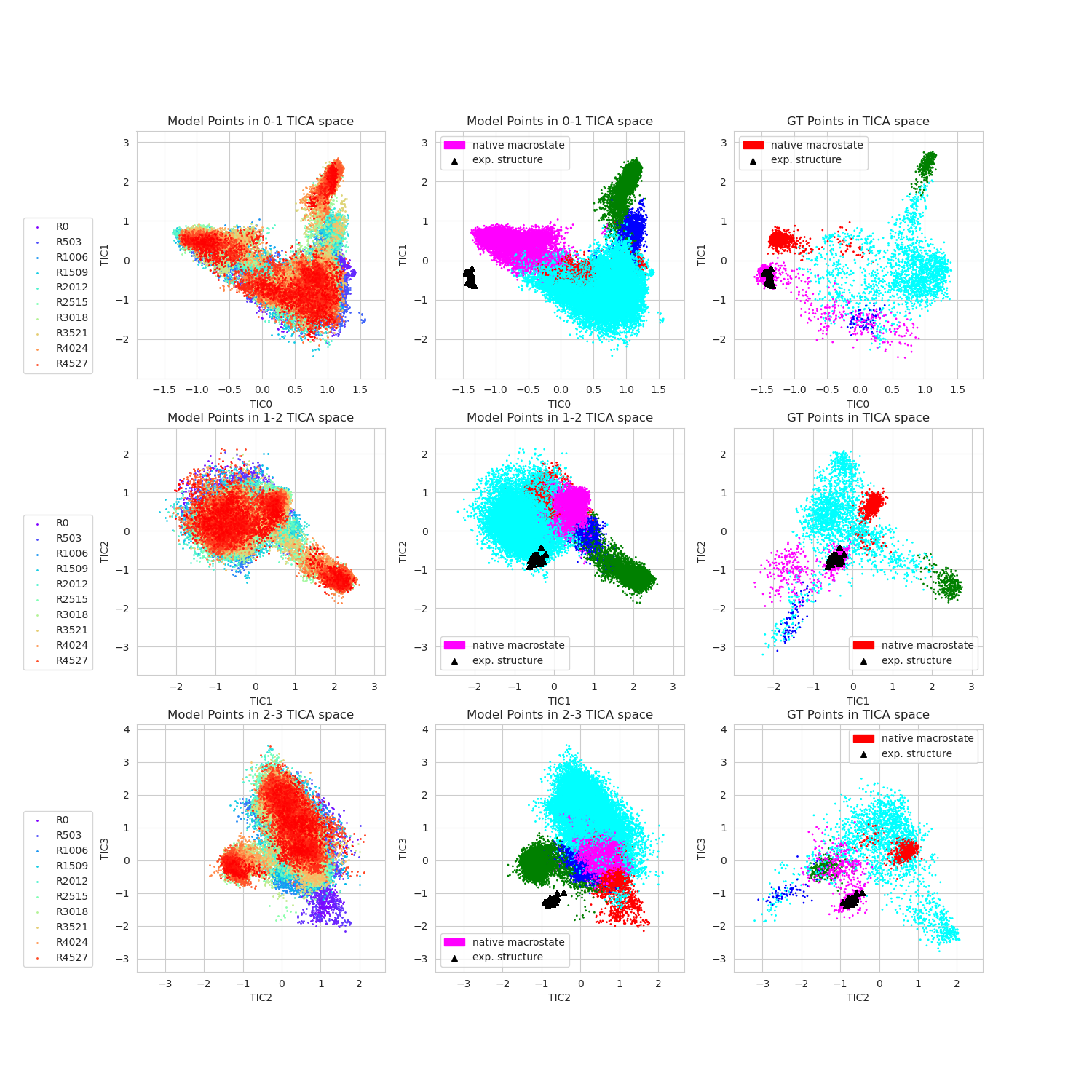}
    \caption{TICA space projections for Trp-cage - Implicit Solvent All Atom MD}
    \label{fgr:tica_spaces_trpcage}
\end{figure}
\clearpage
\begin{table}[h!]
\centering
\resizebox{\textwidth}{!}{%
\begin{tabular}{|l|c|c|c|c|c|c|c|c|}
\hline
\textbf{Metric} & \textbf{TIC 0} & \textbf{TIC 1} & \textbf{TIC 2} & \textbf{TIC 3} & \textbf{Bonds} & \textbf{Angles} & \textbf{Dihedrals} & \textbf{Gyration} \\
\hline
KL & 0.4130 & 0.1685 & 0.6113 & 0.6131 & 0.0110 & 0.0319 & 0.0264 & 0.1554 \\
W1 & 0.4044 & 0.1643 & 0.3060 & 0.6601 & 0.0008 & 0.0317 & 0.0847 & 0.0622 \\
\hline
\end{tabular}
}
\caption{KL and W1 metrics for Trp-cage with Implicit All Atom MD}
\end{table}
\clearpage

\subsubsection{WW Domain}

\begin{figure}[htbp]
    \centering
    \includegraphics[trim={0 0 0 0.75cm},clip,width=0.5\linewidth]{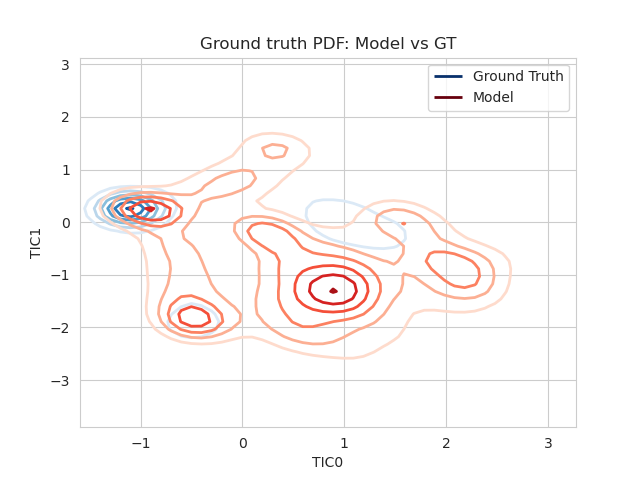}
    \caption{TICA contours for WW Domain - Implicit Solvent All Atom MD}
    \label{fgr:tica_contours_wwdomain}
\end{figure}

\begin{figure}[htbp]
    \centering
    \includegraphics[trim={0 0 0 0.75cm},clip,width=0.5\linewidth]{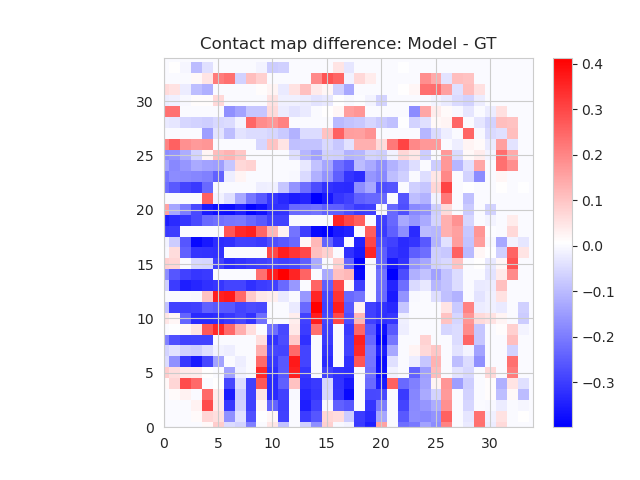}
    \caption{Contact map for WW Domain - Implicit Solvent All Atom MD}
\end{figure}

\begin{figure}[htbp]
    \centering
    \includegraphics[trim={0 0 0 0cm},clip,width=\linewidth]{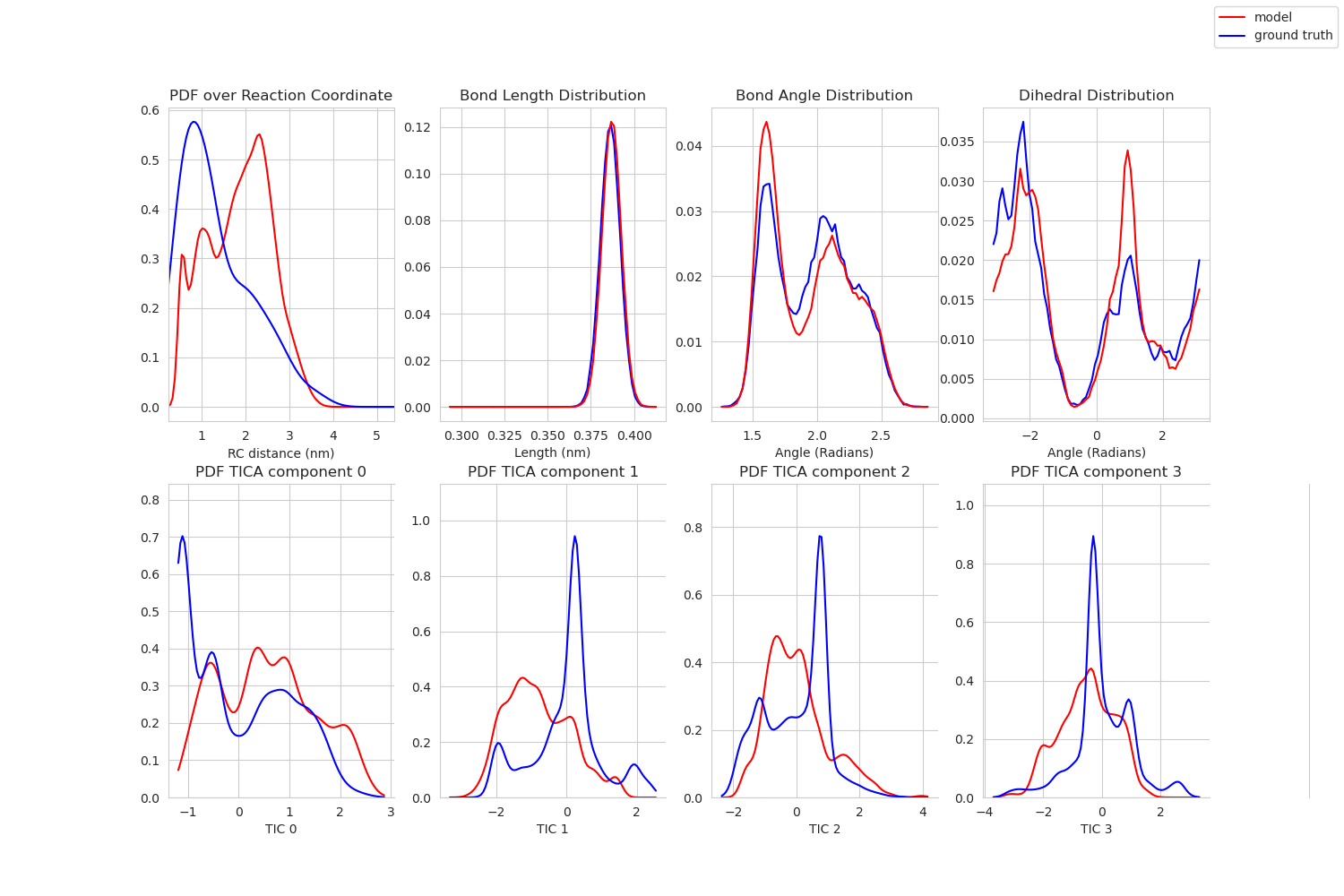}
    \caption{PDFs for WW Domain - Implicit Solvent All Atom MD}
    \label{fgr:pdfs_wwdomain}
\end{figure}

\begin{figure}[htbp]
    \centering
    \includegraphics[trim={0 0 0 0},clip,width=\linewidth]{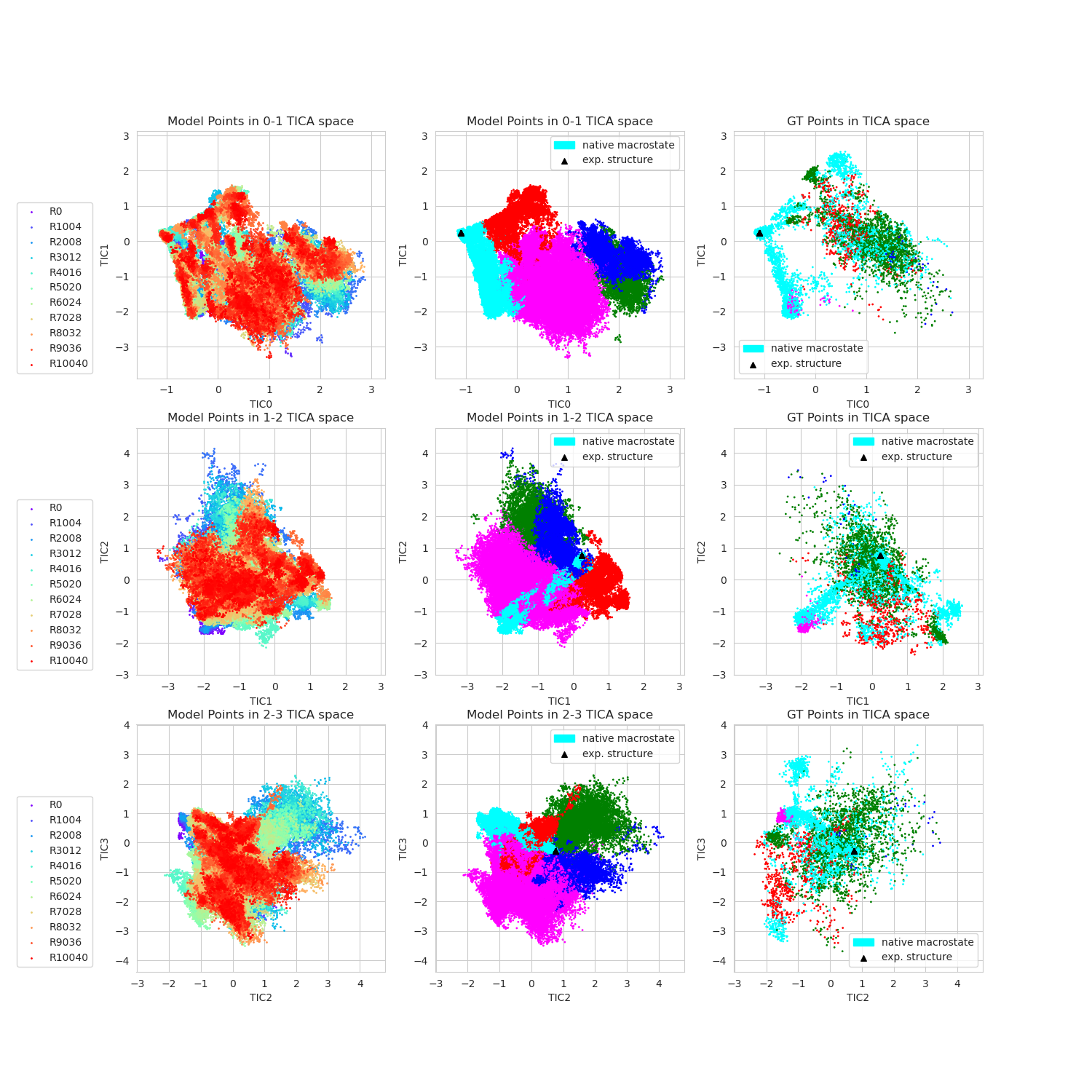}
    \caption{TICA space projections for WW Domain - Implicit Solvent All Atom MD}
    \label{fgr:tica_spaces_wwdomain}
\end{figure}
\clearpage
\begin{table}[h!]
\centering
\resizebox{\textwidth}{!}{%
\begin{tabular}{|l|c|c|c|c|c|c|c|c|}
\hline
\textbf{Metric} & \textbf{TIC 0} & \textbf{TIC 1} & \textbf{TIC 2} & \textbf{TIC 3} & \textbf{Bonds} & \textbf{Angles} & \textbf{Dihedrals} & \textbf{Gyration} \\
\hline
KL & 0.3988 & 0.8827 & 0.4217 & 0.5459 & 0.0130 & 0.0264 & 0.0107 & 0.2359 \\
W1 & 0.4461 & 0.7947 & 0.3019 & 0.5170 & 0.0009 & 0.0320 & 0.1700 & 0.1226 \\
\hline
\end{tabular}
}
\caption{KL and W1 metrics for WWdomain with Implicit All Atom MD}
\end{table}
\clearpage

\subsection{Full Benchmark Diagrams - CGSchNet Fully Trained Model}

\subsubsection{A3D}

\begin{figure}[htbp]
    \centering
    \includegraphics[trim={0 0 0 0.75cm},clip,width=0.5\linewidth]{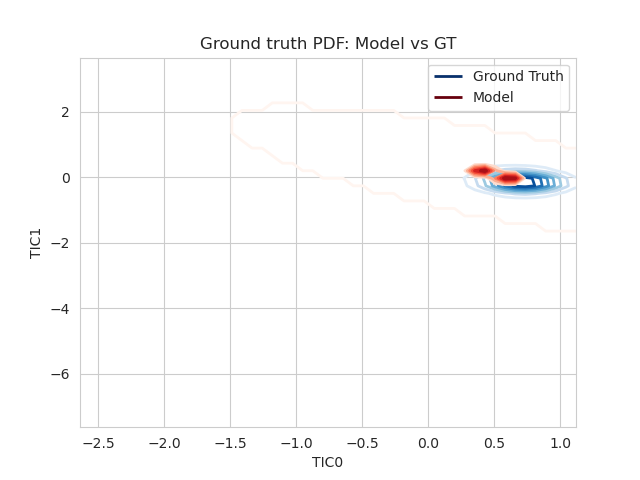}
    \caption{TICA contours for A3D - CGSchNet Fully Trained Model}
    \label{fgr:tica_contours_a3d}
\end{figure}

\begin{figure}[htbp]
    \centering
    \includegraphics[trim={0 0 0 0.75cm},clip,width=0.5\linewidth]{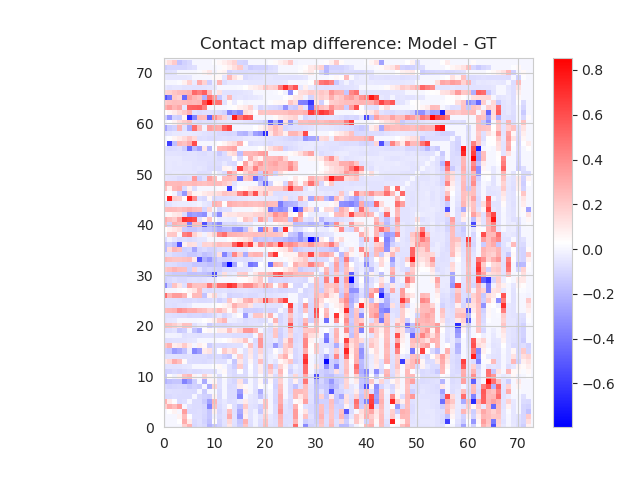}
    \caption{Contact map for A3D - CGSchNet Fully Trained Model}
\end{figure}

\begin{figure}[htbp]
    \centering
    \includegraphics[trim={0 0 0 0cm},clip,width=\linewidth]{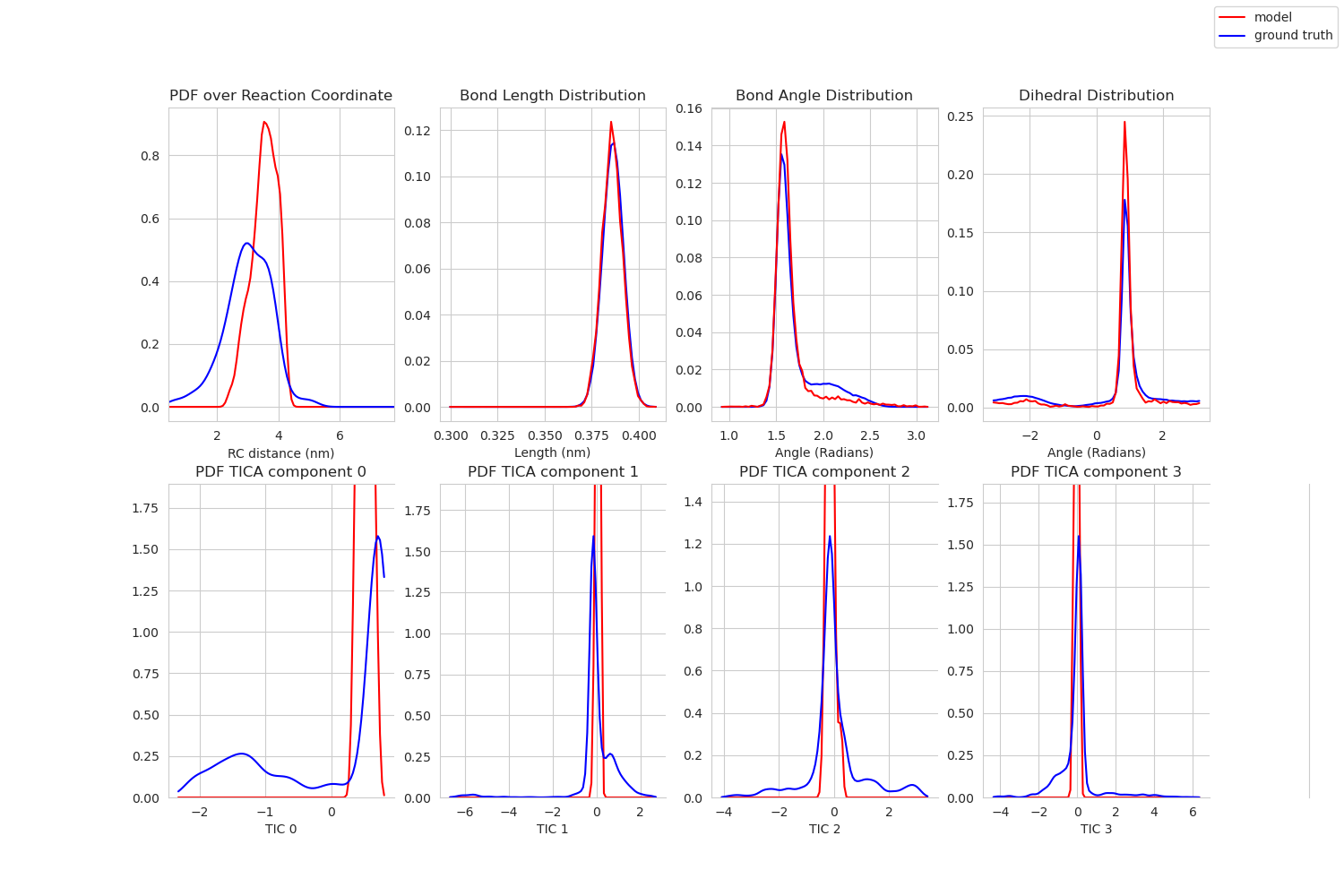}
    \caption{PDFs for A3D - CGSchNet Fully Trained Model}
    \label{fgr:pdfs_a3d}
\end{figure}

\begin{figure}[htbp]
    \centering
    \includegraphics[trim={0 0 0 0},clip,width=\linewidth]{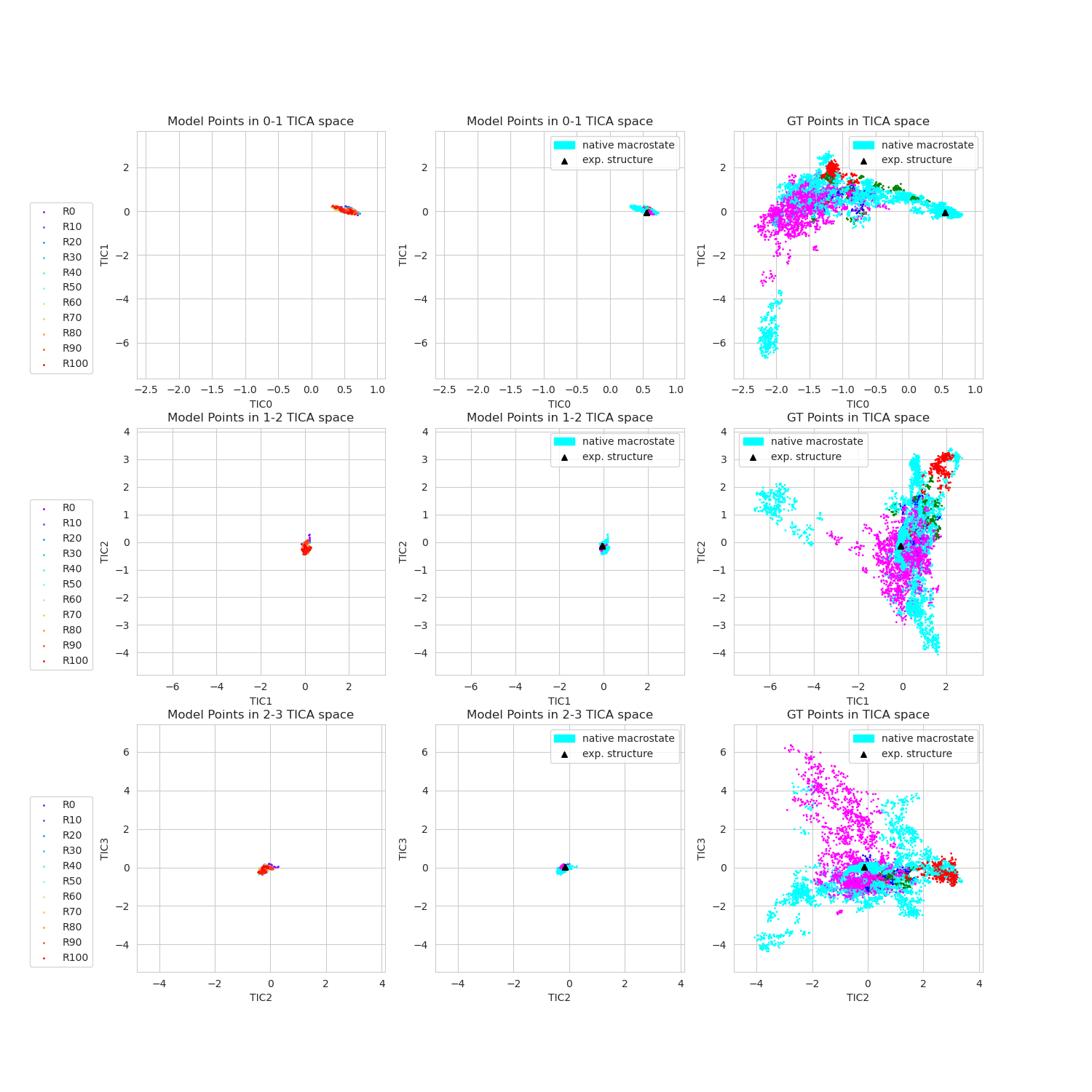}
    \caption{TICA space projections for A3D - CGSchNet Fully Trained Model}
    \label{fgr:tica_spaces_a3d}
\end{figure}
\clearpage
\begin{table}[h!]
\centering
\resizebox{\textwidth}{!}{%
\begin{tabular}{|l|c|c|c|c|c|c|c|c|}
\hline
\textbf{Metric} & \textbf{TIC 0} & \textbf{TIC 1} & \textbf{TIC 2} & \textbf{TIC 3} & \textbf{Bonds} & \textbf{Angles} & \textbf{Dihedrals} & \textbf{Gyration} \\
\hline
KL & 4.8922 & 3.8715 & 2.8953 & 2.8212 & 0.0048 & 0.0851 & 0.1601 & 2.9601 \\
W1 & 0.7346 & 0.4382 & 0.4880 & 0.4373 & 0.0004 & 0.0529 & 0.1107 & 0.1115 \\
\hline
\end{tabular}
}
\caption{KL and W1 metrics for a3D with the CGSchNet Fully Trained Model}
\end{table}

\clearpage

\subsubsection{BBA}

\begin{figure}[htbp]
    \centering
    \includegraphics[trim={0 0 0 0.75cm},clip,width=0.5\linewidth]{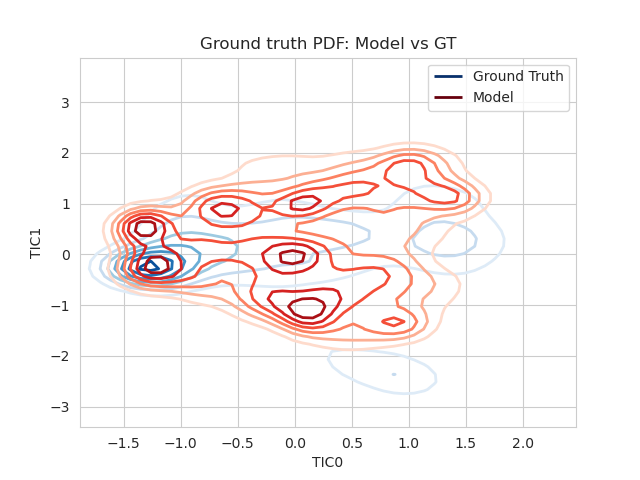}
    \caption{TICA contours for BBA - CGSchNet Fully Trained Model}
    \label{fgr:tica_contours_bba}
\end{figure}

\begin{figure}[htbp]
    \centering
    \includegraphics[trim={0 0 0 0.75cm},clip,width=0.5\linewidth]{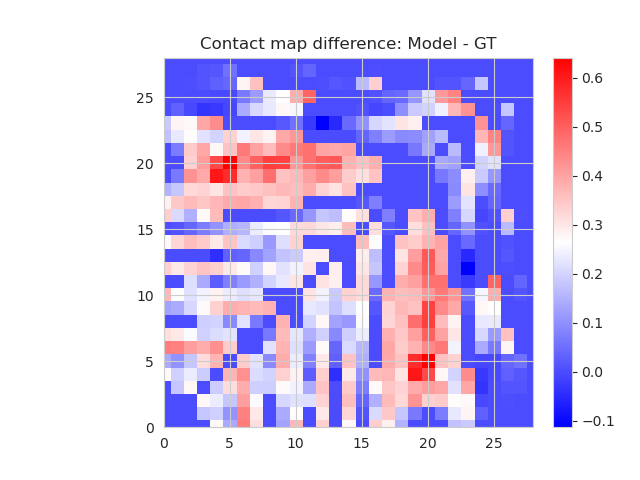}
    \caption{Contact map for BBA - CGSchNet Fully Trained Model}
\end{figure}

\begin{figure}[htbp]
    \centering
    \includegraphics[trim={0 0 0 0cm},clip,width=\linewidth]{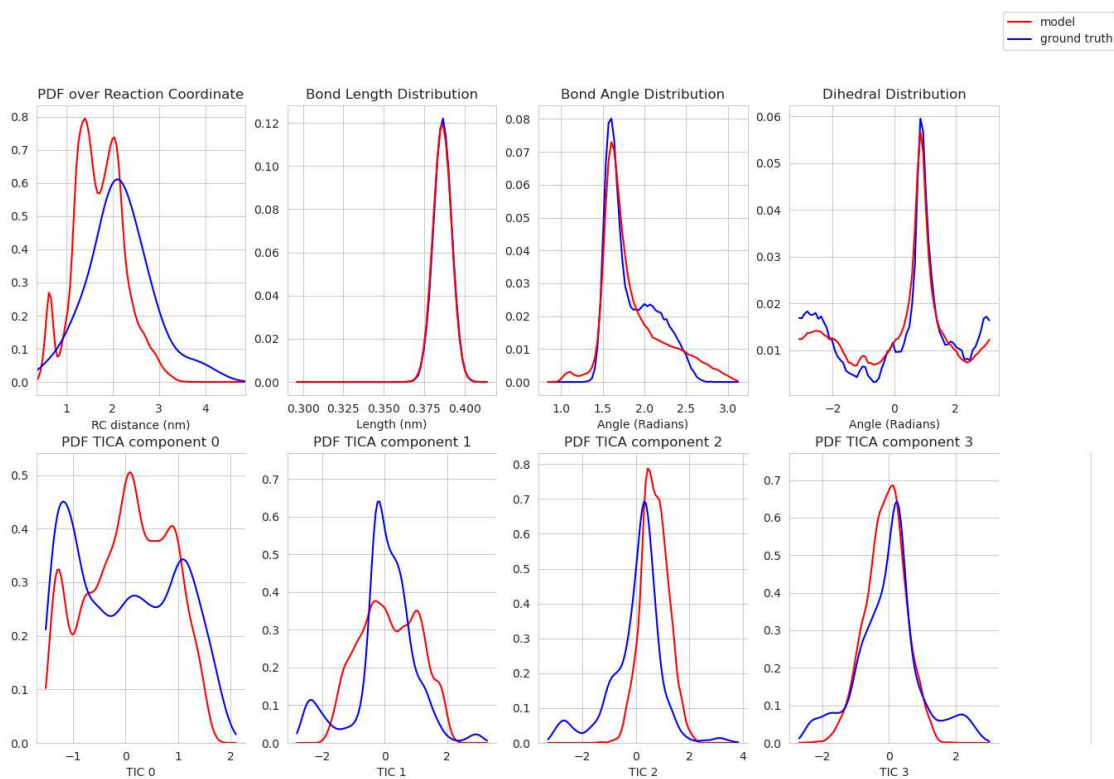}
    \caption{PDFs for BBA - CGSchNet Fully Trained Model}
    \label{fgr:pdfs_bba}
\end{figure}

\begin{figure}[htbp]
    \centering
    \includegraphics[trim={0 0 0 0},clip,width=\linewidth]{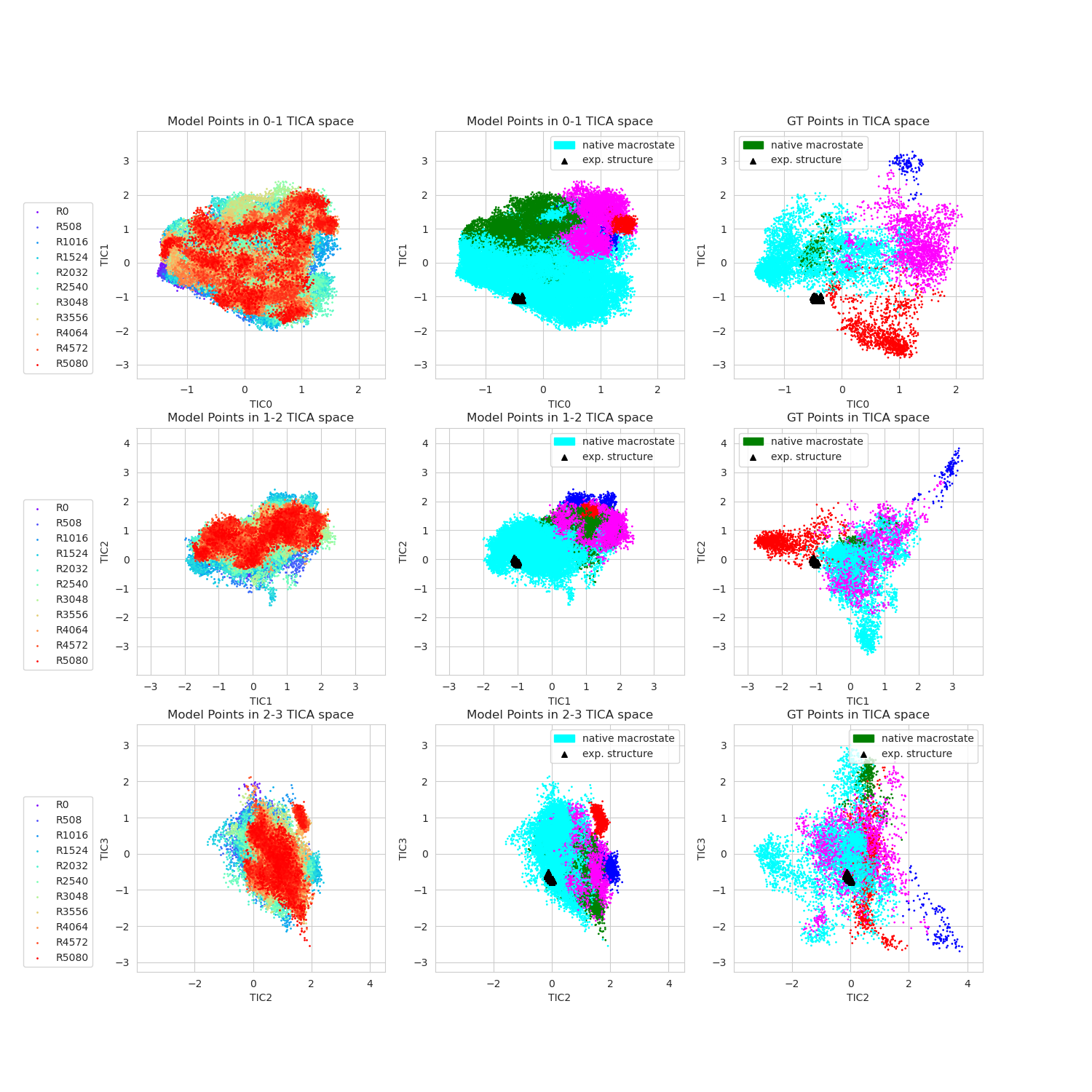}
    \caption{TICA space projections for BBA - CGSchNet Fully Trained Model}
    \label{fgr:tica_spaces_bba}
\end{figure}
\clearpage
\begin{table}[h!]
\centering
\resizebox{\textwidth}{!}{%
\begin{tabular}{|l|c|c|c|c|c|c|c|c|}
\hline
\textbf{Metric} & \textbf{TIC 0} & \textbf{TIC 1} & \textbf{TIC 2} & \textbf{TIC 3} & \textbf{Bonds} & \textbf{Angles} & \textbf{Dihedrals} & \textbf{Gyration} \\
\hline
KL & 0.1614 & 0.8866 & 1.3594 & 0.6290 & 0.0086 & 0.1100 & 0.0925 & 3.3238 \\
W1 & 0.2109 & 0.2781 & 0.7300 & 0.2835 & 0.0006 & 0.0441 & 0.3215 & 0.2103 \\
\hline
\end{tabular}
}
\caption{KL and W1 metrics for BBA with the CGSchNet Fully Trained Model}
\end{table}

\clearpage

\subsubsection{Chignolin}

\begin{figure}[htbp]
    \centering
    \includegraphics[trim={0 0 0 0.75cm},clip,width=0.5\linewidth]{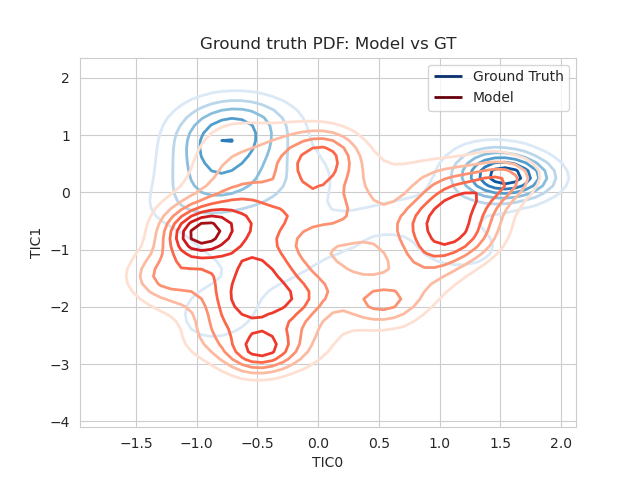}
    \caption{TICA contours for Chignolin - CGSchNet Fully Trained Model}
    \label{fgr:tica_contours_chignolin}
\end{figure}

\begin{figure}[htbp]
    \centering
    \includegraphics[trim={0 0 0 0.75cm},clip,width=0.5\linewidth]{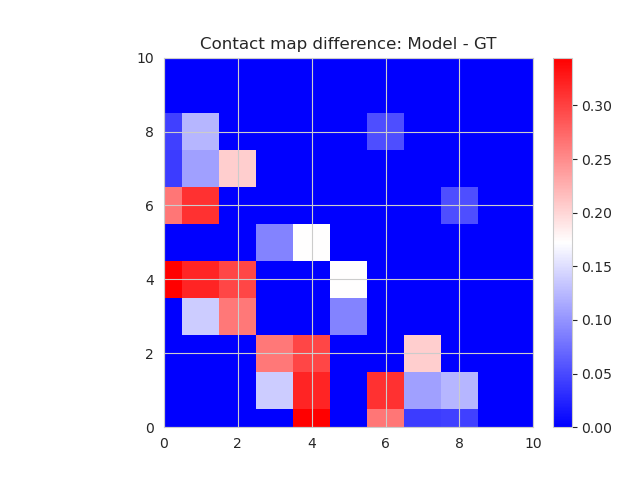}
    \caption{Contact map for Chignolin - CGSchNet Fully Trained Model}
\end{figure}

\begin{figure}[htbp]
    \centering
    \includegraphics[trim={0 0 0 0cm},clip,width=\linewidth]{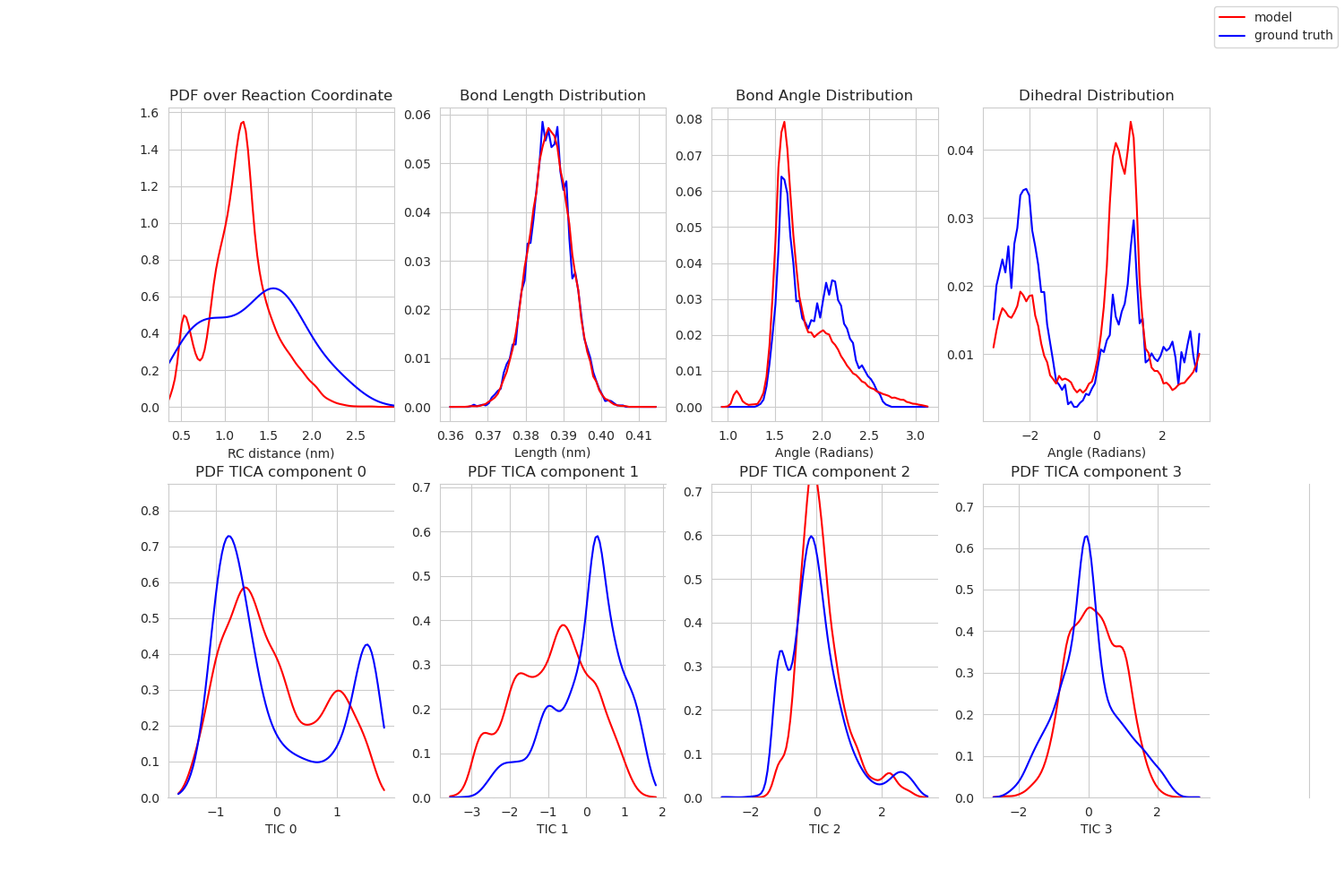}
    \caption{PDFs for Chignolin - CGSchNet Fully Trained Model}
    \label{fgr:pdfs_chignolin}
\end{figure}

\begin{figure}[htbp]
    \centering
    \includegraphics[trim={0 0 0 0},clip,width=\linewidth]{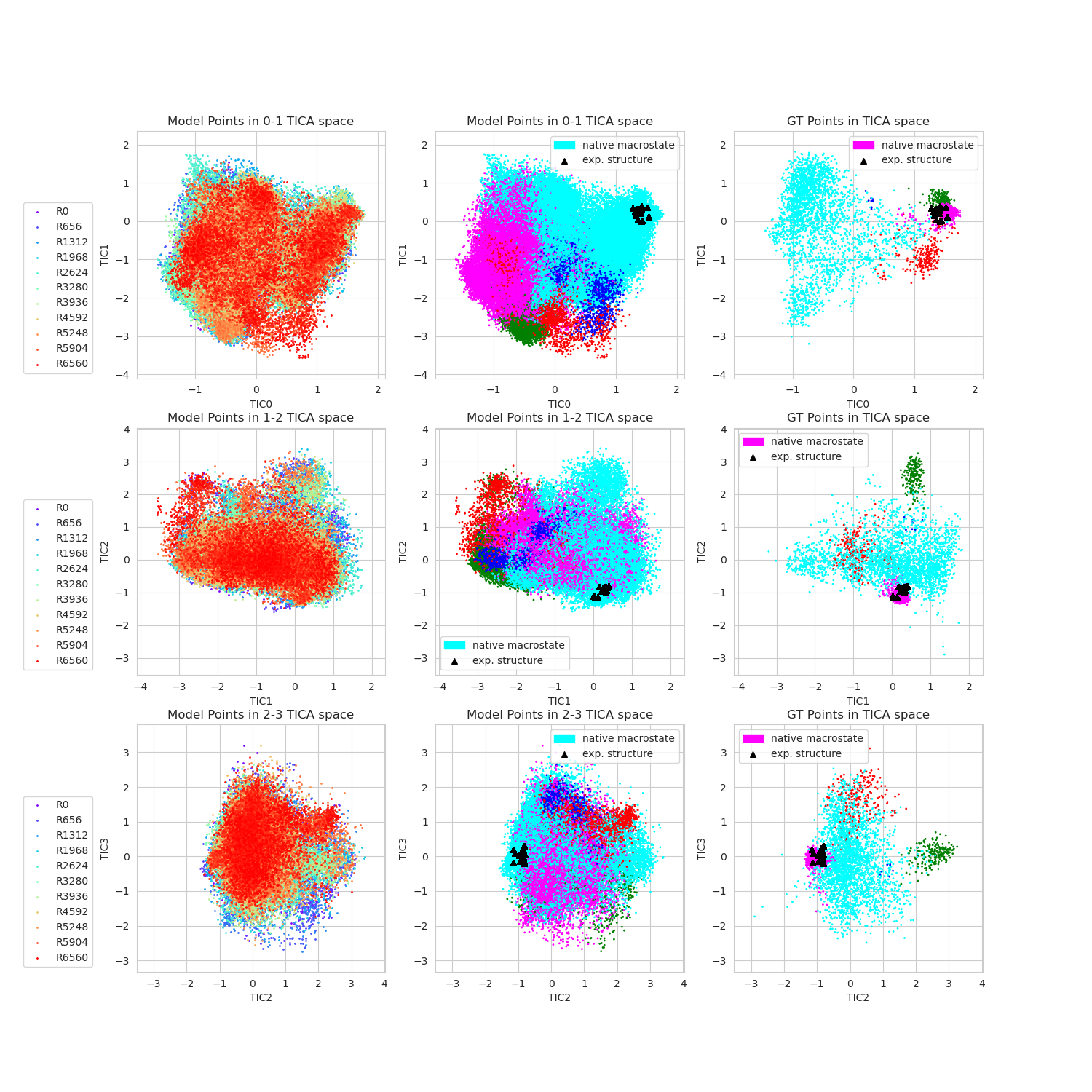}
    \caption{TICA space projections for Chignolin - CGSchNet Fully Trained Model}
    \label{fgr:tica_spaces_chignolin}
\end{figure}
\clearpage

\begin{table}[h!]
\centering
\resizebox{\textwidth}{!}{%
\begin{tabular}{|l|c|c|c|c|c|c|c|c|}
\hline
\textbf{Metric} & \textbf{TIC 0} & \textbf{TIC 1} & \textbf{TIC 2} & \textbf{TIC 3} & \textbf{Bonds} & \textbf{Angles} & \textbf{Dihedrals} & \textbf{Gyration} \\
\hline
KL & 0.2368 & 0.4806 & 0.1442 & 0.1353 & 0.0033 & 0.1331 & 0.0996 & 0.9233 \\
W1 & 0.1904 & 0.8934 & 0.2074 & 0.2368 & 0.0003 & 0.0871 & 0.5291 & 0.0947 \\
\hline
\end{tabular}
}
\caption{KL and W1 metrics for Chignolin with the CGSchNet Fully Trained Model}
\end{table}

\clearpage

\subsubsection{Homeodomain}

\begin{figure}[htbp]
    \centering
    \includegraphics[trim={0 0 0 0.75cm},clip,width=0.5\linewidth]{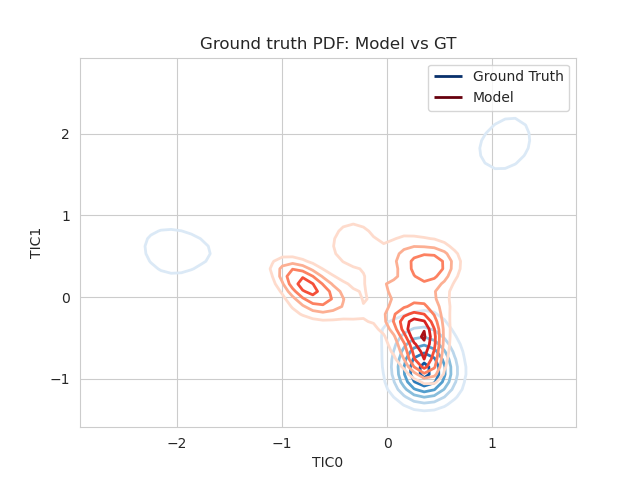}
    \caption{TICA contours for Homeodomain - CGSchNet Fully Trained Model}
    \label{fgr:tica_contours_homeodomain}
\end{figure}

\begin{figure}[htbp]
    \centering
    \includegraphics[trim={0 0 0 0.75cm},clip,width=0.5\linewidth]{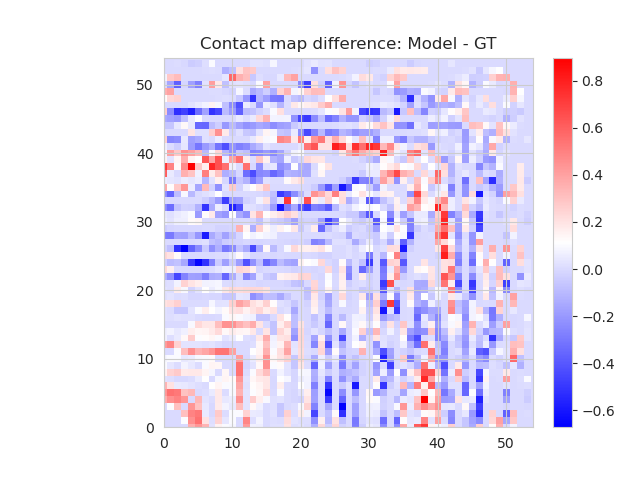}
    \caption{Contact map for Homeodomain - CGSchNet Fully Trained Model}
\end{figure}

\begin{figure}[htbp]
    \centering
    \includegraphics[trim={0 0 0 0cm},clip,width=\linewidth]{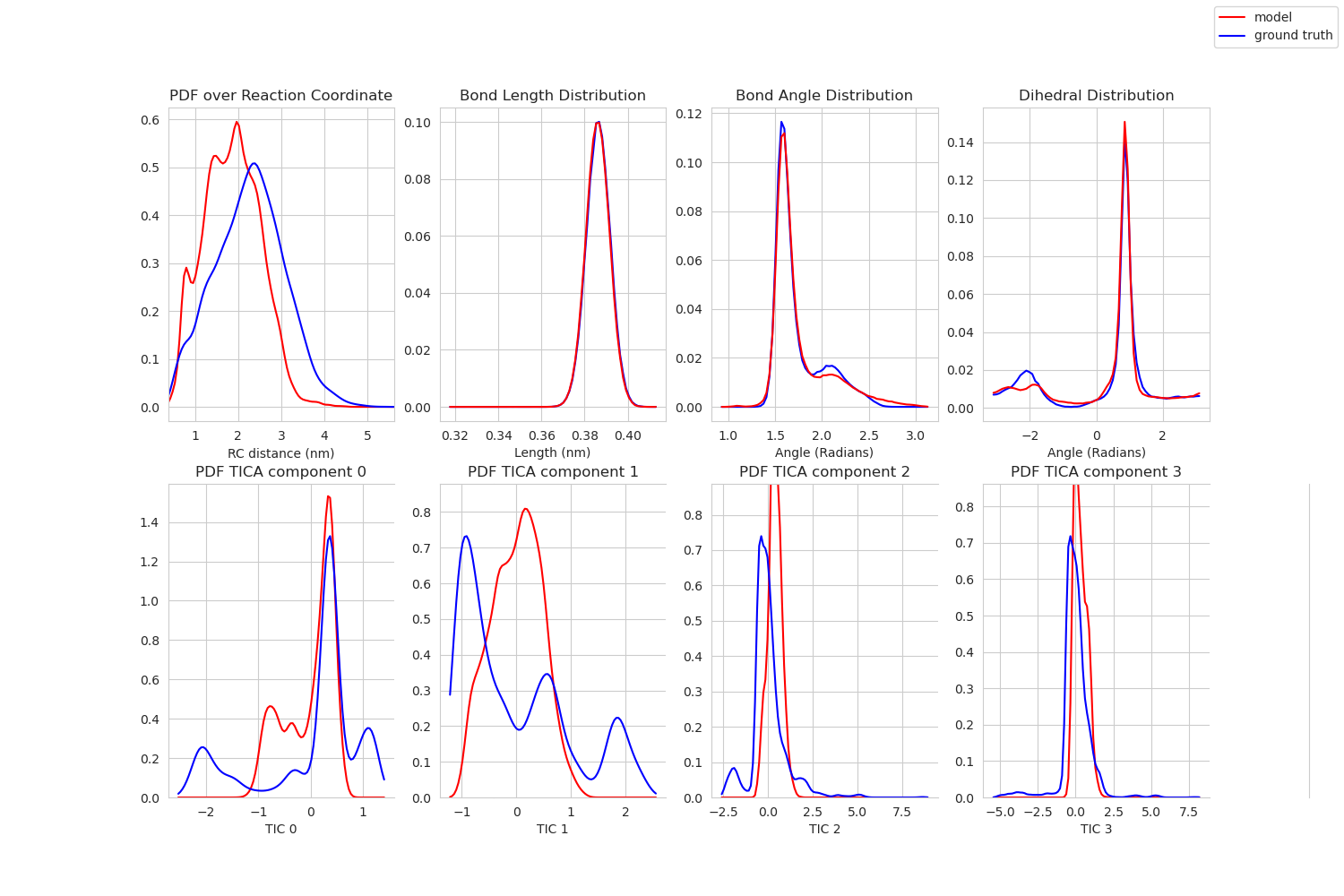}
    \caption{PDFs for Homeodomain - CGSchNet Fully Trained Model}
    \label{fgr:pdfs_homeodomain}
\end{figure}

\begin{figure}[htbp]
    \centering
    \includegraphics[trim={0 0 0 0},clip,width=\linewidth]{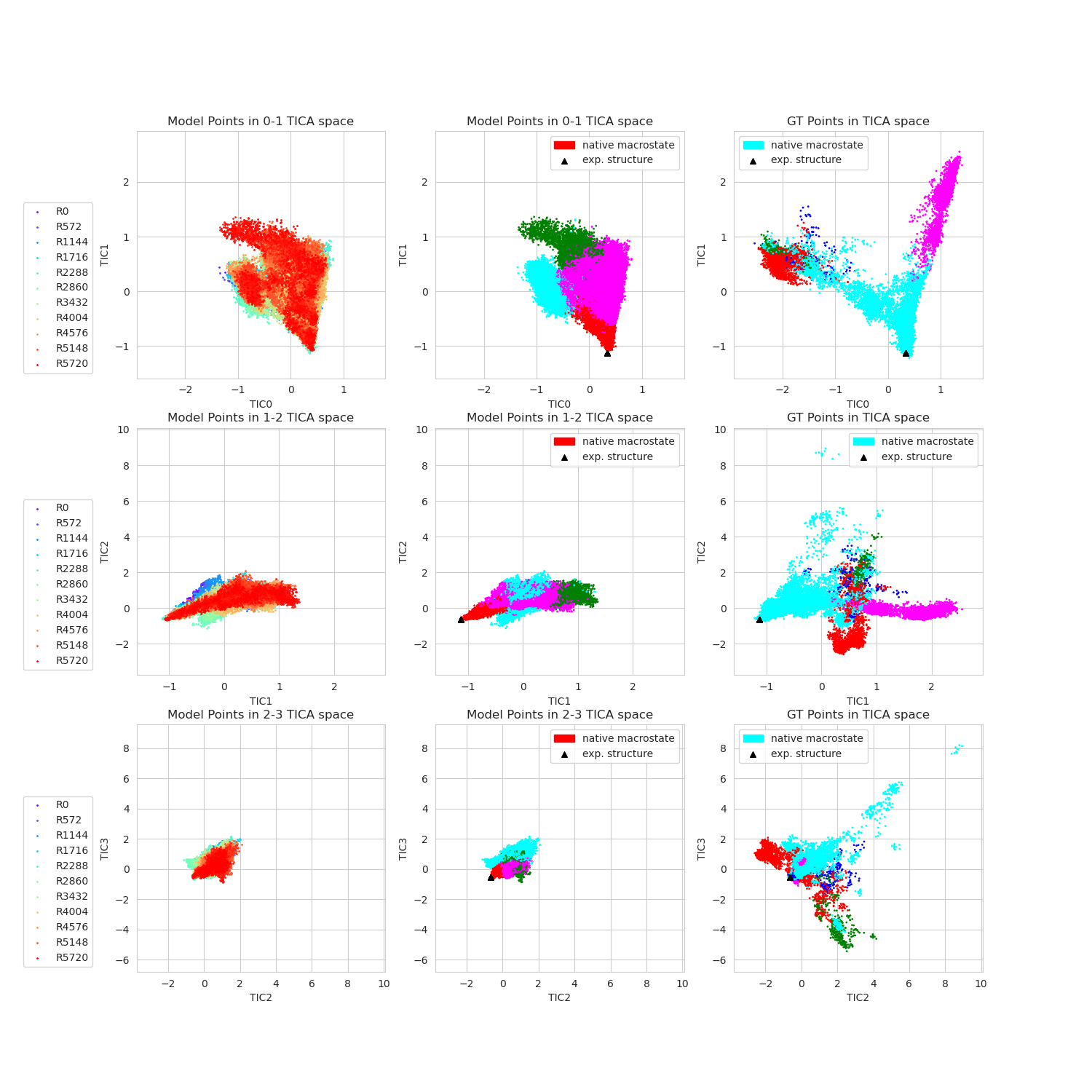}
    \caption{TICA space projections for Homeodomain - CGSchNet Fully Trained Model}
    \label{fgr:tica_spaces_homeodomain}
\end{figure}
\clearpage
\begin{table}[h!]
\centering
\resizebox{\textwidth}{!}{%
\begin{tabular}{|l|c|c|c|c|c|c|c|c|}
\hline
\textbf{Metric} & \textbf{TIC 0} & \textbf{TIC 1} & \textbf{TIC 2} & \textbf{TIC 3} & \textbf{Bonds} & \textbf{Angles} & \textbf{Dihedrals} & \textbf{Gyration} \\
\hline
KL & 3.7665 & 2.1008 & 1.7079 & 1.0527 & 0.0015 & 0.0390 & 0.0552 & 0.2076 \\
W1 & 0.4284 & 0.4722 & 0.5890 & 0.4316 & 0.0002 & 0.0158 & 0.2223 & 0.0200 \\
\hline
\end{tabular}
}
\caption{KL and W1 metrics for Homeodomain with the CGSchNet Fully Trained Model}
\end{table}
\clearpage

\subsubsection{$\lambda$-repressor}

\begin{figure}[htbp]
    \centering
    \includegraphics[trim={0 0 0 0.75cm},clip,width=0.5\linewidth]{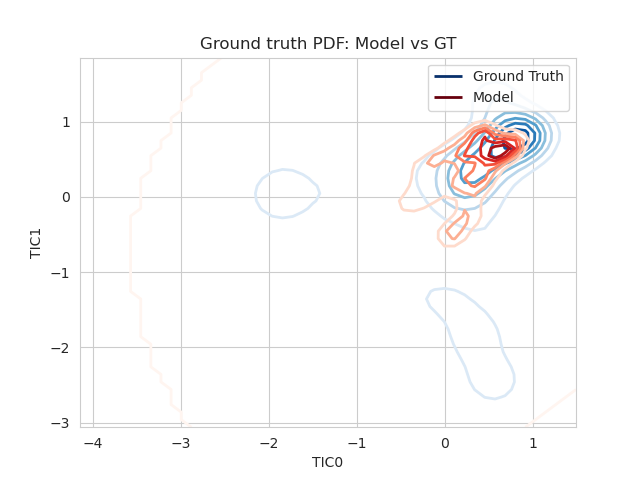}
    \caption{TICA contours for $\lambda$-repressor - CGSchNet Fully Trained Model}
    \label{fgr:tica_contours_lambda}
\end{figure}

\begin{figure}[htbp]
    \centering
    \includegraphics[trim={0 0 0 0.75cm},clip,width=0.5\linewidth]{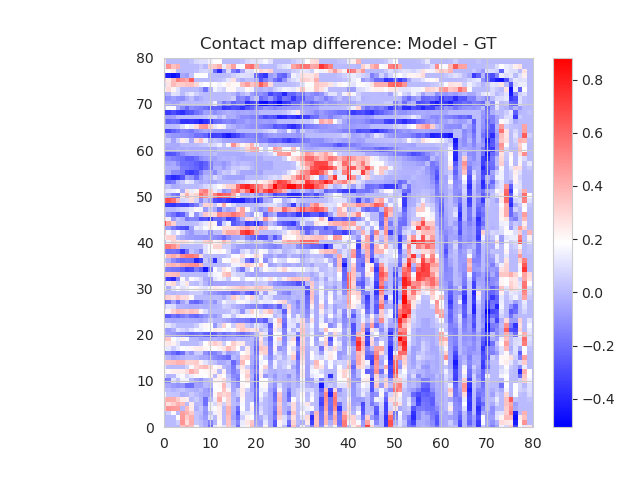}
    \caption{Contact map for $\lambda$-repressor - CGSchNet Fully Trained Model}
\end{figure}

\begin{figure}[htbp]
    \centering
    \includegraphics[trim={0 0 0 0cm},clip,width=\linewidth]{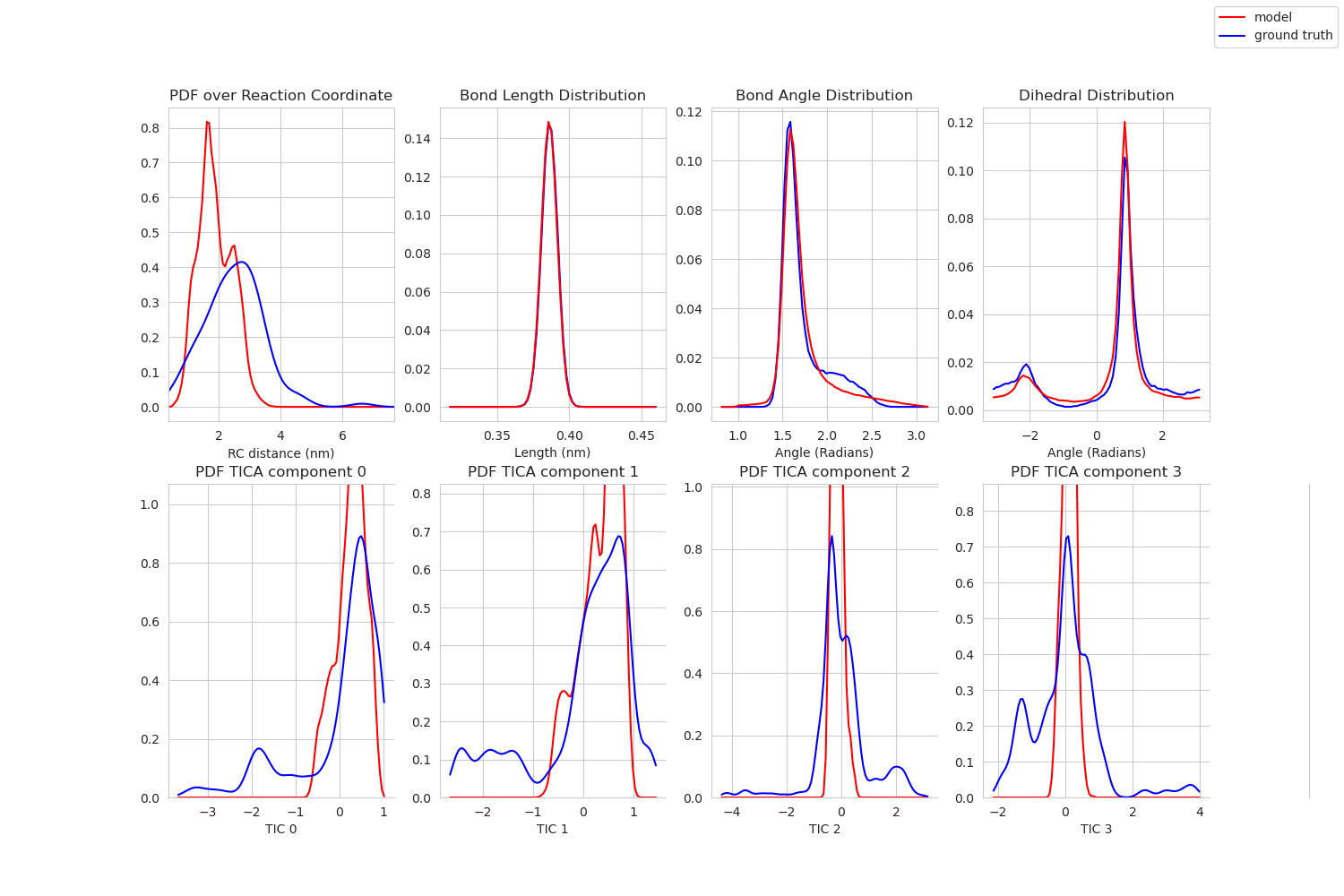}
    \caption{PDFs for $\lambda$-repressor - CGSchNet Fully Trained Model}
    \label{fgr:pdfs_lambda}
\end{figure}

\begin{figure}[htbp]
    \centering
    \includegraphics[trim={0 0 0 0},clip,width=\linewidth]{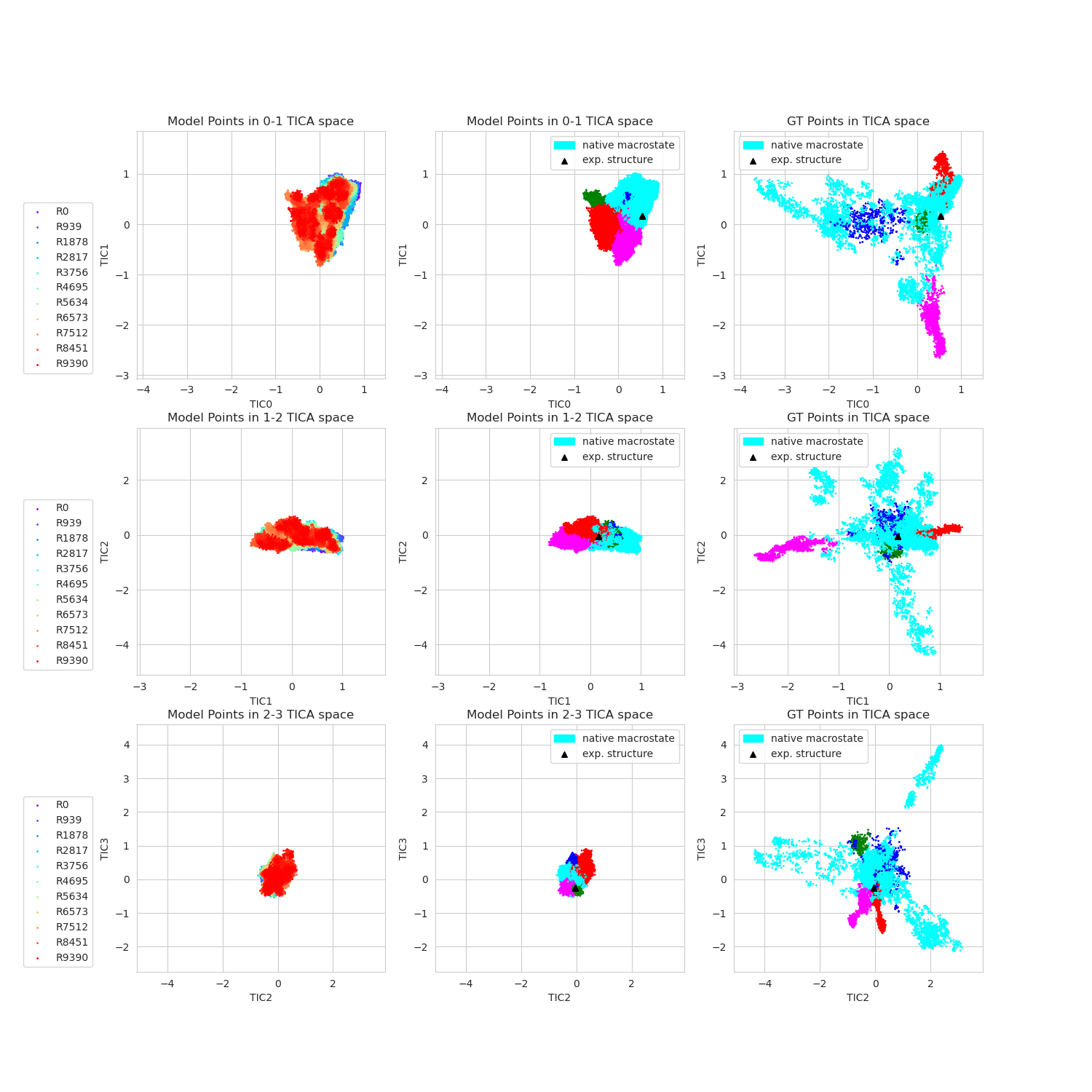}
    \caption{TICA space projections for $\lambda$-repressor - CGSchNet Fully Trained Model}
    \label{fgr:tica_spaces_lambda}
\end{figure}
\clearpage

\begin{table}[h!]
\centering
\resizebox{\textwidth}{!}{%
\begin{tabular}{|l|c|c|c|c|c|c|c|c|}
\hline
\textbf{Metric} & \textbf{TIC 0} & \textbf{TIC 1} & \textbf{TIC 2} & \textbf{TIC 3} & \textbf{Bonds} & \textbf{Angles} & \textbf{Dihedrals} & \textbf{Gyration} \\
\hline
KL & 2.1323 & 2.5487 & 2.8529 & 4.3626 & 0.0019 & 0.0748 & 0.0361 & 0.5359 \\
W1 & 0.4016 & 0.4392 & 0.5014 & 0.5771 & 0.0003 & 0.0569 & 0.1836 & 0.0681 \\
\hline
\end{tabular}
}
\caption{KL and W1 metrics for $\lambda$-repressor with the CGSchNet Fully Trained Model}
\end{table}

\clearpage

\subsubsection{Protein B}

\begin{figure}[htbp]
    \centering
    \includegraphics[trim={0 0 0 0.75cm},clip,width=0.5\linewidth]{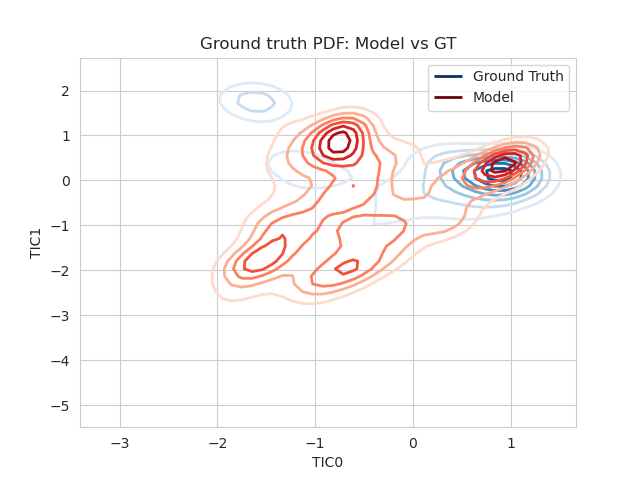}
    \caption{TICA contours for Protein B - CGSchNet Fully Trained Model}
    \label{fgr:tica_contours_proteinb}
\end{figure}

\begin{figure}[htbp]
    \centering
    \includegraphics[trim={0 0 0 0.75cm},clip,width=0.5\linewidth]{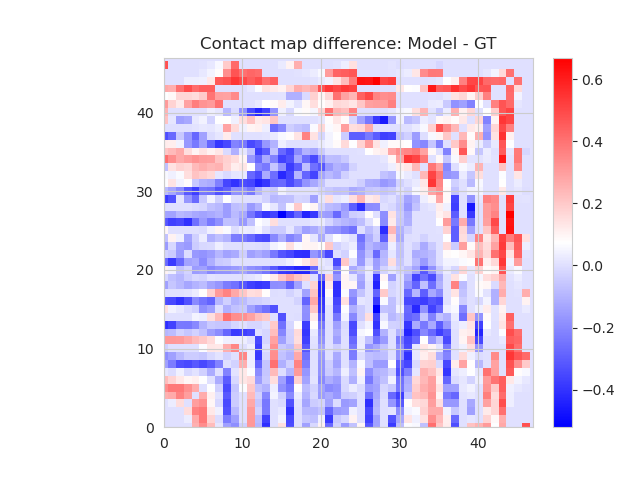}
    \caption{Contact map for Protein B - CGSchNet Fully Trained Model}
\end{figure}

\begin{figure}[htbp]
    \centering
    \includegraphics[trim={0 0 0 0cm},clip,width=\linewidth]{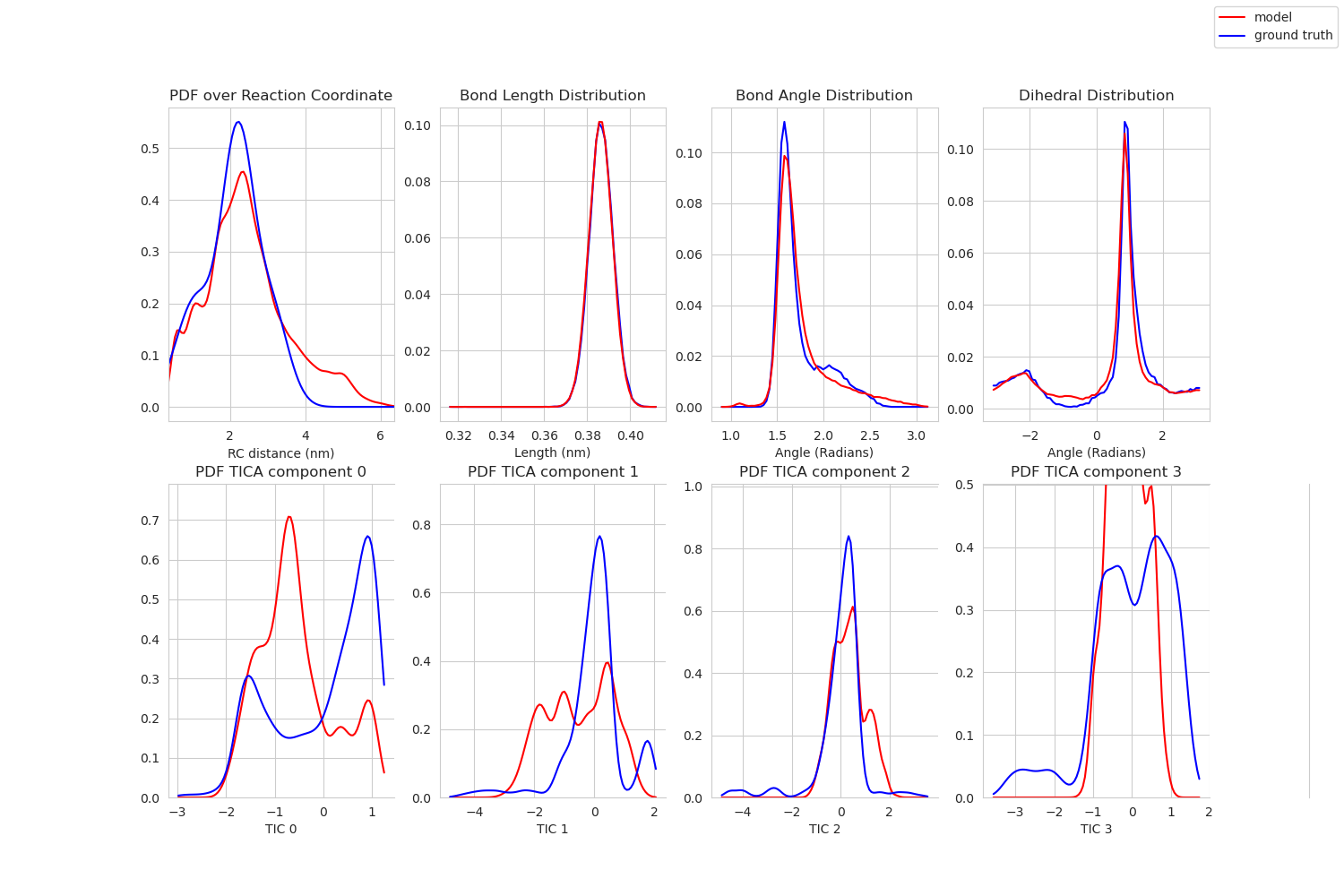}
    \caption{PDFs for Protein B - CGSchNet Fully Trained Model}
    \label{fgr:pdfs_proteinb}
\end{figure}

\begin{figure}[htbp]
    \centering
    \includegraphics[trim={0 0 0 0},clip,width=\linewidth]{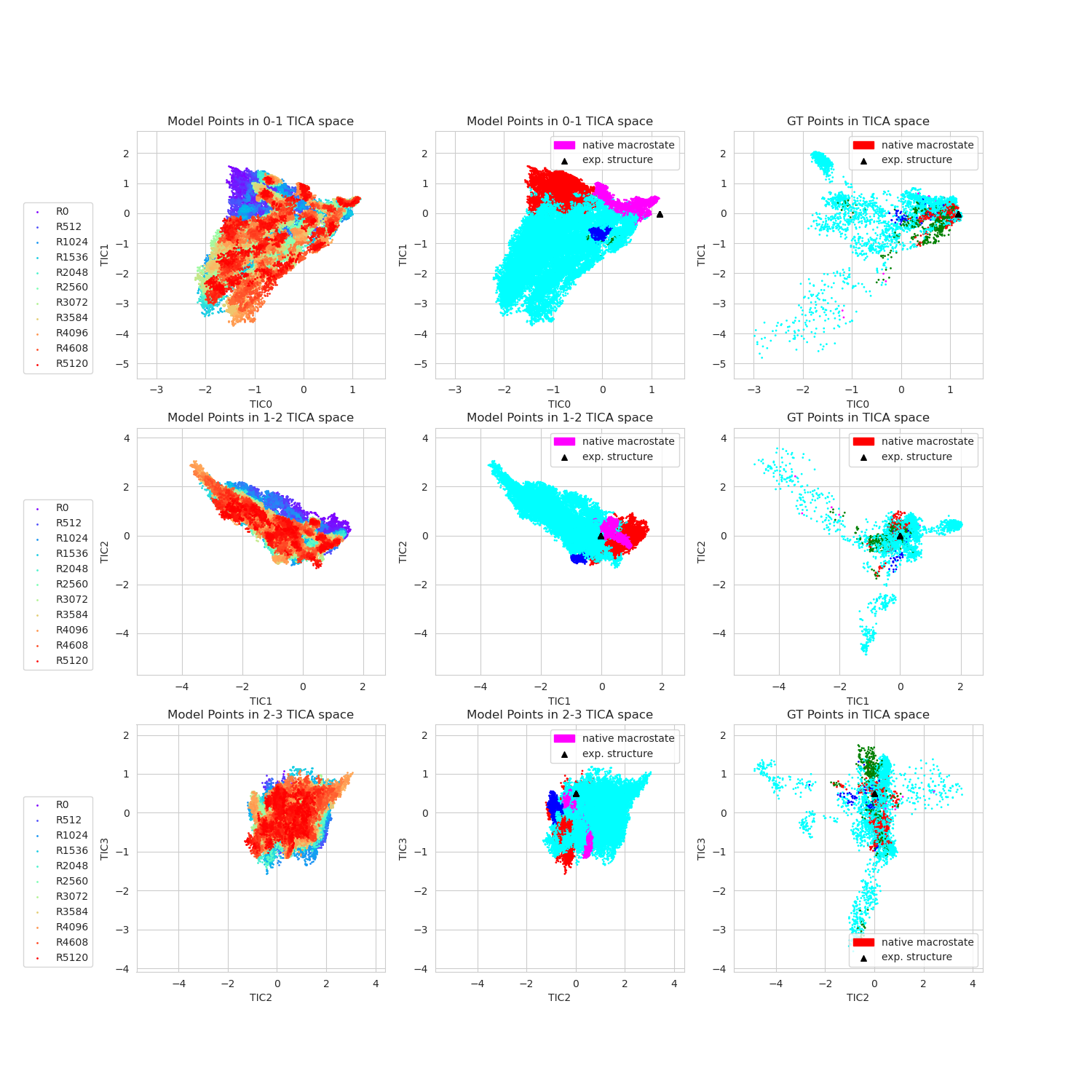}
    \caption{TICA space projections for Protein B - CGSchNet Fully Trained Model}
    \label{fgr:tica_spaces_proteinb}
\end{figure}
\clearpage

\begin{table}[h!]
\centering
\resizebox{\textwidth}{!}{%
\begin{tabular}{|l|c|c|c|c|c|c|c|c|}
\hline
\textbf{Metric} & \textbf{TIC 0} & \textbf{TIC 1} & \textbf{TIC 2} & \textbf{TIC 3} & \textbf{Bonds} & \textbf{Angles} & \textbf{Dihedrals} & \textbf{Gyration} \\
\hline
KL & 0.4294 & 0.7843 & 0.7043 & 1.6948 & 0.0008 & 0.4226 & 0.0644 & 1.0199 \\
W1 & 0.5474 & 0.5817 & 0.3853 & 0.4518 & 0.0000 & 0.2287 & 0.1878 & 0.1688 \\
\hline
\end{tabular}
}
\caption{KL and W1 metrics for Protein B with the CGSchNet Fully Trained Model}
\end{table}

\clearpage

\subsubsection{Protein G}

\begin{figure}[htbp]
    \centering
    \includegraphics[trim={0 0 0 0.75cm},clip,width=0.5\linewidth]{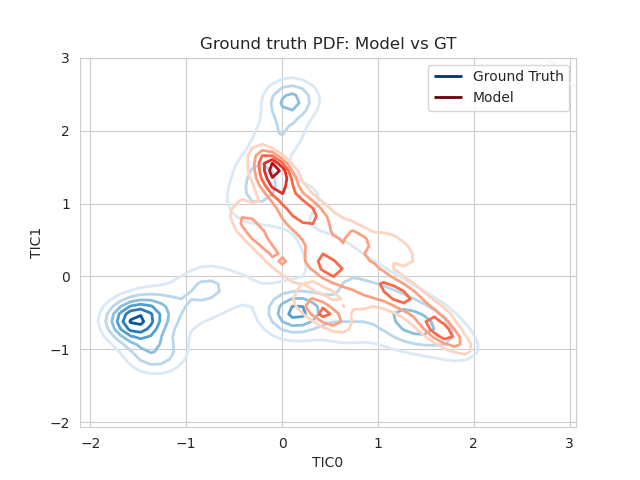}
    \caption{TICA contours for Protein G - CGSchNet Fully Trained Model}
    \label{fgr:tica_contours_proteing}
\end{figure}

\begin{figure}[htbp]
    \centering
    \includegraphics[trim={0 0 0 0.75cm},clip,width=0.5\linewidth]{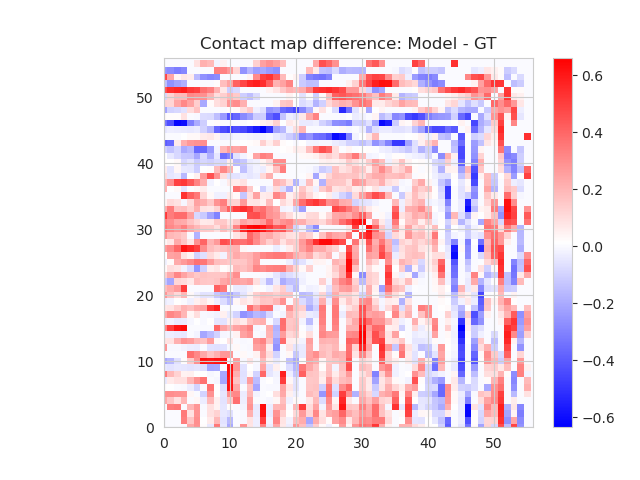}
    \caption{Contact map for Protein G - CGSchNet Fully Trained Model}
\end{figure}

\begin{figure}[htbp]
    \centering
    \includegraphics[trim={0 0 0 0cm},clip,width=\linewidth]{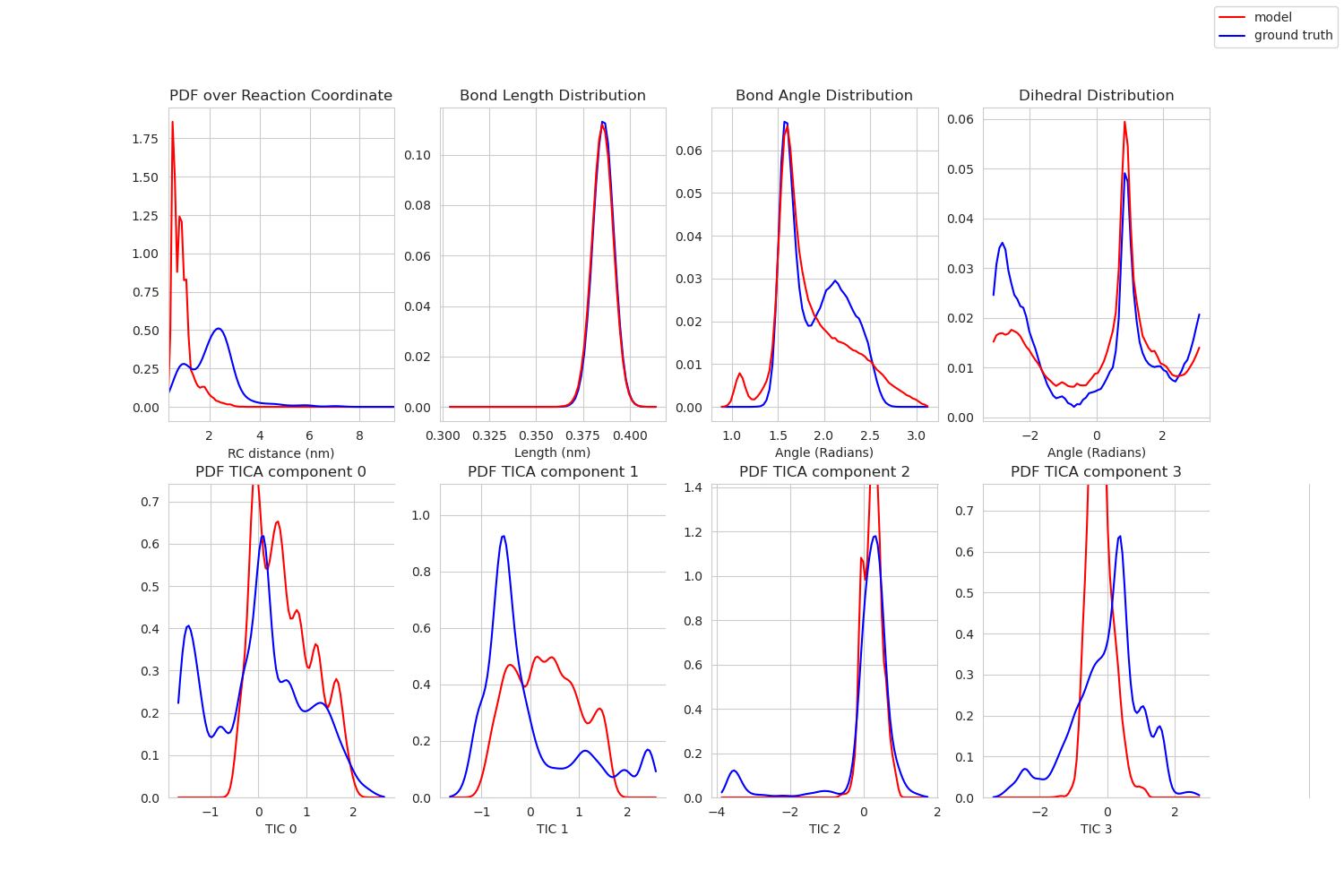}
    \caption{PDFs for Protein G - CGSchNet Fully Trained Model}
    \label{fgr:pdfs_proteing}
\end{figure}

\begin{figure}[htbp]
    \centering
    \includegraphics[trim={0 0 0 0},clip,width=\linewidth]{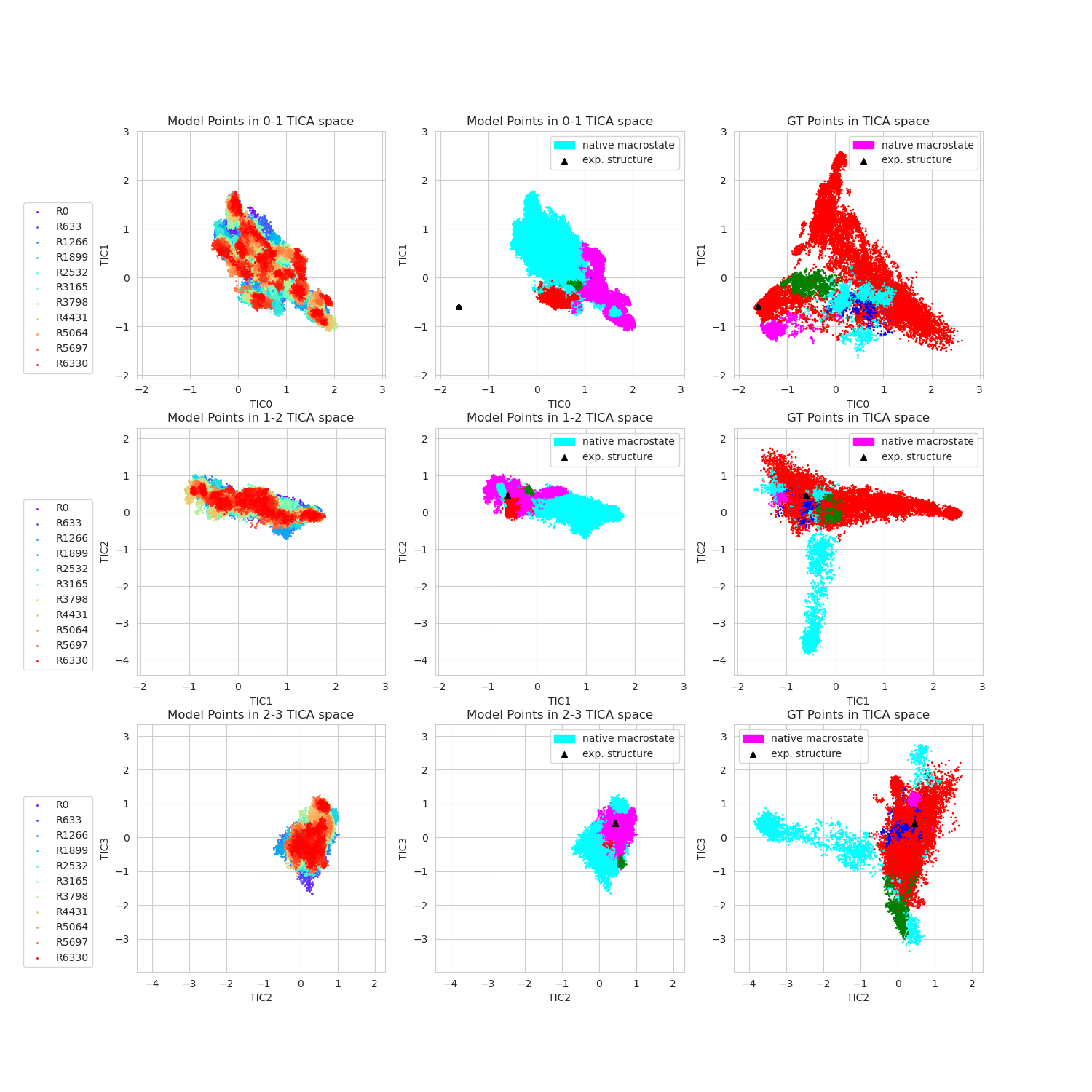}
    \caption{TICA space projections for Protein G - CGSchNet Fully Trained Model}
    \label{fgr:tica_spaces_proteing}
\end{figure}
\clearpage

\begin{table}[h!]
\centering
\resizebox{\textwidth}{!}{%
\begin{tabular}{|l|c|c|c|c|c|c|c|c|}
\hline
\textbf{Metric} & \textbf{TIC 0} & \textbf{TIC 1} & \textbf{TIC 2} & \textbf{TIC 3} & \textbf{Bonds} & \textbf{Angles} & \textbf{Dihedrals} & \textbf{Gyration} \\
\hline
KL & 3.1382 & 1.3059 & 1.0656 & 2.2134 & 0.0076 & 0.1656 & 0.0848 & 2.1418 \\
W1 & 0.5210 & 0.4907 & 0.3105 & 0.5869 & 0.0007 & 0.0951 & 0.4118 & 0.1361 \\
\hline
\end{tabular}
}
\caption{KL and W1 metrics for Protein G with the CGSchNet Fully Trained Model}
\end{table}

\clearpage

\subsubsection{Trp-cage}

\begin{figure}[htbp]
    \centering
    \includegraphics[trim={0 0 0 0.75cm},clip,width=0.5\linewidth]{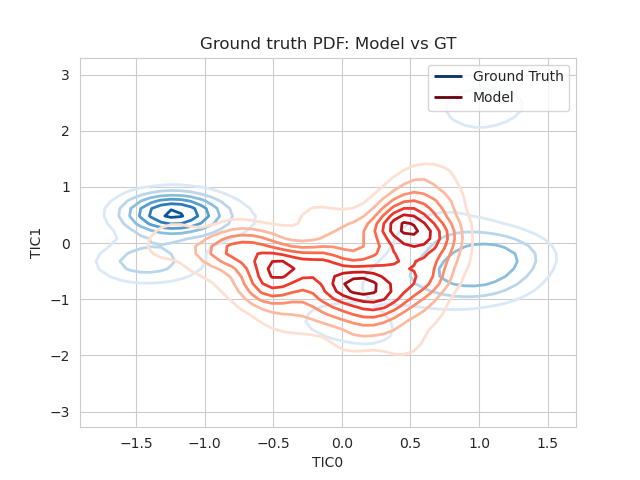}
    \caption{TICA contours for Trp-cage - CGSchNet Fully Trained Model}
    \label{fgr:tica_contours_trpcage}
\end{figure}

\begin{figure}[htbp]
    \centering
    \includegraphics[trim={0 0 0 0.75cm},clip,width=0.5\linewidth]{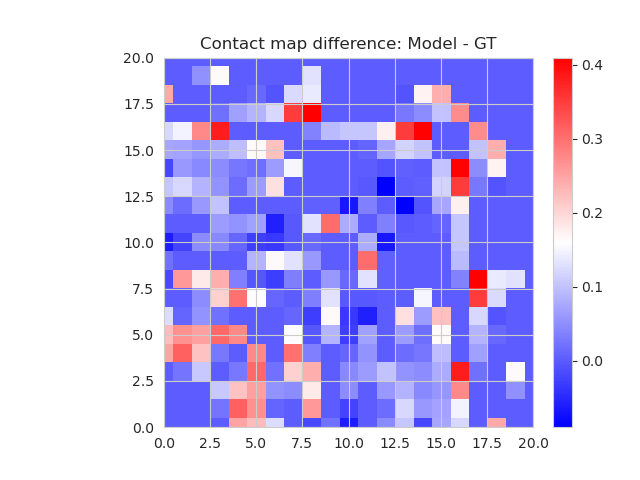}
    \caption{Contact map for Trp-cage - CGSchNet Fully Trained Model}
\end{figure}

\begin{figure}[htbp]
    \centering
    \includegraphics[trim={0 0 0 0cm},clip,width=\linewidth]{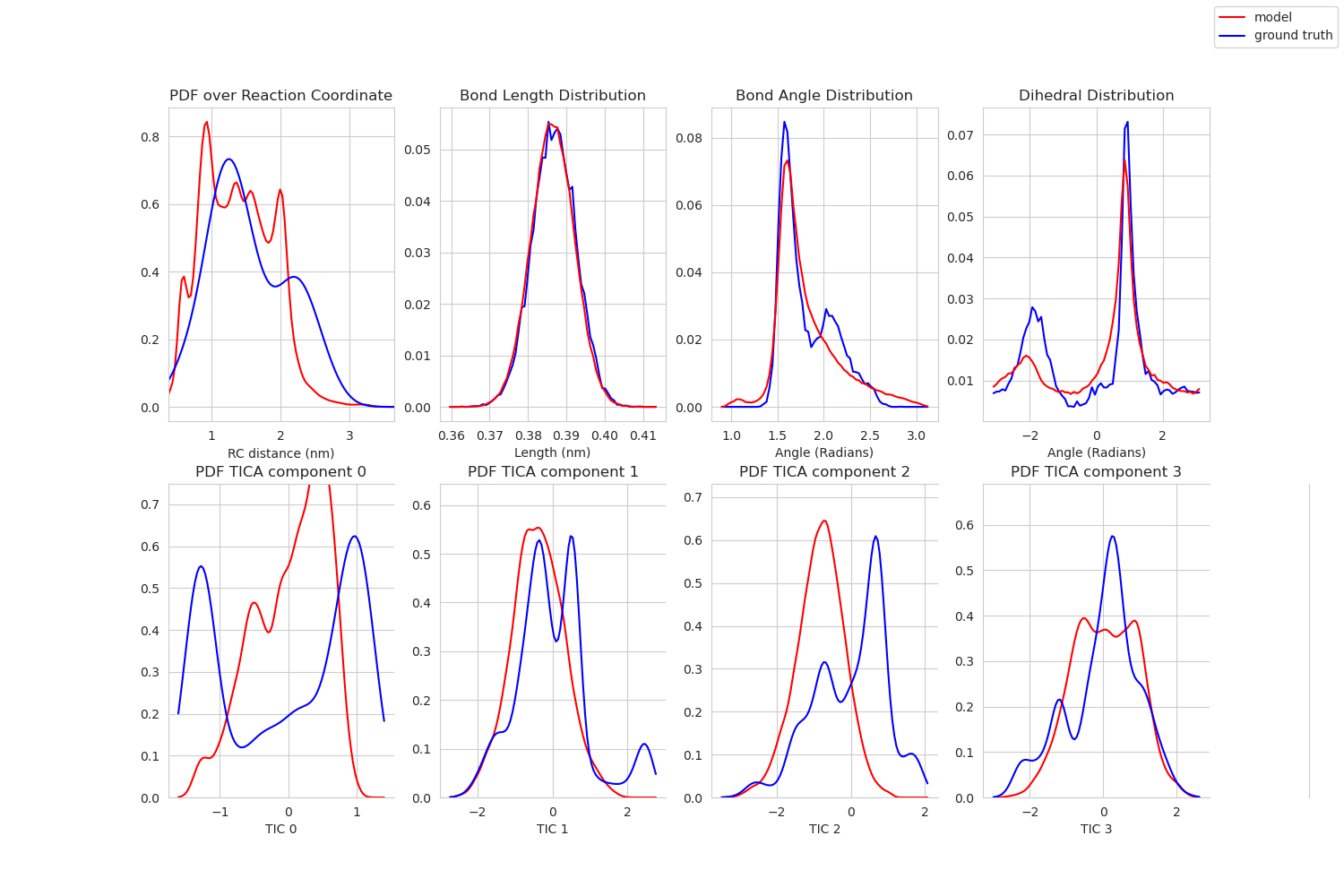}
    \caption{PDFs for Trp-cage - CGSchNet Fully Trained Model}
    \label{fgr:pdfs_trpcage}
\end{figure}

\begin{figure}[htbp]
    \centering
    \includegraphics[trim={0 0 0 0},clip,width=\linewidth]{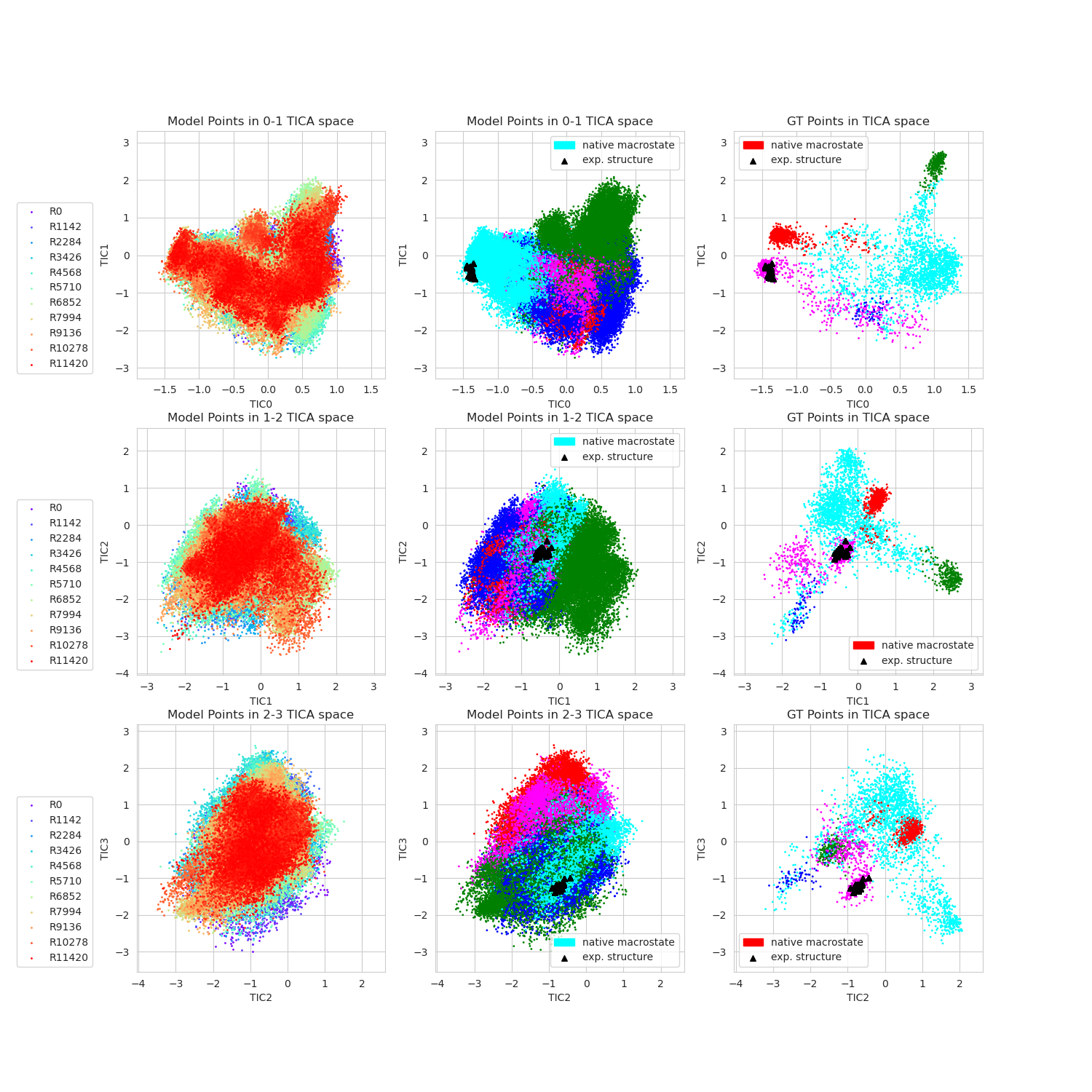}
    \caption{TICA space projections for Trp-cage - CGSchNet Fully Trained Model}
    \label{fgr:tica_spaces_trpcage}
\end{figure}
\clearpage

\begin{table}[h!]
\centering
\resizebox{\textwidth}{!}{%
\begin{tabular}{|l|c|c|c|c|c|c|c|c|}
\hline
\textbf{Metric} & \textbf{TIC 0} & \textbf{TIC 1} & \textbf{TIC 2} & \textbf{TIC 3} & \textbf{Bonds} & \textbf{Angles} & \textbf{Dihedrals} & \textbf{Gyration} \\
\hline
KL & 1.6817 & 0.8146 & 1.5487 & 0.1519 & 0.0038 & 0.1234 & 0.0554 & 0.9295 \\
W1 & 0.4348 & 0.3665 & 0.8268 & 0.1396 & 0.0004 & 0.0460 & 0.1854 & 0.1023 \\
\hline
\end{tabular}
}
\caption{KL and W1 metrics for Trp-cage with the CGSchNet Fully Trained Model}
\end{table}

\clearpage

\subsubsection{WW Domain}

\begin{figure}[htbp]
    \centering
    \includegraphics[trim={0 0 0 0.75cm},clip,width=0.5\linewidth]{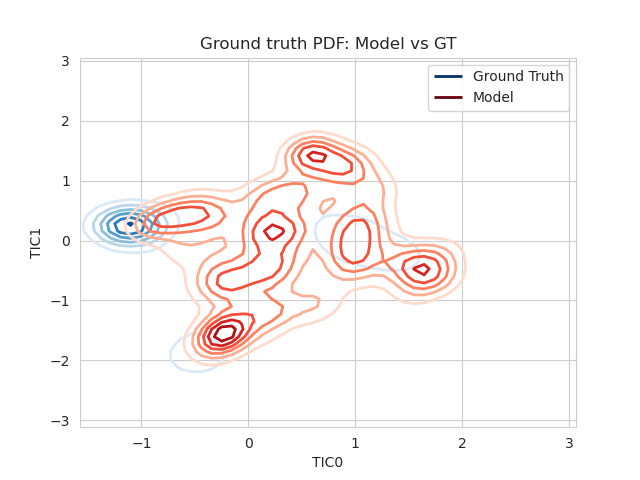}
    \caption{TICA contours for WW Domain - CGSchNet Fully Trained Model}
    \label{fgr:tica_contours_wwdomain}
\end{figure}

\begin{figure}[htbp]
    \centering
    \includegraphics[trim={0 0 0 0.75cm},clip,width=0.5\linewidth]{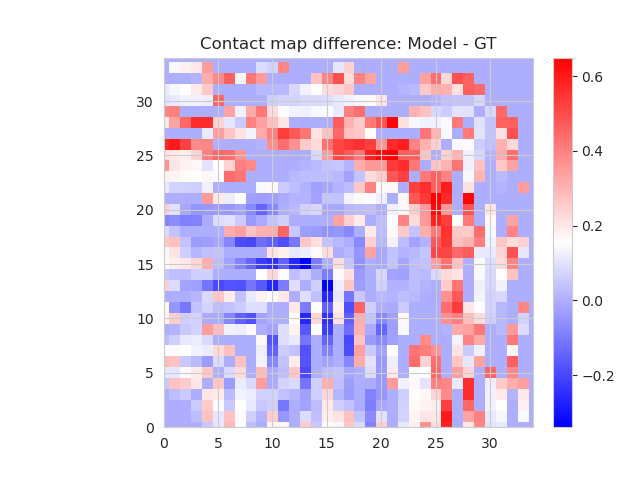}
    \caption{Contact map for WW Domain - CGSchNet Fully Trained Model}
\end{figure}

\begin{figure}[htbp]
    \centering
    \includegraphics[trim={0 0 0 0cm},clip,width=\linewidth]{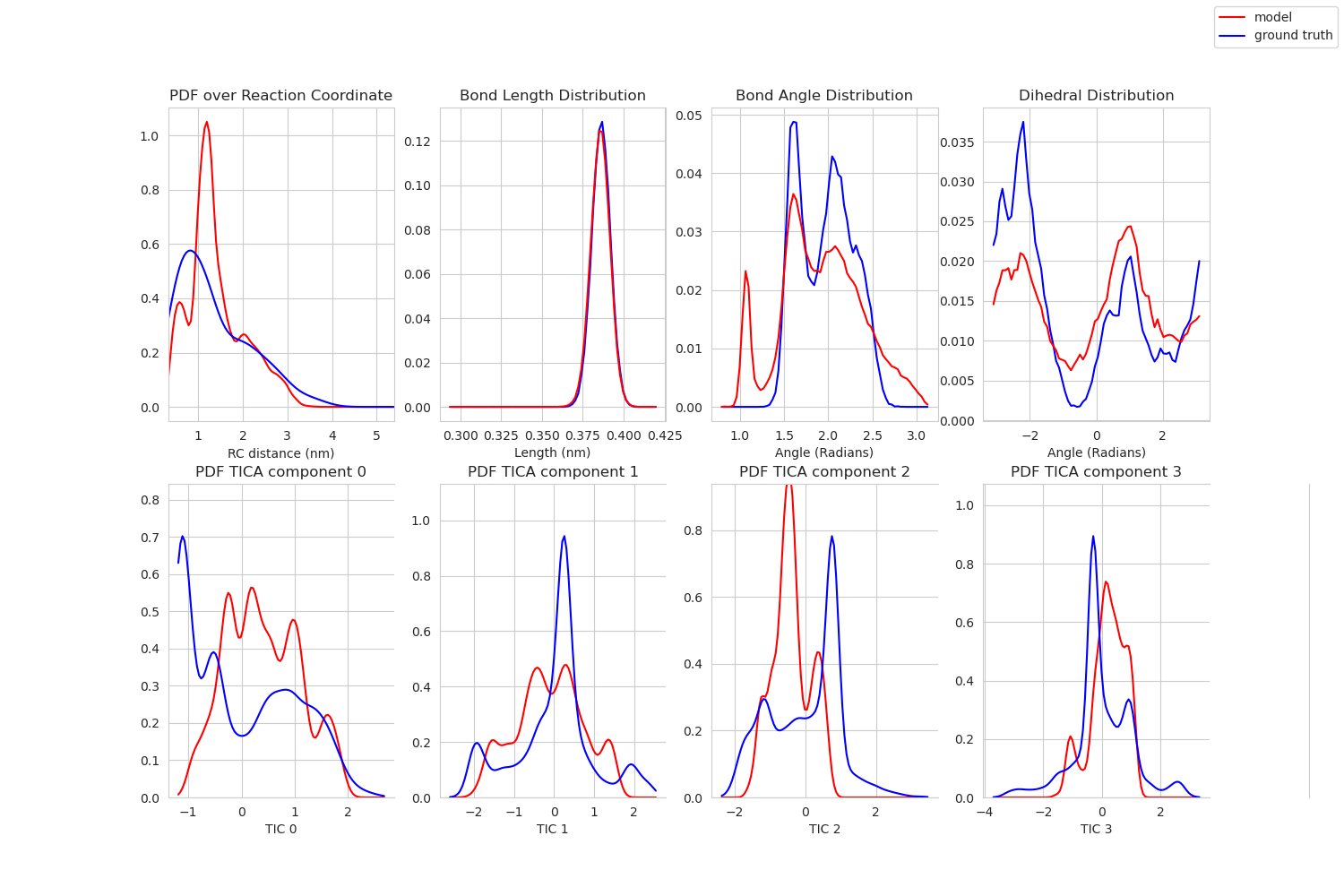}
    \caption{PDFs for WW Domain - CGSchNet Fully Trained Model}
    \label{fgr:pdfs_wwdomain}
\end{figure}

\begin{figure}[htbp]
    \centering
    \includegraphics[trim={0 0 0 0},clip,width=\linewidth]{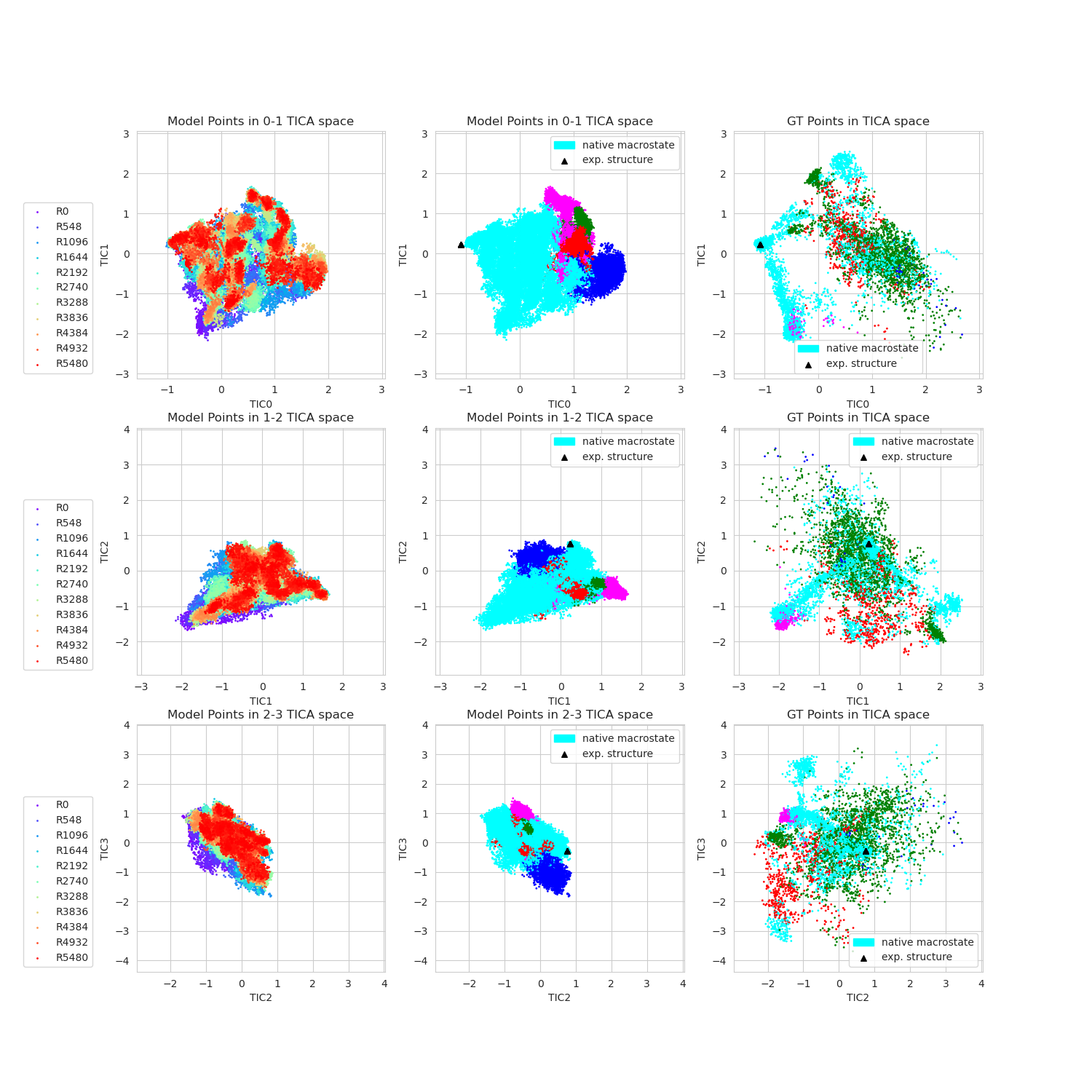}
    \caption{TICA space projections for WW Domain - CGSchNet Fully Trained Model}
    \label{fgr:tica_spaces_wwdomain}
\end{figure}
\clearpage

\begin{table}[h!]
\centering
\resizebox{\textwidth}{!}{%
\begin{tabular}{|l|c|c|c|c|c|c|c|c|}
\hline
\textbf{Metric} & \textbf{TIC 0} & \textbf{TIC 1} & \textbf{TIC 2} & \textbf{TIC 3} & \textbf{Bonds} & \textbf{Angles} & \textbf{Dihedrals} & \textbf{Gyration} \\
\hline
KL & 0.9144 & 0.6805 & 1.7252 & 1.1981 & 0.0059 & 0.2254 & 0.0841 & 1.9351 \\
W1 & 0.3312 & 0.2071 & 0.5608 & 0.3849 & 0.0005 & 0.0356 & 0.4661 & 0.1449 \\
\hline
\end{tabular}
}
\caption{KL and W1 metrics for WWdomain with the CGSchNet Fully Trained Model}
\end{table}

\clearpage

\subsection{Full Benchmark Diagrams - CG ML Under Trained Model}

\begin{figure}
    \includegraphics[trim={0 15in 0 2in},clip,scale=0.275]{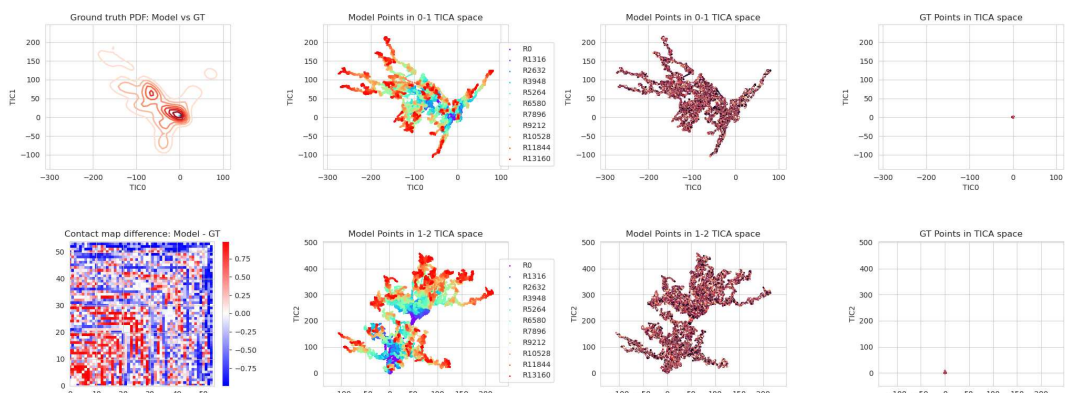}
    \caption{Homeodomain Protein with a poorly trained CGSchNet model which has exploded as an example. Note that this is not the under-trained model used in the analysis, only as an example of what an explosion may look like. Note how the ground truth appears almost non-existent when compared to the model, and the contact map contains lots of deep red and blue suggesting large differences in values. }
  \label{fgr:homeodomain_exploded}
\end{figure}
\clearpage

\subsubsection{A3D}

\begin{figure}[htbp]
    \centering
    \includegraphics[trim={0 0 0 0.75cm},clip,width=0.5\linewidth]{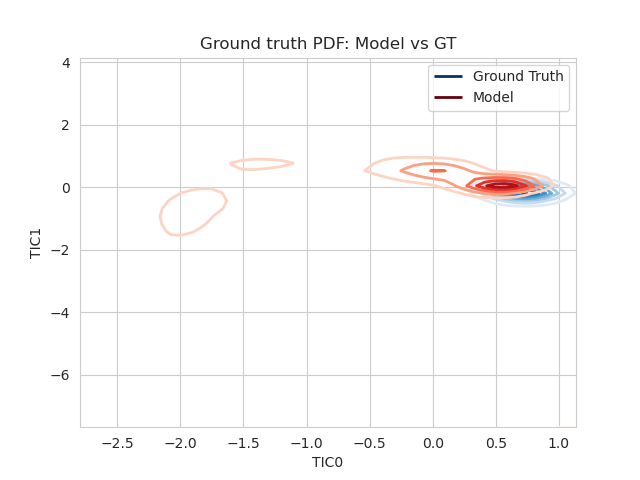}
    \caption{TICA contours for A3D - CGSchNet Under-Trained Model}
    \label{fgr:tica_contours_a3d}
\end{figure}

\begin{figure}[htbp]
    \centering
    \includegraphics[trim={0 0 0 0.75cm},clip,width=0.5\linewidth]{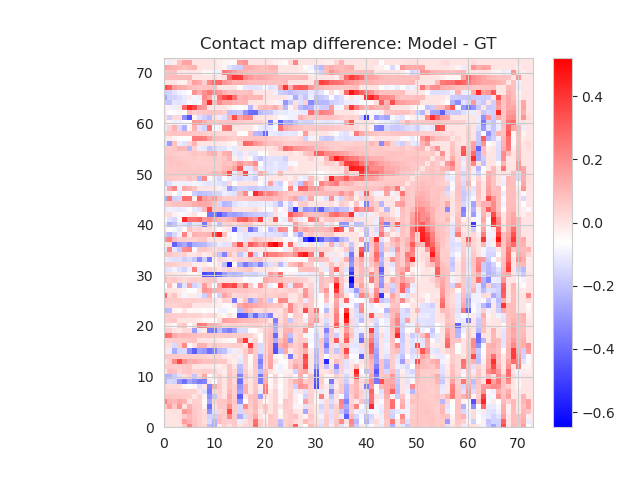}
    \caption{Contact map for A3D - CGSchNet Under-Trained Model}
\end{figure}

\begin{figure}[htbp]
    \centering
    \includegraphics[trim={0 0 0 0cm},clip,width=\linewidth]{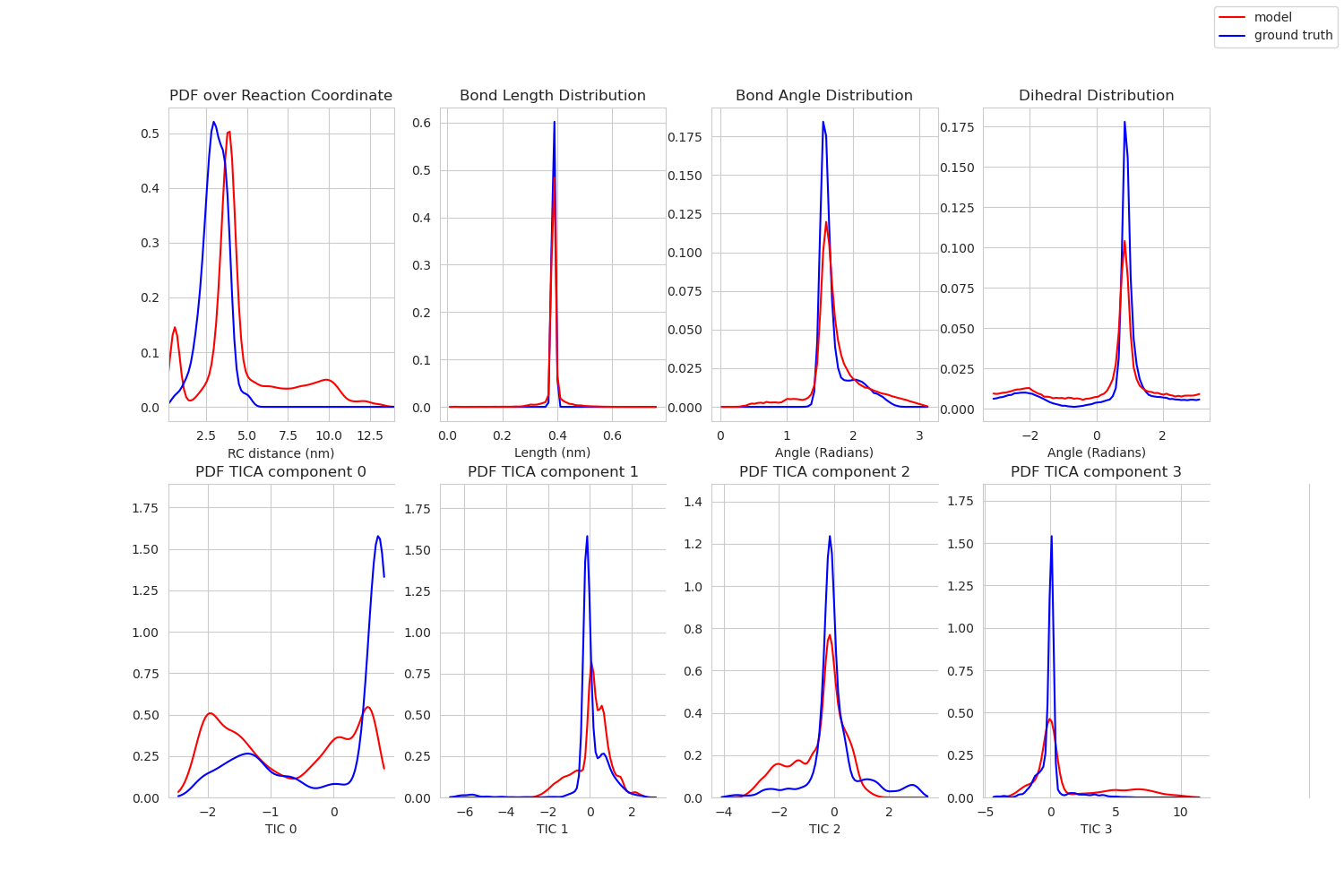}
    \caption{PDFs for A3D - CGSchNet Under-Trained Model}
    \label{fgr:pdfs_a3d}
\end{figure}

\begin{figure}[htbp]
    \centering
    \includegraphics[trim={0 0 0 0},clip,width=\linewidth]{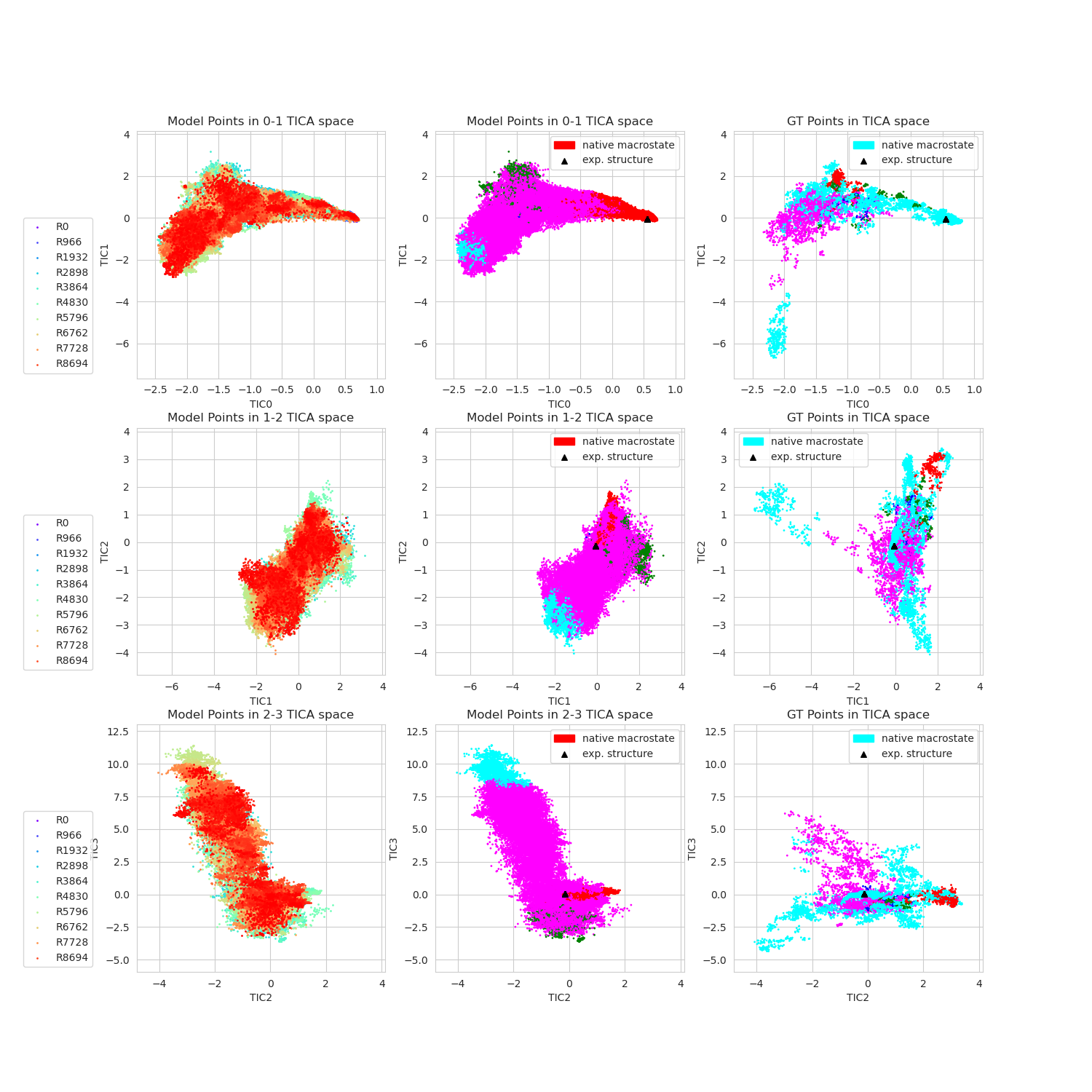}
    \caption{TICA space projections for A3D - CGSchNet Under-Trained Model. Note how the TICA space explored is much larger when compared to the Implicit All Atom and the Fully Trained model. This is an unexpected anomaly and led us to investigate what in the model could potentially lead to this behavior. While we have not found exactly why, the benchmark result has helped guide our investigation. Despite this better explored TICA space, the structural stability of the bond lengths, bond angles, and diherdal angles is decreased when compared the fully trained model. This is because while the model was able to explore more area, it was not as stable and many segments imploded. }
    \label{fgr:tica_spaces_a3d}
\end{figure}
\clearpage

\begin{table}[h!]
\centering
\resizebox{\textwidth}{!}{%
\begin{tabular}{|l|c|c|c|c|c|c|c|c|}
\hline
\textbf{Metric} & \textbf{TIC 0} & \textbf{TIC 1} & \textbf{TIC 2} & \textbf{TIC 3} & \textbf{Bonds} & \textbf{Angles} & \textbf{Dihedrals} & \textbf{Gyration} \\
\hline
KL & 0.7080 & 0.8750 & 0.7492 & 0.6468 & 0.1545 & 0.2076 & 0.2134 & 0.8945 \\
W1 & 0.6139 & 0.3653 & 0.5287 & 1.5593 & 0.0031 & 0.0885 & 0.4242 & 0.5089 \\
\hline
\end{tabular}
}
\caption{KL and W1 metrics for a3d}
\end{table}

\clearpage

\subsubsection{BBA}

\begin{figure}[htbp]
    \centering
    \includegraphics[trim={0 0 0 0.75cm},clip,width=0.5\linewidth]{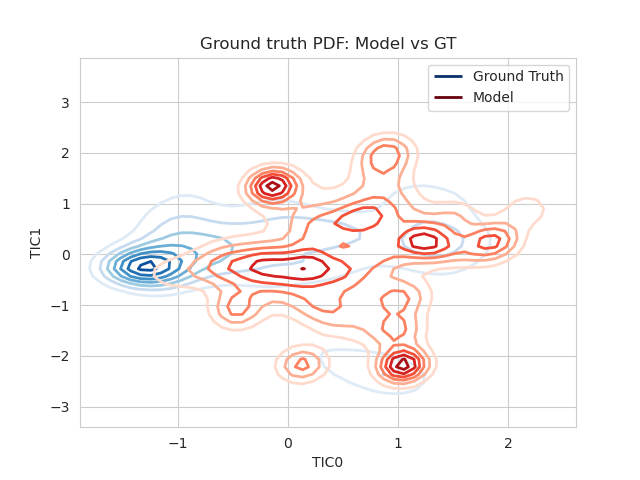}
    \caption{TICA contours for BBA - CGSchNet Under-Trained Model}
    \label{fgr:tica_contours_bba}
\end{figure}

\begin{figure}[htbp]
    \centering
    \includegraphics[trim={0 0 0 0.75cm},clip,width=0.5\linewidth]{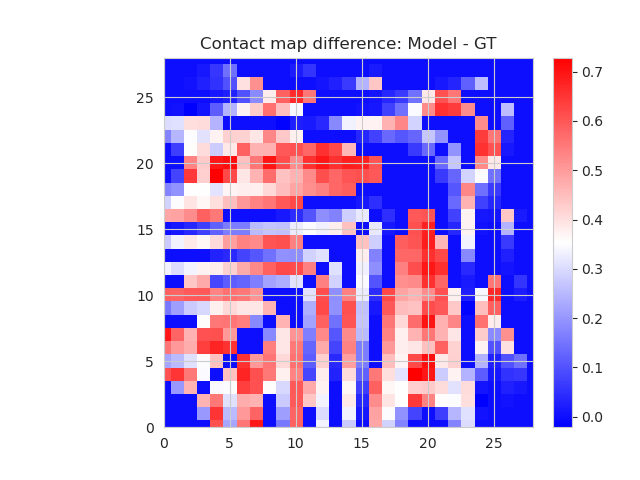}
    \caption{Contact map for BBA - CGSchNet Under-Trained Model}
\end{figure}

\begin{figure}[htbp]
    \centering
    \includegraphics[trim={0 0 0 0cm},clip,width=\linewidth]{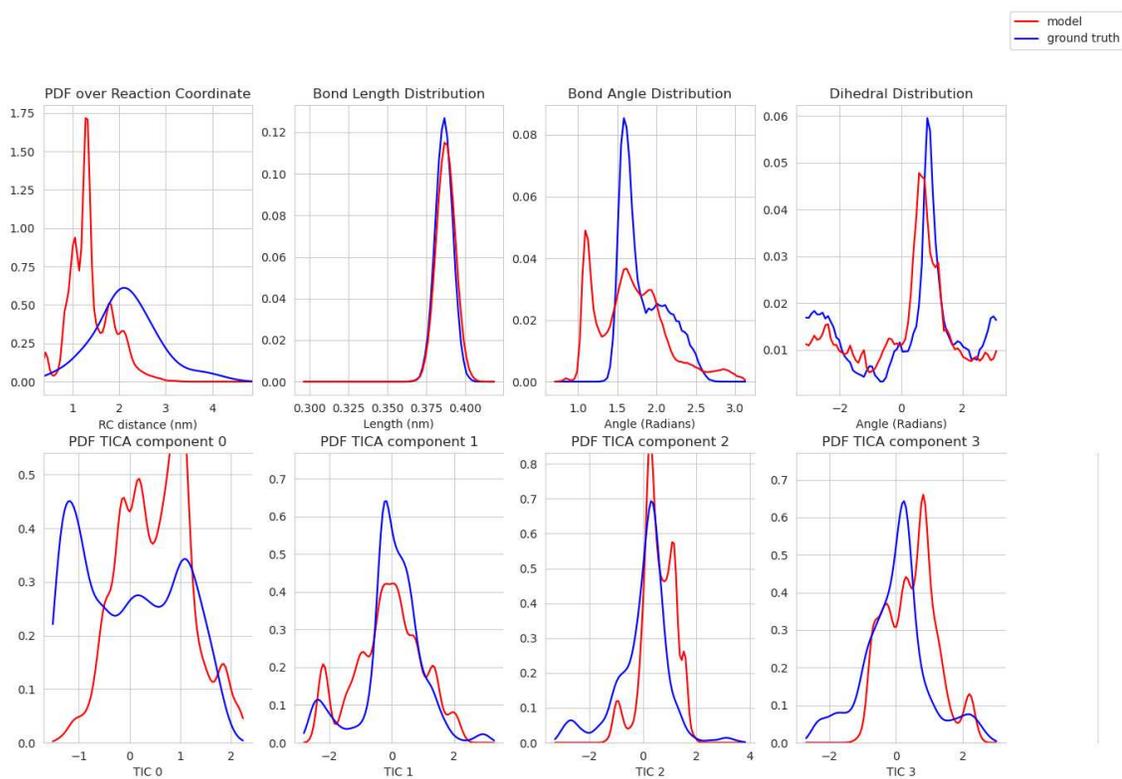}
    \caption{PDFs for BBA - CGSchNet Under-Trained Model}
    \label{fgr:pdfs_bba}
\end{figure}

\begin{figure}[htbp]
    \centering
    \includegraphics[trim={0 0 0 0},clip,width=\linewidth]{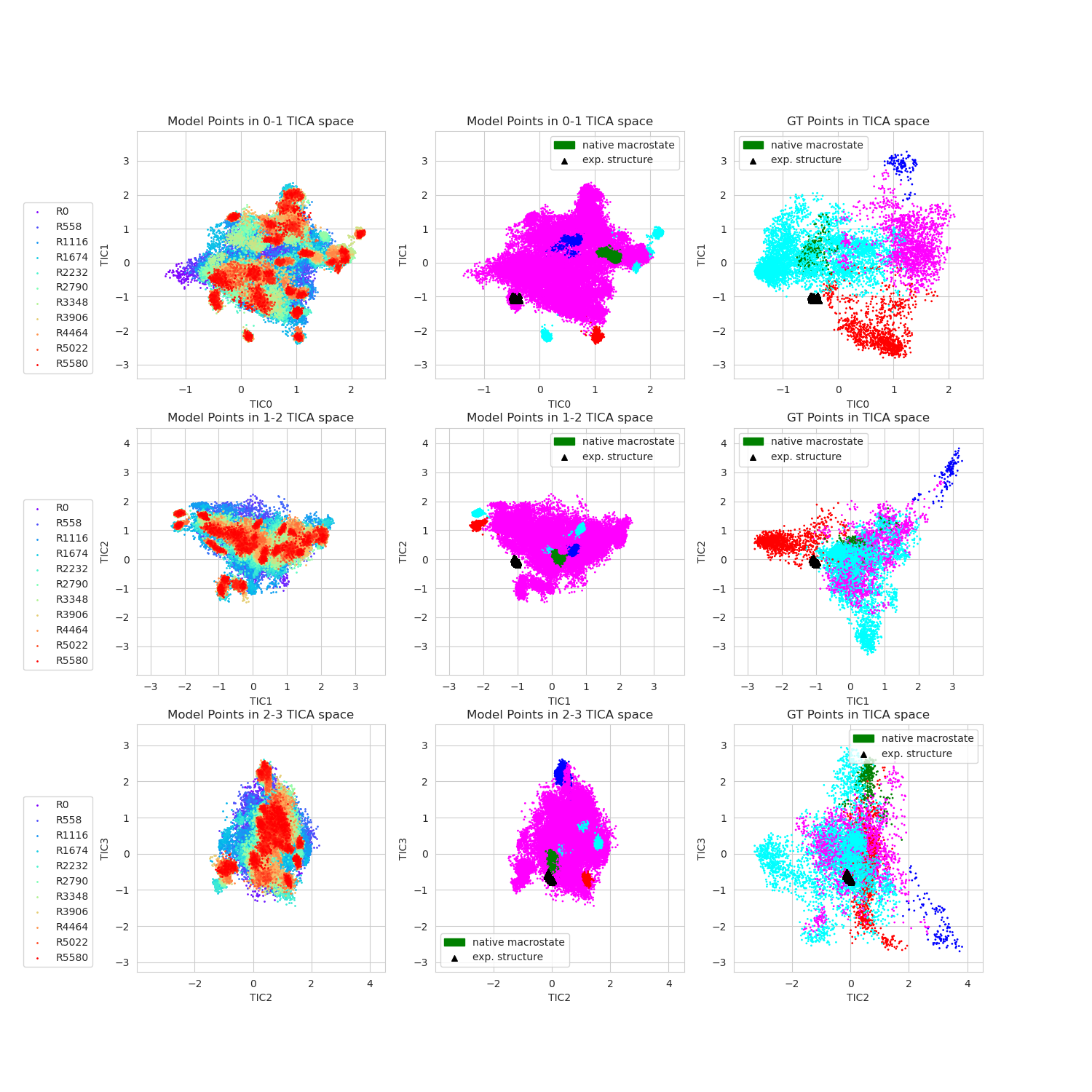}
    \caption{TICA space projections for BBA - CGSchNet Under-Trained Model}
    \label{fgr:tica_spaces_bba}
\end{figure}
\clearpage

\begin{table}[h!]
\centering
\resizebox{\textwidth}{!}{%
\begin{tabular}{|l|c|c|c|c|c|c|c|c|}
\hline
\textbf{Metric} & \textbf{TIC 0} & \textbf{TIC 1} & \textbf{TIC 2} & \textbf{TIC 3} & \textbf{Bonds} & \textbf{Angles} & \textbf{Dihedrals} & \textbf{Gyration} \\
\hline
KL & 0.6981 & 0.2568 & 1.0406 & 0.9517 & 0.0154 & 0.4750 & 0.2025 & 5.7981 \\
W1 & 0.4902 & 0.2095 & 0.5975 & 0.4876 & 0.0007 & 0.0605 & 0.4272 & 0.3480 \\
\hline
\end{tabular}
}
\caption{KL and W1 metrics for BBA with the CGSchNet Under-Trained Model}
\end{table}

\clearpage

\subsubsection{Chignolin}

\begin{figure}[htbp]
    \centering
    \includegraphics[trim={0 0 0 0.75cm},clip,width=0.5\linewidth]{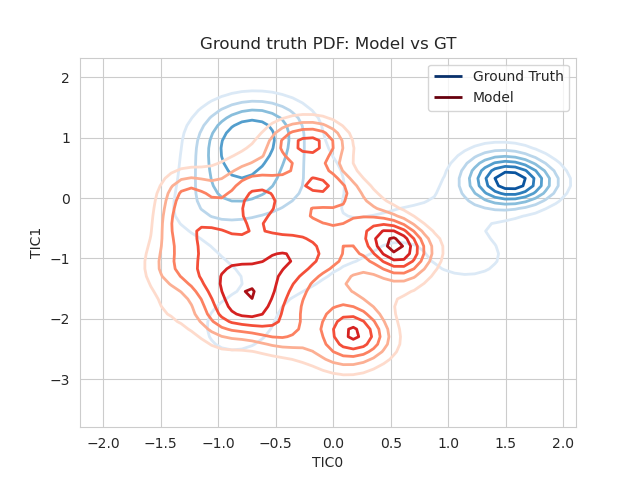}
    \caption{TICA contours for Chignolin - CGSchNet Under-Trained Model}
    \label{fgr:tica_contours_chignolin}
\end{figure}

\begin{figure}[htbp]
    \centering
    \includegraphics[trim={0 0 0 0.75cm},clip,width=0.5\linewidth]{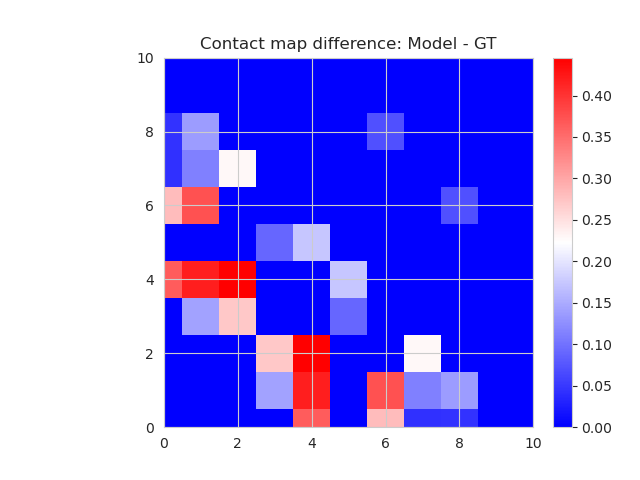}
    \caption{Contact map for Chignolin - CGSchNet Under-Trained Model}
\end{figure}

\begin{figure}[htbp]
    \centering
    \includegraphics[trim={0 0 0 0cm},clip,width=\linewidth]{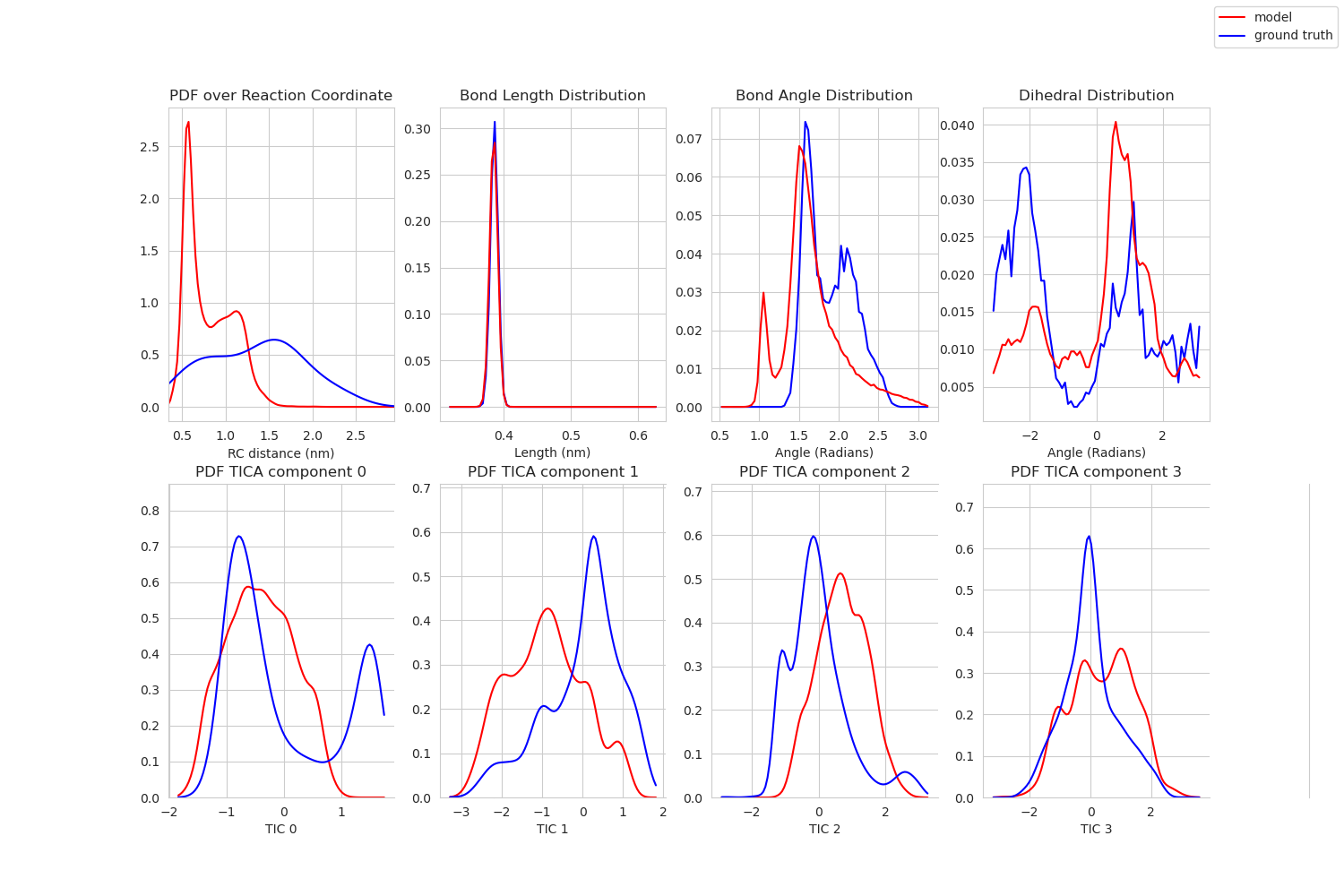}
    \caption{PDFs for Chignolin - CGSchNet Under-Trained Model}
    \label{fgr:pdfs_chignolin}
\end{figure}

\begin{figure}[htbp]
    \centering
    \includegraphics[trim={0 0 0 0},clip,width=\linewidth]{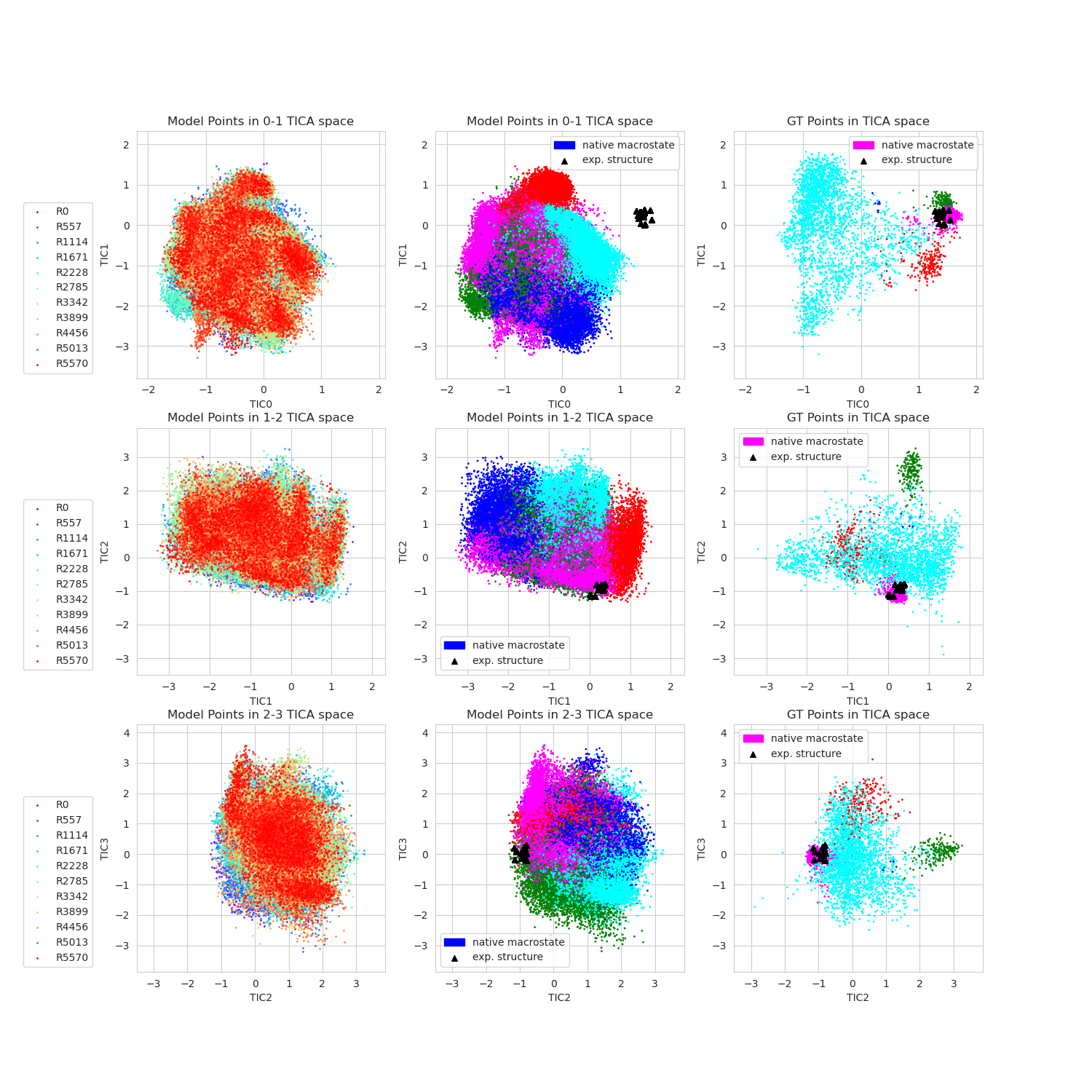}
    \caption{TICA space projections for Chignolin - CGSchNet Under-Trained Model}
    \label{fgr:tica_spaces_chignolin}
\end{figure}
\clearpage

\begin{table}[h!]
\centering
\resizebox{\textwidth}{!}{%
\begin{tabular}{|l|c|c|c|c|c|c|c|c|}
\hline
\textbf{Metric} & \textbf{TIC 0} & \textbf{TIC 1} & \textbf{TIC 2} & \textbf{TIC 3} & \textbf{Bonds} & \textbf{Angles} & \textbf{Dihedrals} & \textbf{Gyration} \\
\hline
KL & 2.8054 & 0.5400 & 0.7951 & 0.1570 & 0.0842 & 0.5551 & 0.1776 & 4.6291 \\
W1 & 0.3609 & 0.8948 & 0.7436 & 0.3715 & 0.0023 & 0.1803 & 0.6544 & 0.1778 \\
\hline
\end{tabular}
}
\caption{KL and W1 metrics for Chignolin with the CGSchNet Under-Trained Model}
\end{table}

\clearpage

\subsubsection{Homeodomain}

\begin{figure}[htbp]
    \centering
    \includegraphics[trim={0 0 0 0.75cm},clip,width=0.5\linewidth]{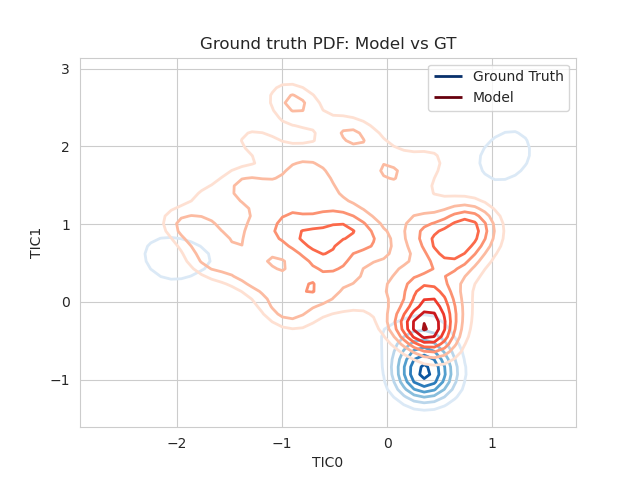}
    \caption{TICA contours for Homeodomain - CGSchNet Under-Trained Model}
    \label{fgr:tica_contours_homeodomain}
\end{figure}

\begin{figure}[htbp]
    \centering
    \includegraphics[trim={0 0 0 0.75cm},clip,width=0.5\linewidth]{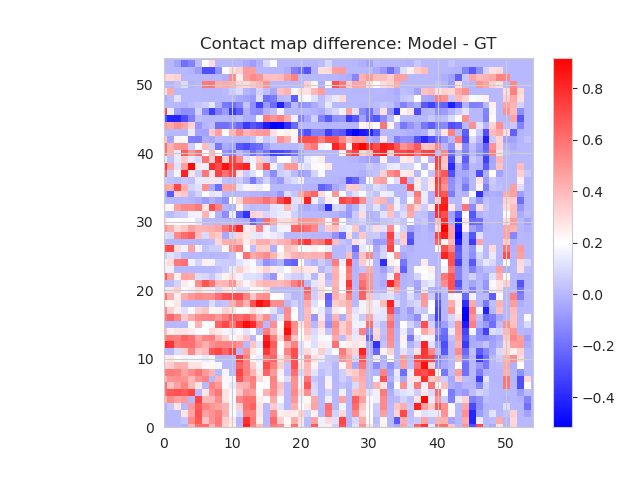}
    \caption{Contact map for Homeodomain - CGSchNet Under-Trained Model}
\end{figure}

\begin{figure}[htbp]
    \centering
    \includegraphics[trim={0 0 0 0cm},clip,width=\linewidth]{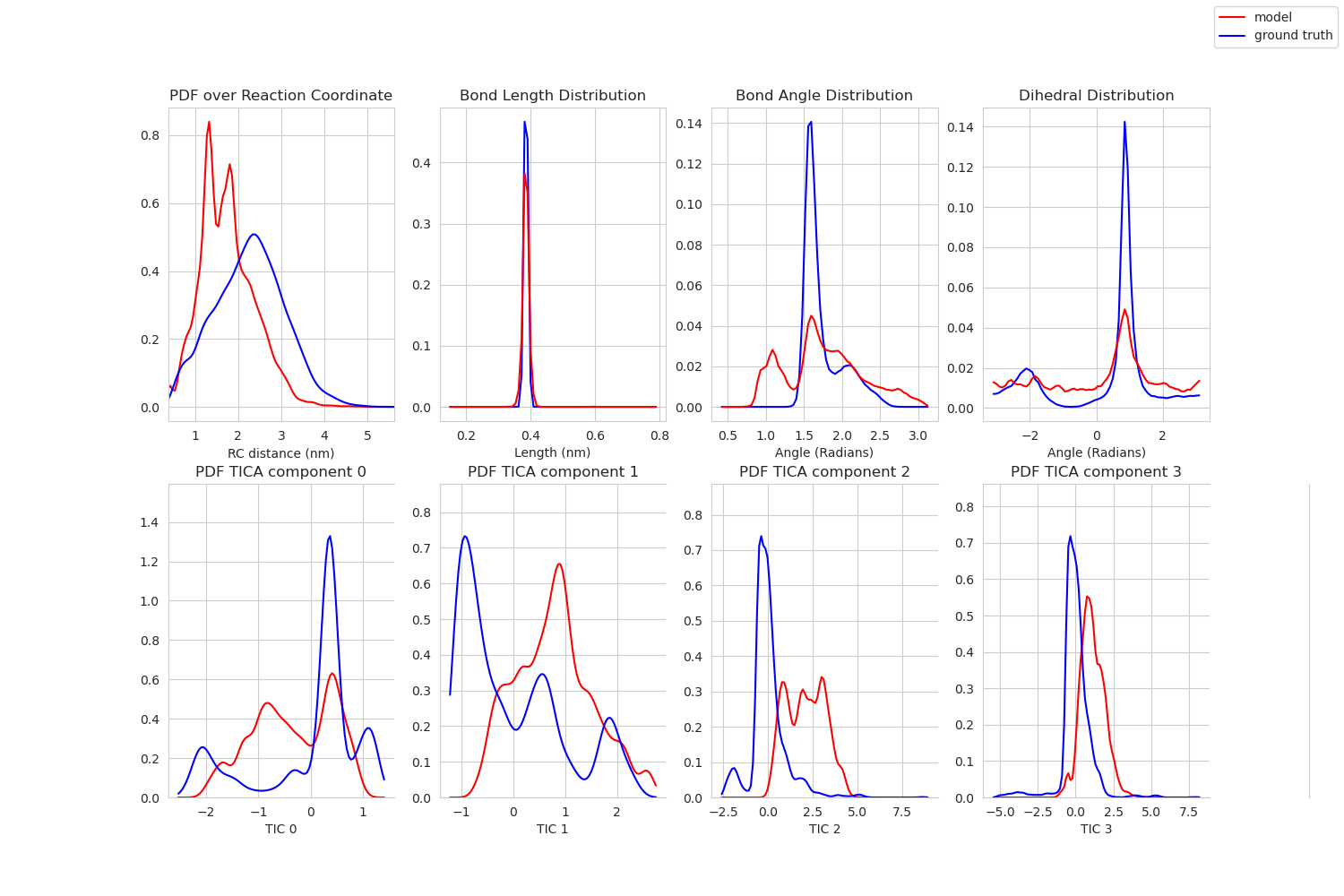}
    \caption{PDFs for Homeodomain - CGSchNet Under-Trained Model}
    \label{fgr:pdfs_homeodomain}
\end{figure}

\begin{figure}[htbp]
    \centering
    \includegraphics[trim={0 0 0 0},clip,width=\linewidth]{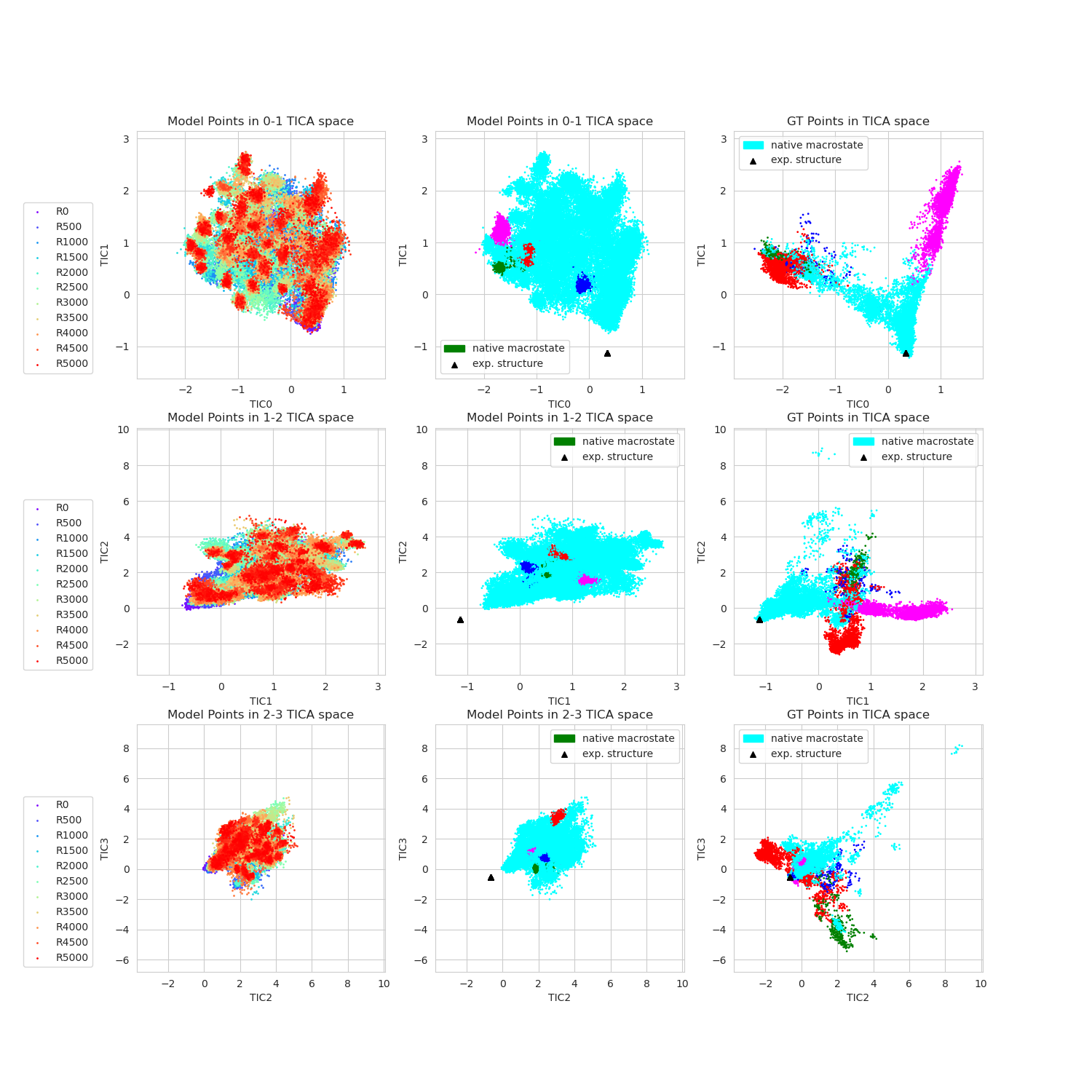}
    \caption{TICA space projections for Homeodomain - CGSchNet Under-Trained Model}
    \label{fgr:tica_spaces_homeodomain}
\end{figure}
\clearpage

\begin{table}[h!]
\centering
\resizebox{\textwidth}{!}{%
\begin{tabular}{|l|c|c|c|c|c|c|c|c|}
\hline
\textbf{Metric} & \textbf{TIC 0} & \textbf{TIC 1} & \textbf{TIC 2} & \textbf{TIC 3} & \textbf{Bonds} & \textbf{Angles} & \textbf{Dihedrals} & \textbf{Gyration} \\
\hline
KL & 0.9845 & 2.1244 & 4.7541 & 1.4255 & 0.1561 & 0.6603 & 0.3144 & 1.4470 \\
W1 & 0.4917 & 0.7547 & 2.1111 & 1.1386 & 0.0026 & 0.1845 & 0.3238 & 0.1846 \\
\hline
\end{tabular}
}
\caption{KL and W1 metrics for Homeodomain with the CGSchNet Under-Trained Model}
\end{table}

\clearpage

\subsubsection{$\lambda$-repressor}

\begin{figure}[htbp]
    \centering
    \includegraphics[trim={0 0 0 0.75cm},clip,width=0.5\linewidth]{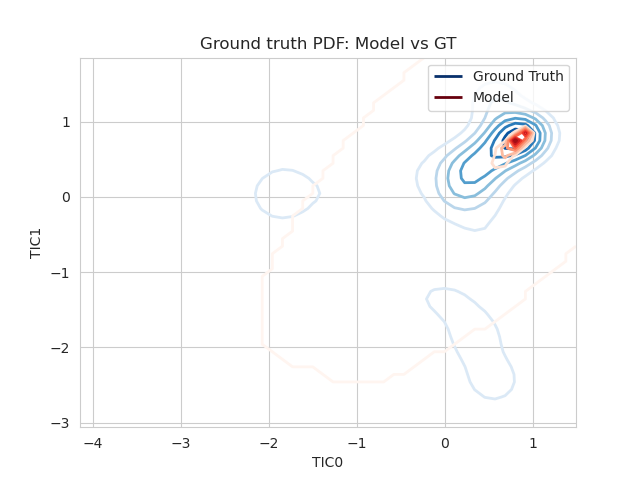}
    \caption{TICA contours for $\lambda$-repressor - CGSchNet Under-Trained Model}
    \label{fgr:tica_contours_lambda}
\end{figure}

\begin{figure}[htbp]
    \centering
    \includegraphics[trim={0 0 0 0.75cm},clip,width=0.5\linewidth]{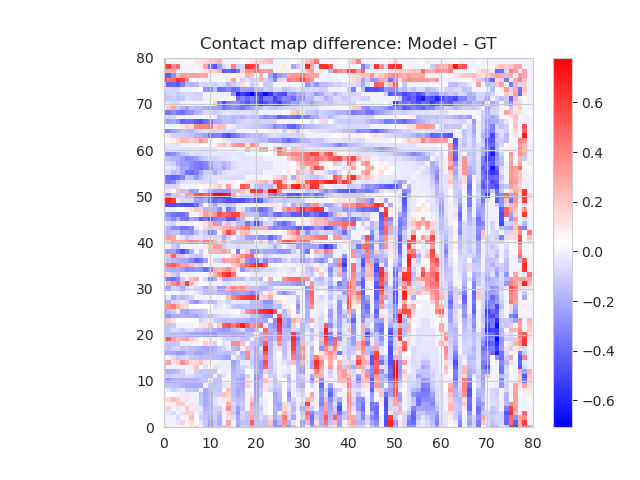}
    \caption{Contact map for $\lambda$-repressor - CGSchNet Under-Trained Model}
\end{figure}

\begin{figure}[htbp]
    \centering
    \includegraphics[trim={0 0 0 0cm},clip,width=\linewidth]{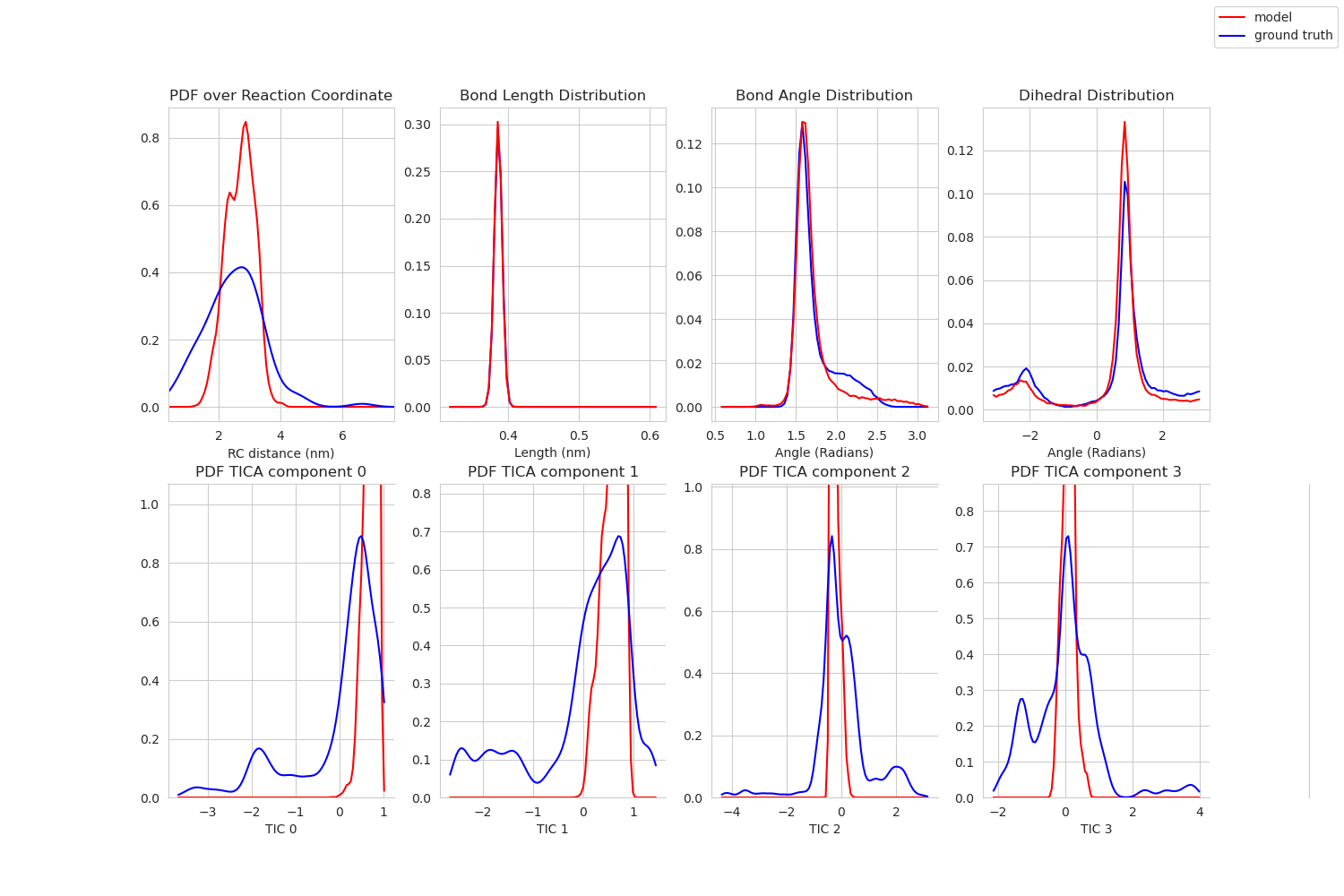}
    \caption{PDFs for $\lambda$-repressor - CGSchNet Under-Trained Model}
    \label{fgr:pdfs_lambda}
\end{figure}

\begin{figure}[htbp]
    \centering
    \includegraphics[trim={0 0 0 0},clip,width=\linewidth]{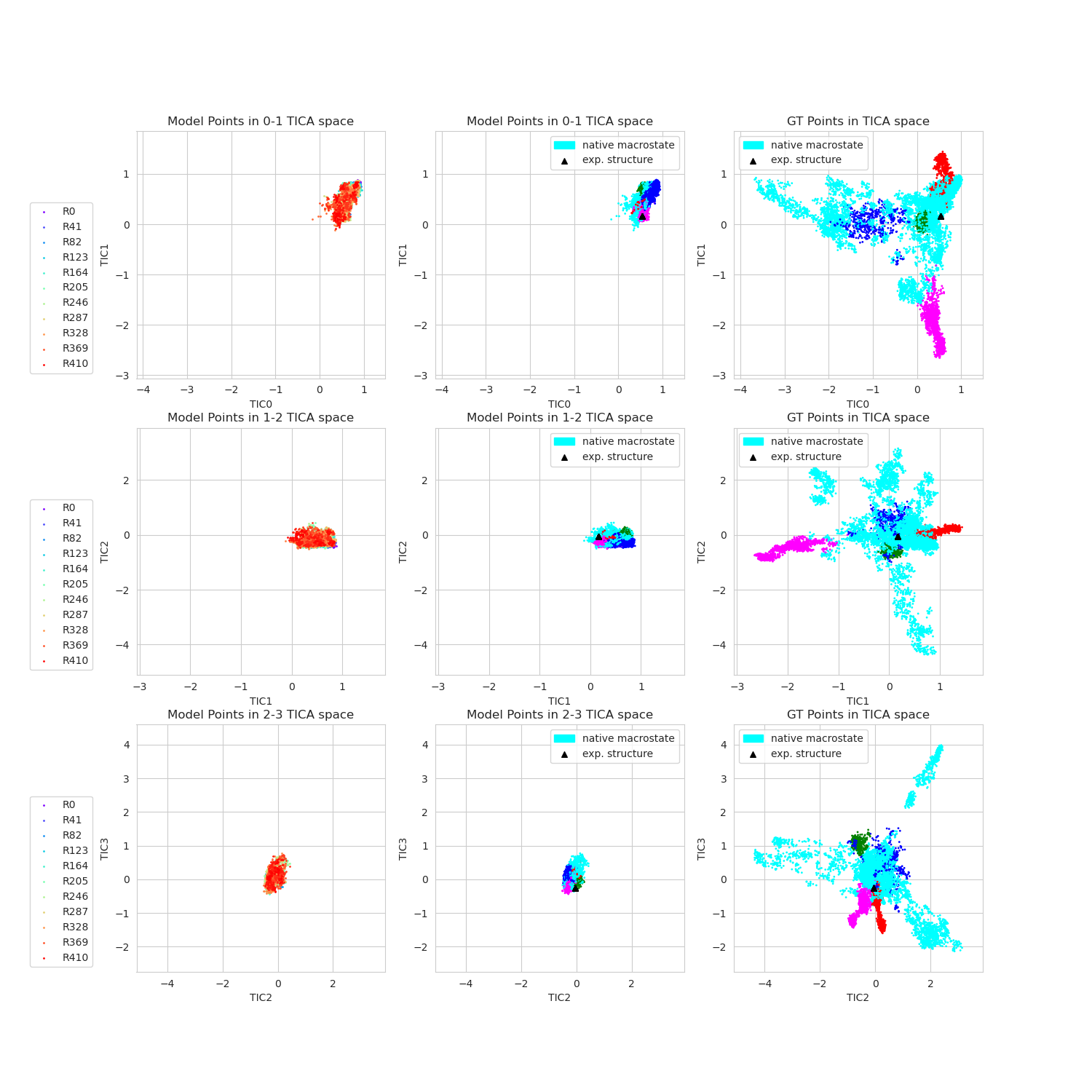}
    \caption{TICA space projections for $\lambda$-repressor - CGSchNet Under-Trained Model}
    \label{fgr:tica_spaces_lambda}
\end{figure}
\clearpage
\begin{table}[h!]
\centering
\resizebox{\textwidth}{!}{%
\begin{tabular}{|l|c|c|c|c|c|c|c|c|}
\hline
\textbf{Metric} & \textbf{TIC 0} & \textbf{TIC 1} & \textbf{TIC 2} & \textbf{TIC 3} & \textbf{Bonds} & \textbf{Angles} & \textbf{Dihedrals} & \textbf{Gyration} \\
\hline
KL & 3.1606 & 3.7179 & 3.9355 & 4.5957 & 0.0025 & 0.1091 & 0.0660 & 0.8406 \\
W1 & 0.7695 & 0.6949 & 0.5835 & 0.6023 & 0.0002 & 0.0641 & 0.2406 & 0.1035 \\
\hline
\end{tabular}
}
\caption{KL and W1 metrics for lambda with the CGSchNet Under-Trained Model}
\end{table}
\clearpage

\subsubsection{Protein B}

\begin{figure}[htbp]
    \centering
    \includegraphics[trim={0 0 0 0.75cm},clip,width=0.5\linewidth]{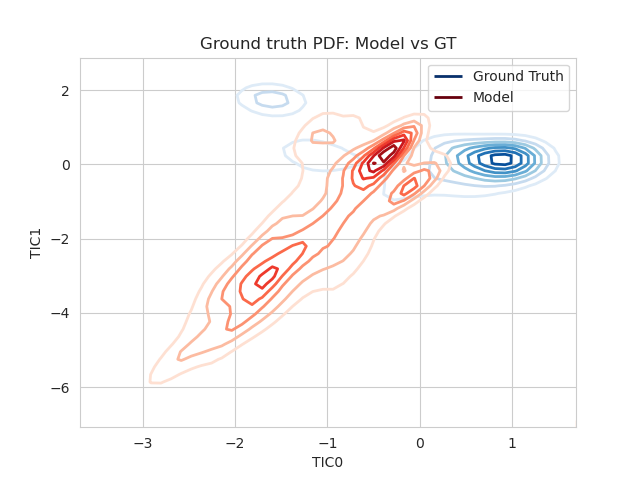}
    \caption{TICA contours for Protein B - CGSchNet Under-Trained Model}
    \label{fgr:tica_contours_proteinb}
\end{figure}

\begin{figure}[htbp]
    \centering
    \includegraphics[trim={0 0 0 0.75cm},clip,width=0.5\linewidth]{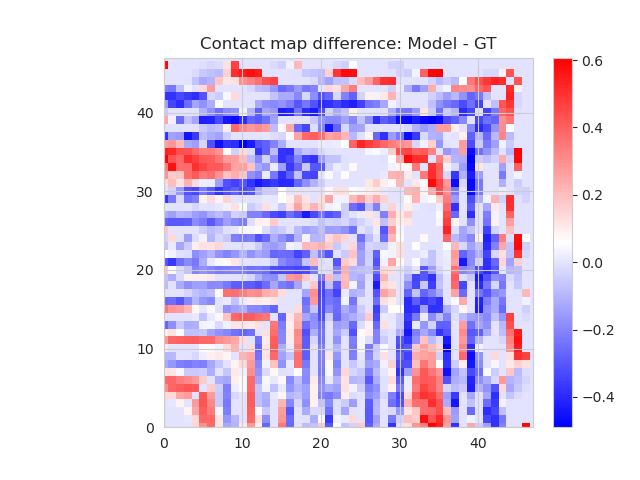}
    \caption{Contact map for Protein B - CGSchNet Under-Trained Model}
\end{figure}

\begin{figure}[htbp]
    \centering
    \includegraphics[trim={0 0 0 0cm},clip,width=\linewidth]{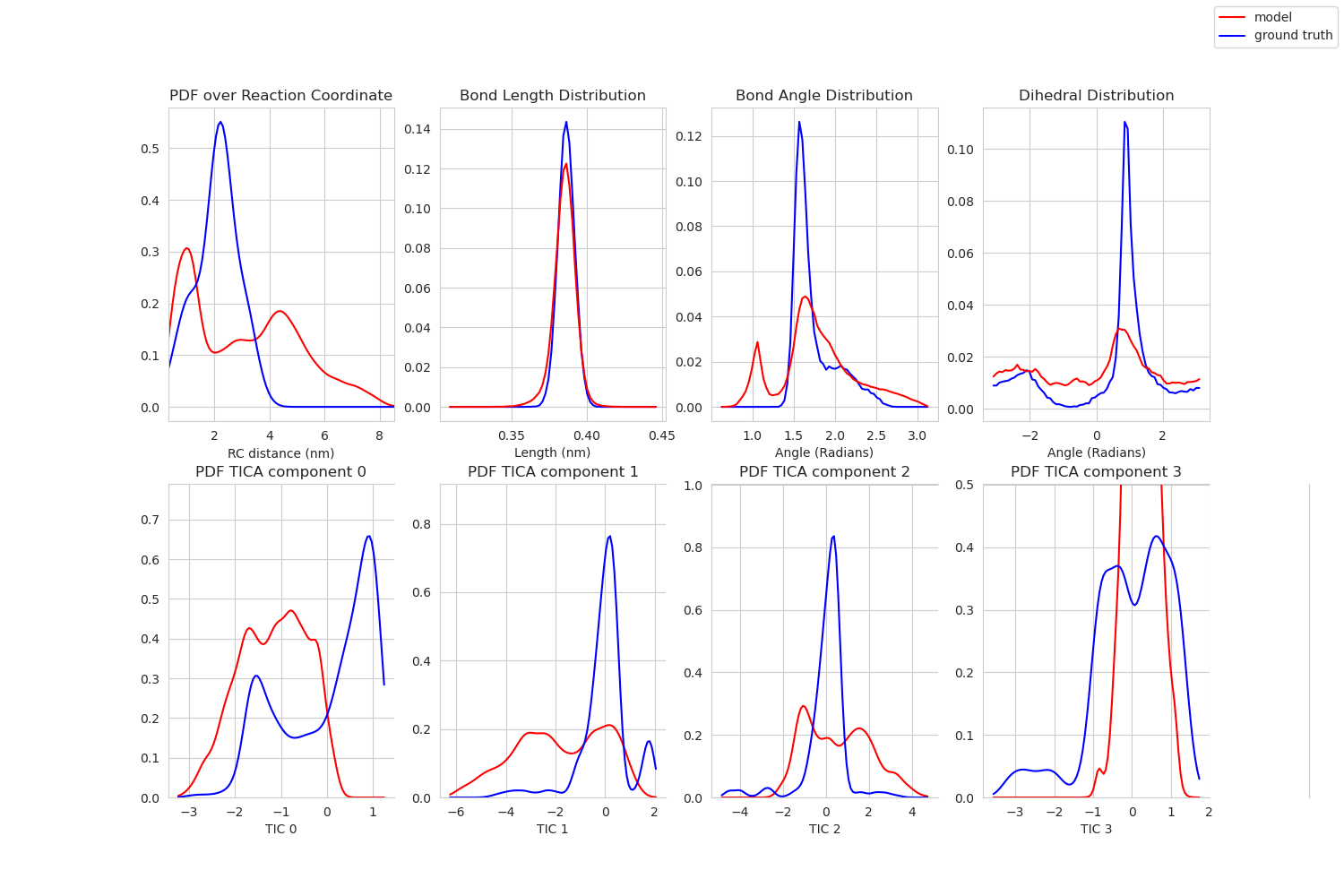}
    \caption{PDFs for Protein B - CGSchNet Under-Trained Model}
    \label{fgr:pdfs_proteinb}
\end{figure}

\begin{figure}[htbp]
    \centering
    \includegraphics[trim={0 0 0 0},clip,width=\linewidth]{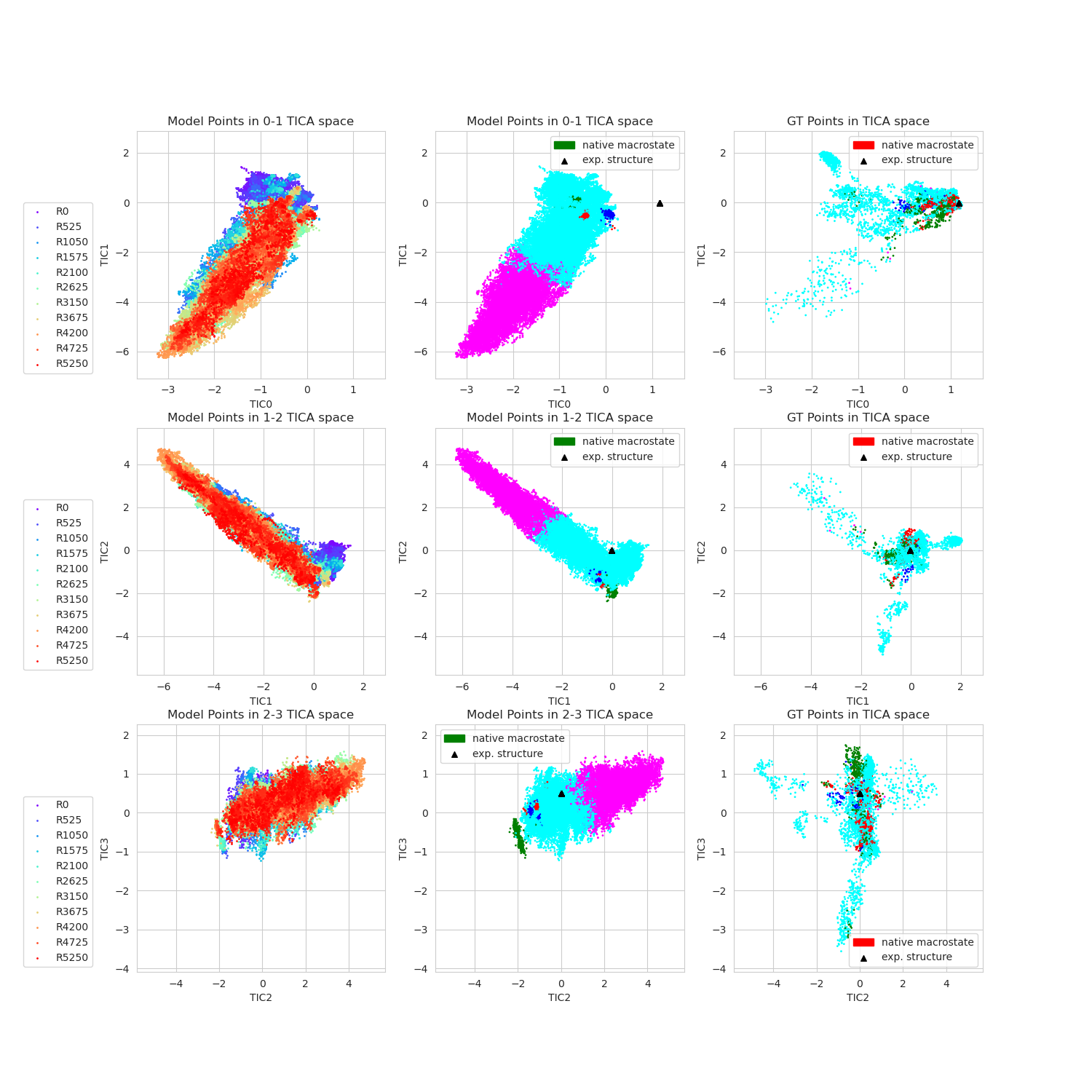}
    \caption{TICA space projections for Protein B - CGSchNet Under-Trained Model}
    \label{fgr:tica_spaces_proteinb}
\end{figure}
\clearpage

\begin{table}[h!]
\centering
\resizebox{\textwidth}{!}{%
\begin{tabular}{|l|c|c|c|c|c|c|c|c|}
\hline
\textbf{Metric} & \textbf{TIC 0} & \textbf{TIC 1} & \textbf{TIC 2} & \textbf{TIC 3} & \textbf{Bonds} & \textbf{Angles} & \textbf{Dihedrals} & \textbf{Gyration} \\
\hline
KL & 5.6274 & 0.9901 & 1.2067 & 1.3488 & 0.0570 & 0.5957 & 0.2948 & 1.9395 \\
W1 & 1.1227 & 1.7827 & 0.8935 & 0.4680 & 0.0014 & 0.1276 & 0.4887 & 0.4768 \\
\hline
\end{tabular}
}
\caption{KL and W1 metrics for Protein B with the CGSchNet Under-Trained Model}
\end{table}

\clearpage

\subsubsection{Protein G}

\begin{figure}[htbp]
    \centering
    \includegraphics[trim={0 0 0 0.75cm},clip,width=0.5\linewidth]{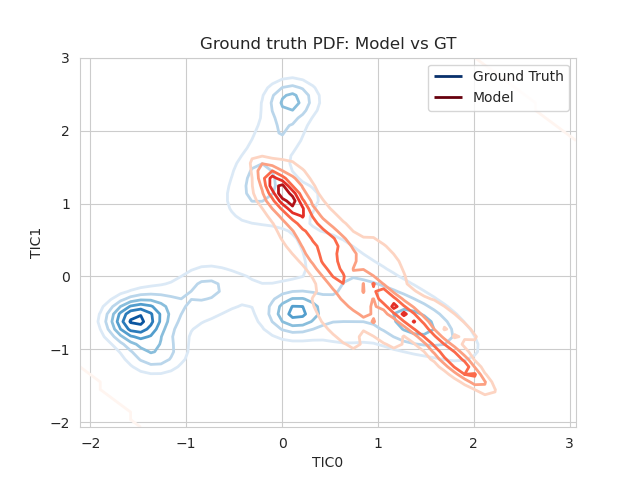}
    \caption{TICA contours for Protein G - CGSchNet Under-Trained Model}
    \label{fgr:tica_contours_proteing}
\end{figure}

\begin{figure}[htbp]
    \centering
    \includegraphics[trim={0 0 0 0.75cm},clip,width=0.5\linewidth]{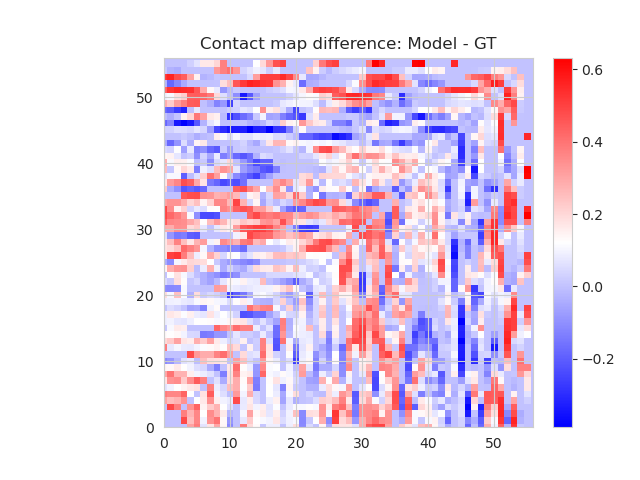}
    \caption{Contact map for Protein G - CGSchNet Under-Trained Model}
\end{figure}

\begin{figure}[htbp]
    \centering
    \includegraphics[trim={0 0 0 0cm},clip,width=\linewidth]{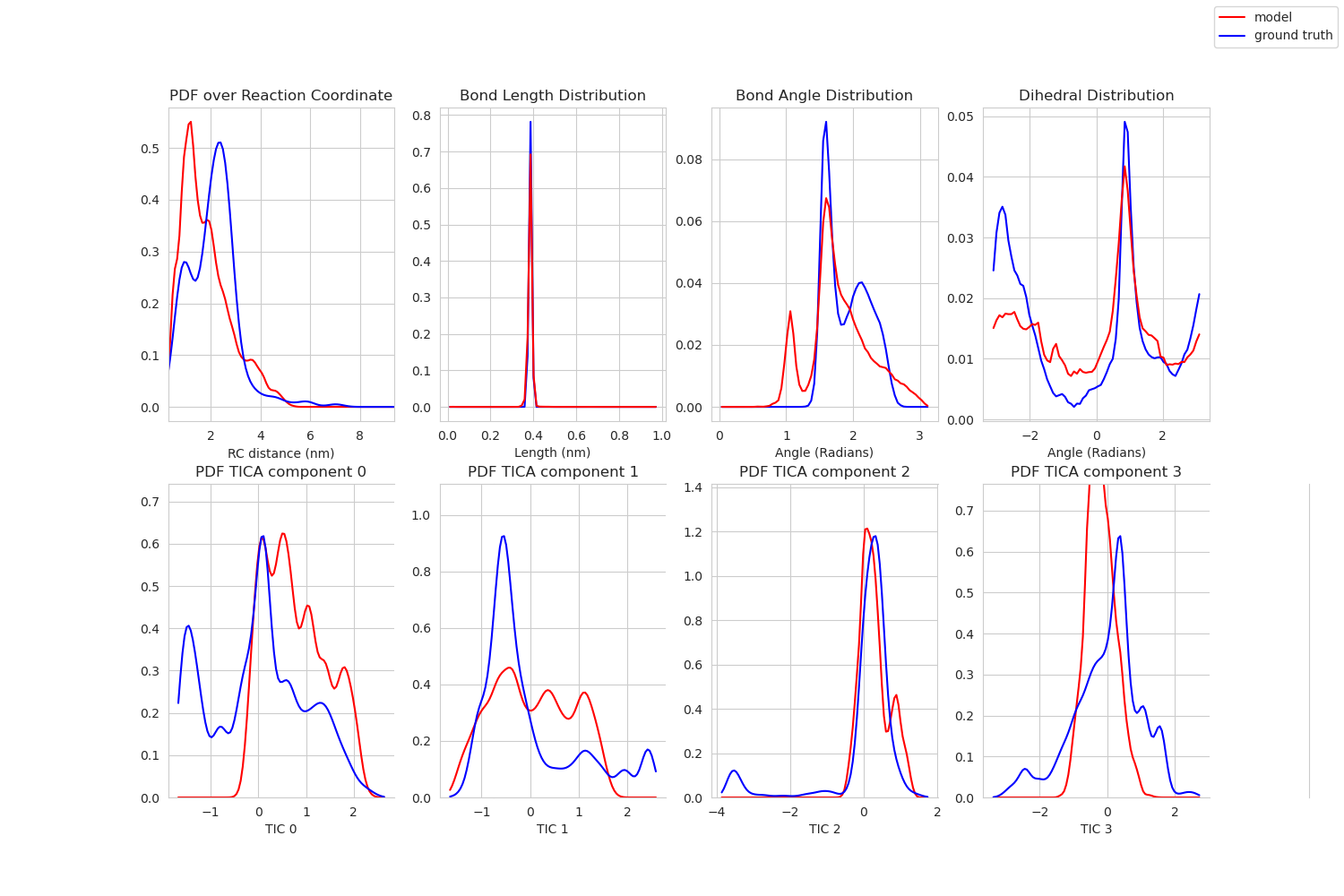}
    \caption{PDFs for Protein G - CGSchNet Under-Trained Model}
    \label{fgr:pdfs_proteing}
\end{figure}

\begin{figure}[htbp]
    \centering
    \includegraphics[trim={0 0 0 0},clip,width=\linewidth]{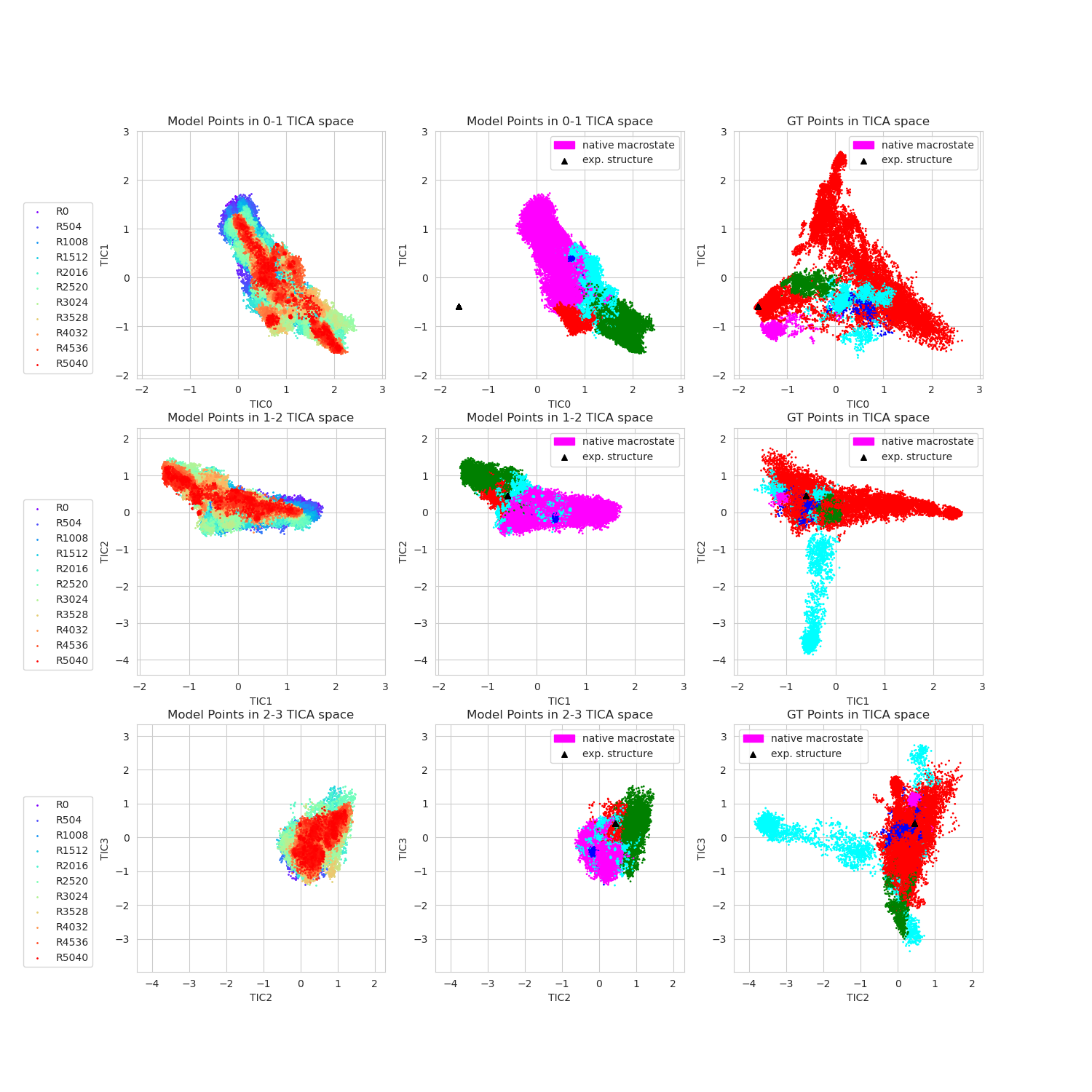}
    \caption{TICA space projections for Protein G - CGSchNet Under-Trained Model}
    \label{fgr:tica_spaces_proteing}
\end{figure}
\clearpage

\begin{table}[h!]
\centering
\resizebox{\textwidth}{!}{%
\begin{tabular}{|l|c|c|c|c|c|c|c|c|}
\hline
\textbf{Metric} & \textbf{TIC 0} & \textbf{TIC 1} & \textbf{TIC 2} & \textbf{TIC 3} & \textbf{Bonds} & \textbf{Angles} & \textbf{Dihedrals} & \textbf{Gyration} \\
\hline
KL & 3.4520 & 1.0588 & 1.0475 & 1.6905 & 0.0708 & 0.2731 & 0.1056 & 0.9801 \\
W1 & 0.7457 & 0.2944 & 0.3005 & 0.5256 & 0.0019 & 0.0471 & 0.2949 & 0.1475 \\
\hline
\end{tabular}
}
\caption{KL and W1 metrics for Protein G with the CGSchNet Under-Trained Model}
\end{table}

\clearpage

\subsubsection{Trp-cage}

\begin{figure}[htbp]
    \centering
    \includegraphics[trim={0 0 0 0.75cm},clip,width=0.5\linewidth]{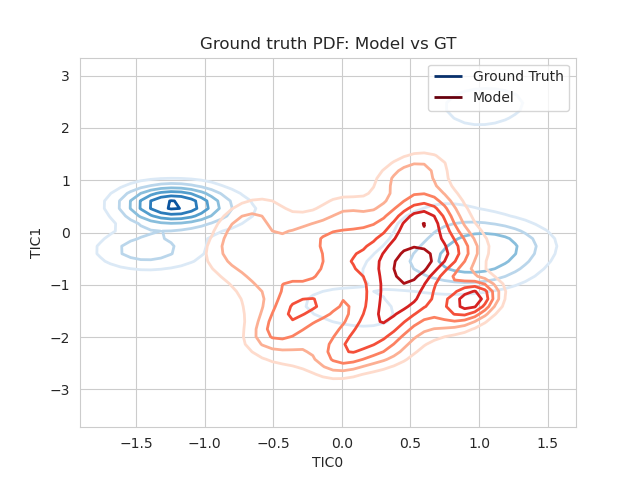}
    \caption{TICA contours for Trp-cage - CGSchNet Under-Trained Model}
    \label{fgr:tica_contours_trpcage}
\end{figure}

\begin{figure}[htbp]
    \centering
    \includegraphics[trim={0 0 0 0.75cm},clip,width=0.5\linewidth]{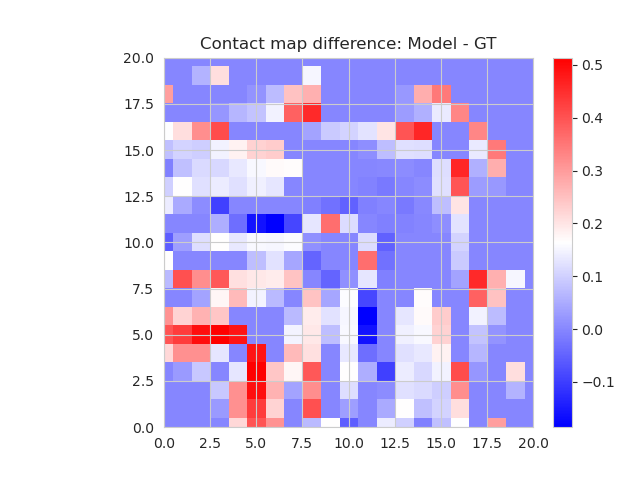}
    \caption{Contact map for Trp-cage - CGSchNet Under-Trained Model}
\end{figure}

\begin{figure}[htbp]
    \centering
    \includegraphics[trim={0 0 0 0cm},clip,width=\linewidth]{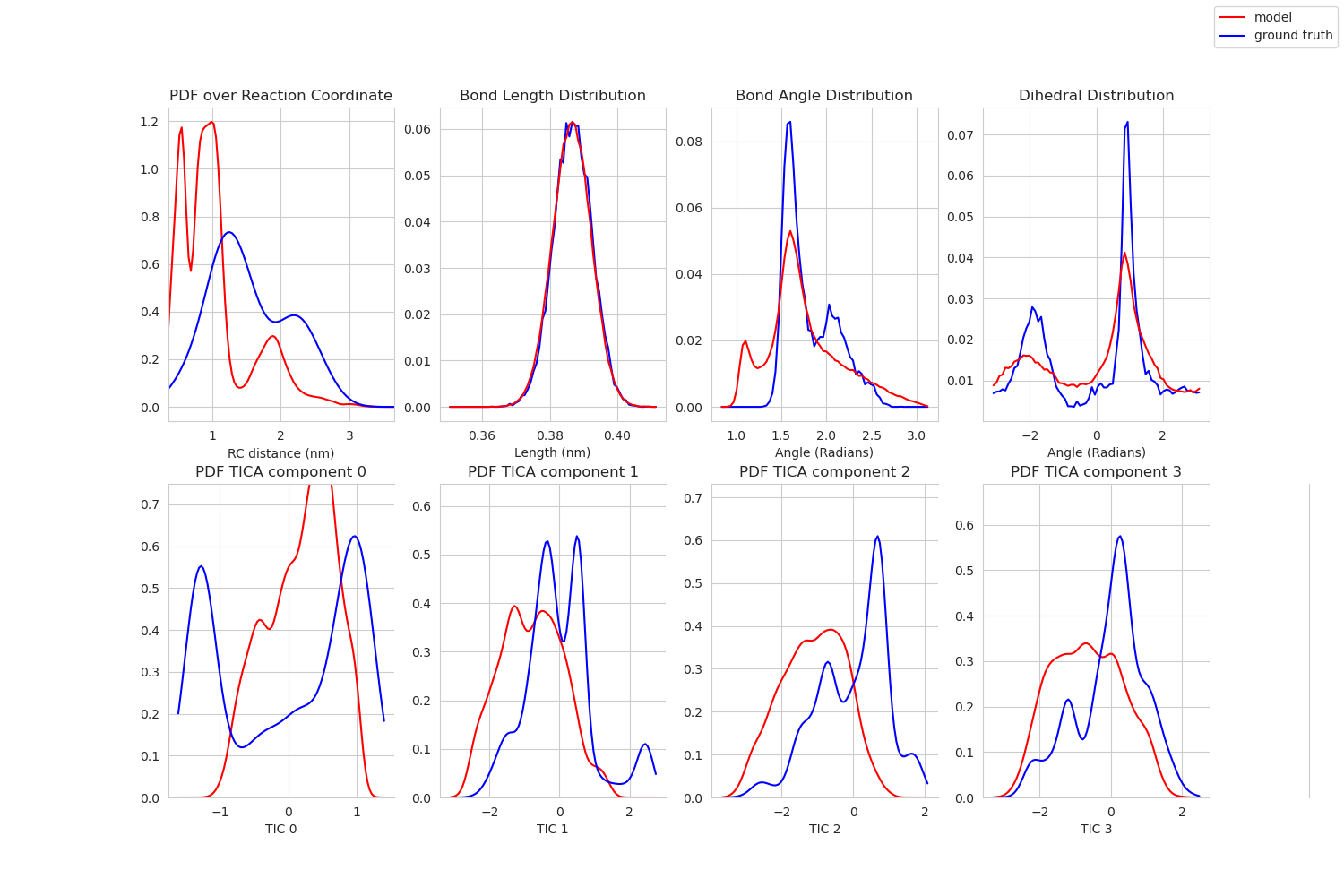}
    \caption{PDFs for Trp-cage - CGSchNet Under-Trained Model}
    \label{fgr:pdfs_trpcage}
\end{figure}

\begin{figure}[htbp]
    \centering
    \includegraphics[trim={0 0 0 0},clip,width=\linewidth]{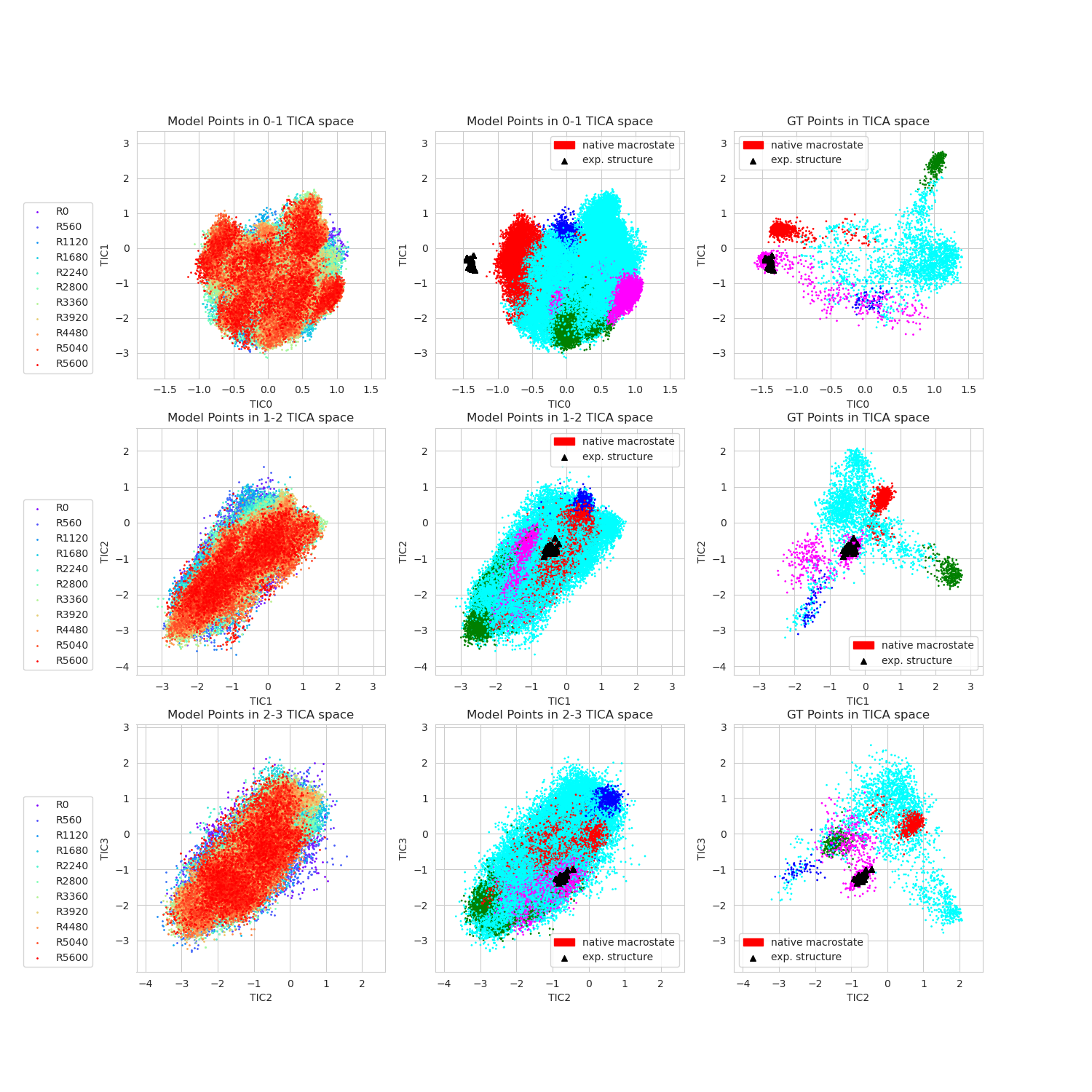}
    \caption{TICA space projections for Trp-cage - CGSchNet Under-Trained Model}
    \label{fgr:tica_spaces_trpcage}
\end{figure}
\clearpage

\begin{table}[h!]
\centering
\resizebox{\textwidth}{!}{%
\begin{tabular}{|l|c|c|c|c|c|c|c|c|}
\hline
\textbf{Metric} & \textbf{TIC 0} & \textbf{TIC 1} & \textbf{TIC 2} & \textbf{TIC 3} & \textbf{Bonds} & \textbf{Angles} & \textbf{Dihedrals} & \textbf{Gyration} \\
\hline
KL & 2.9204 & 1.0706 & 1.3262 & 0.2899 & 0.0024 & 0.1683 & 0.0785 & 1.8363 \\
W1 & 0.4597 & 0.7833 & 1.1118 & 0.6546 & 0.0003 & 0.0565 & 0.1069 & 0.1533 \\
\hline
\end{tabular}
}
\caption{KL and W1 metrics for Trp-cage with the CGSchNet Under-Trained Model}
\end{table}

\clearpage

\subsubsection{WW Domain}

\begin{figure}[htbp]
    \centering
    \includegraphics[trim={0 0 0 0.75cm},clip,width=0.5\linewidth]{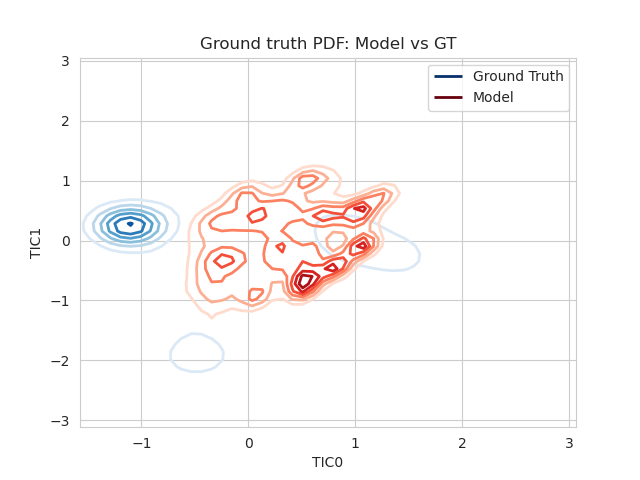}
    \caption{TICA contours for WW Domain - CGSchNet Under-Trained Model}
    \label{fgr:tica_contours_wwdomain}
\end{figure}

\begin{figure}[htbp]
    \centering
    \includegraphics[trim={0 0 0 0.75cm},clip,width=0.5\linewidth]{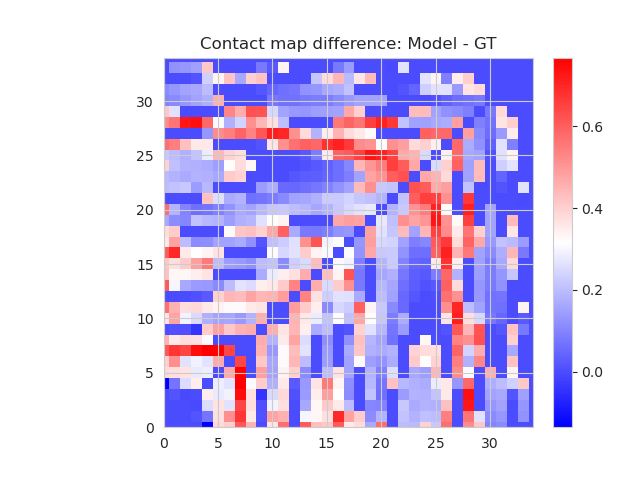}
    \caption{Contact map for WW Domain - CGSchNet Under-Trained Model}
\end{figure}

\begin{figure}[htbp]
    \centering
    \includegraphics[trim={0 0 0 0cm},clip,width=\linewidth]{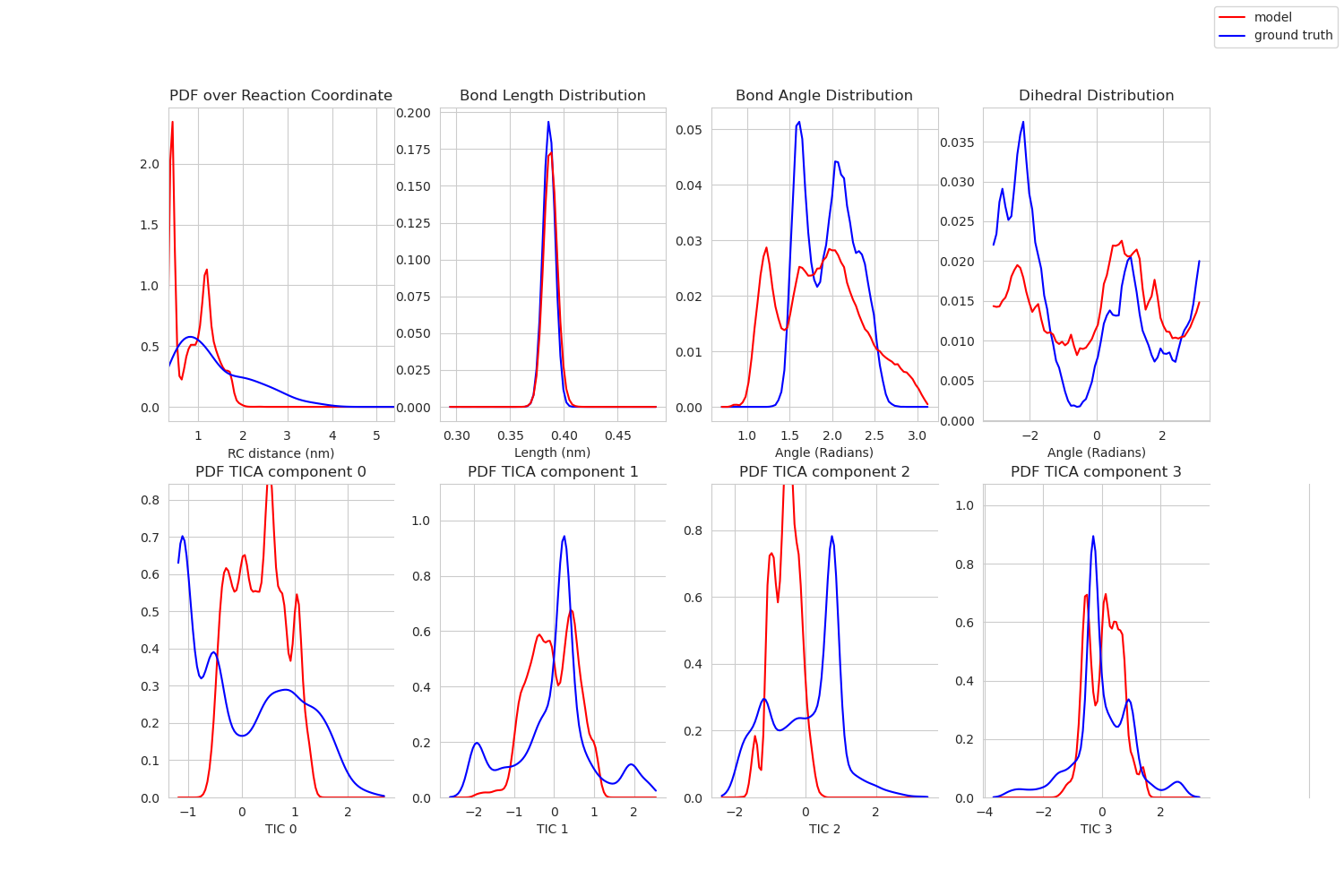}
    \caption{PDFs for WW Domain - CGSchNet Under-Trained Model}
    \label{fgr:pdfs_wwdomain}
\end{figure}

\begin{figure}[htbp]
    \centering
    \includegraphics[trim={0 0 0 0},clip,width=\linewidth]{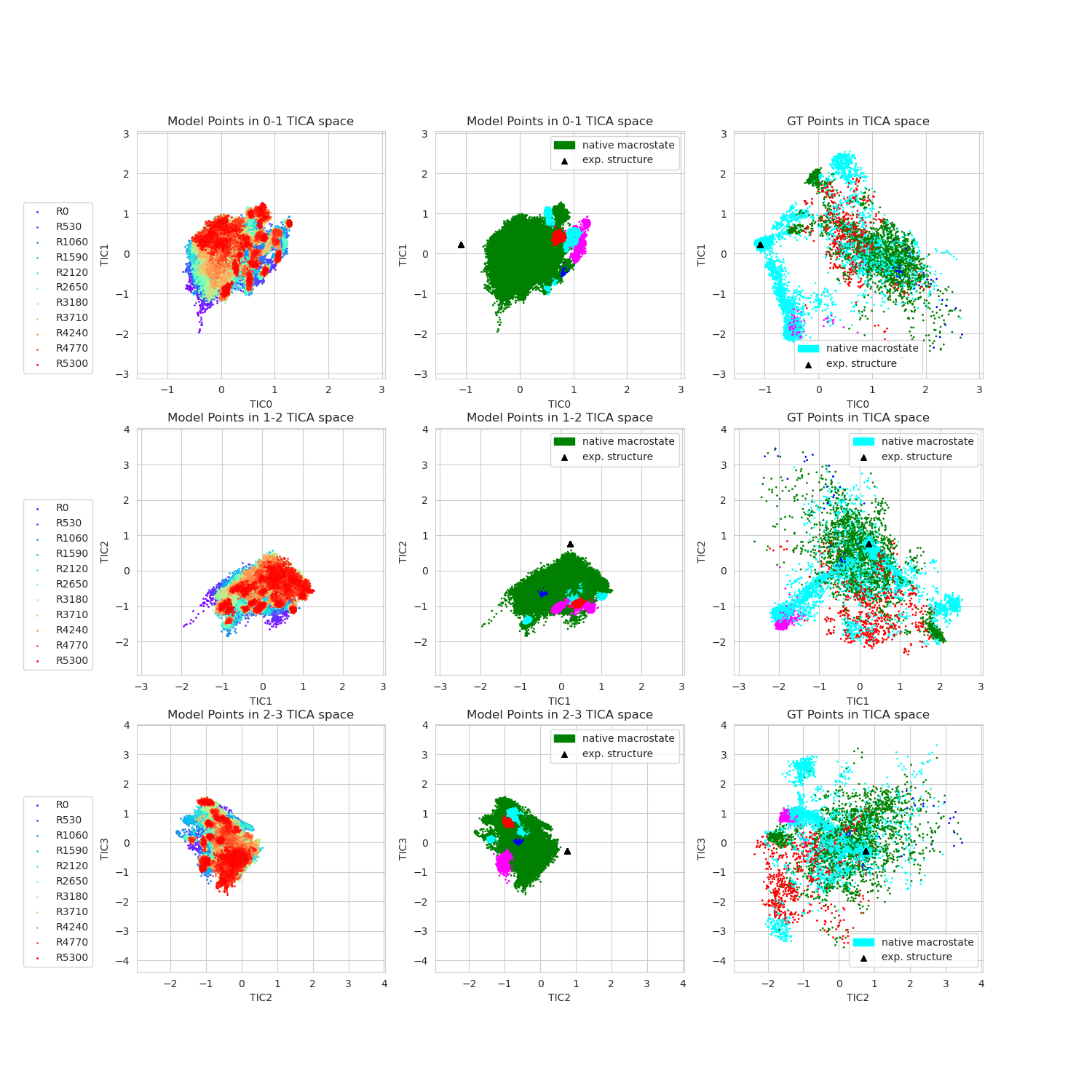}
    \caption{TICA space projections for WW Domain - CGSchNet Under-Trained Model}
    \label{fgr:tica_spaces_wwdomain}
\end{figure}
\clearpage

\begin{table}[h!]
\centering
\resizebox{\textwidth}{!}{%
\begin{tabular}{|l|c|c|c|c|c|c|c|c|}
\hline
\textbf{Metric} & \textbf{TIC 0} & \textbf{TIC 1} & \textbf{TIC 2} & \textbf{TIC 3} & \textbf{Bonds} & \textbf{Angles} & \textbf{Dihedrals} & \textbf{Gyration} \\
\hline
KL & 5.0102 & 1.3795 & 5.4683 & 1.2589 & 0.0465 & 0.3646 & 0.1255 & 6.0218 \\
W1 & 0.4825 & 0.3110 & 0.7272 & 0.3131 & 0.0016 & 0.0659 & 0.5642 & 0.2708 \\
\hline
\end{tabular}
}
\caption{KL and W1 metrics for WWdomain with the CGSchNet Under-Trained Model}
\end{table}

\clearpage